%% file: FINAL_TMLR_fixed.tex
\documentclass[10pt]{article}
\usepackage[preprint]{tmlr}

\input{math_commands.tex}

\usepackage{pifont}
\usepackage{hyperref}
\usepackage{url}
\usepackage{amsmath}
\usepackage{amssymb}
\usepackage{amsthm}
\usepackage{booktabs}
\usepackage{graphicx}
\usepackage{placeins}
\usepackage{needspace}
\usepackage{float}
\usepackage{makecell}
\usepackage{algorithm}
\usepackage{algorithmic}
\usepackage{microtype}

\newtheorem{definition}{Definition}

\newcommand{\cmark}{\ding{51}}

\title{Token Commitment as a Control Surface for Steering Masked Diffusion Generation}

\author{\name Ambroise Aigueperse \email aaiguepe@stanford.edu \\
\addr Stanford University
\AND
\name Hugo Buurmeijer \email hbuurmei@stanford.edu \\
\addr Stanford University
\AND
\name Carmen Amo Alonso \email camoalon@stanford.edu \\
\addr Stanford University
\AND
\name Marco Pavone \email pavone@stanford.edu \\
\addr Stanford University
}

\def\month{MM}
\def\year{YYYY}
\def\openreview{\url{https://openreview.net/forum?id=XXXX}}

\begin{document}

\maketitle

\begin{abstract}
Masked diffusion models repeatedly predict provisional tokens before a sampler decides which predictions become permanent. Existing inference-time steering methods primarily modify activations, logits, or denoising trajectories, while treating this commitment policy as a fixed decoding detail. We instead study token commitment as a diffusion-specific control surface and separate it into two decisions: \emph{which} tokens to commit and \emph{how many} tokens to commit at each step. For the first, we introduce counterfactual target-aware commitment, which ranks unresolved positions by a mixture of model confidence and leave-one-token-out target contribution. For the second, we introduce calibrated adaptive budgets driven by target progress and confidence. Across language and vision-language settings in the LLaDA family, multiple targets, two activation-steering methods, and an unsteered control, both mechanisms yield empirically monotonic target--quality trade-offs in every configuration we evaluate. Adaptive schedules also recover competitive operating points relative to scorer-free stochastic schedules matched in average denoising cost. Because both mechanisms require only a scalar score evaluated on a provisional completion, they apply unchanged when the target is an automatic proxy for task success rather than an output attribute: NLI entailment for dialogue summarization and CLIPScore for captioning, neither of which corresponds to a single reusable steering direction. We reproduce the behavior of both mechanisms on a second masked-diffusion family, Dream-7B, for one control target and one task-success target. These results characterize commitment---which provisional predictions become irreversible, and how rapidly---as an interpretable control surface for steering masked diffusion generation under the conditions we test.
\end{abstract}

\section{Introduction}

Inference-time steering controls a frozen generative model without retraining it, commonly by modifying activations, logits, or guidance signals during generation \citep{dathathri2020plug,turner2023activation,li2023inference,rodriguez2025controlling}. Such interventions are flexible, but stronger steering often improves the target objective at the cost of fluency, grounding, or general generation quality \citep{li2023inference,rodriguez2025controlling,axbench2025}. This motivates control mechanisms that expose interpretable operating points rather than a single fixed intervention strength.

Masked diffusion models expose a decoding decision absent from standard autoregressive generation. Autoregressive models commit tokens in a fixed left-to-right order. Masked diffusion models instead predict provisional tokens at multiple unresolved positions and then choose which predictions become permanent \citep{austin2021structured,sahoo2024simple,nie2025large}. This creates two distinct decisions: \emph{which} tokens to commit and \emph{how many} tokens to commit at each denoising step. The first controls which provisional predictions survive; the second controls how quickly the sequence becomes irreversible. Such models are obtained both by training from scratch with random masking \citep{nie2025large} and by adapting an autoregressive checkpoint \citep{dream2025}; we study LLaDA as the primary family and test whether the observed behavior is reproduced on Dream-7B, an instance of the second construction.

Existing steering methods mostly intervene on what the model predicts, whereas masked-diffusion sampling policies are usually designed for quality, robustness, or acceleration \citep{shnaidman2025activation,ilrr2026,singhal2025feynman,wang2025remasking,klass2025,jazbec2025learning}. We connect these lines of work by making commitment explicitly target-aware. Activation steering changes the provisional predictions produced by the denoiser; commitment control determines which of those predictions survive and how rapidly they become permanent. The two mechanisms are therefore complementary rather than competing. Our goal is not to establish a new best-performing steering method, but to isolate a sampler-level control surface and characterize the operating points it exposes; commitment control acts downstream of whatever intervention produced the provisional predictions, and we hold that intervention fixed throughout so that the commitment weight is the only variable along a sweep.

Our contributions are:
\begin{itemize}
    \item \textbf{Commitment as a steering interface.} Confidence ranking and per-step commitment budgets are standard components of masked-diffusion samplers, but they are treated as decoding details tuned for quality or speed. We give a formal account of the commitment policy that separates it into two components, the scoring map and the budget map, and recast both as \emph{target-aware} control axes. (Section~\ref{sec:problem})
    \item \textbf{Two target-aware controls.} We introduce counterfactual target-aware commitment order and calibrated adaptive commitment budgets, yielding separate controls over token survival and commitment rate. Both require only a scalar target score evaluated on a provisional completion, so both apply to output attributes and to task-success measures alike. (Section~\ref{sec:approach})
    \item \textbf{Controlled empirical characterization.} We evaluate language and vision-language generation, multiple objectives, and no activation steering, DiffMean, and Activation Transport (ACT), across two masked-diffusion model families. Both controls produce empirically monotonic trade-offs. A scorer-free, non-adaptive stochastic schedule matched in average denoising steps separates the effect of state-dependent adaptation from average compute and provides the principal baseline for adaptive-budget competitiveness. (Section~\ref{sec:results})
\end{itemize}

\section{Related Work}
\label{sec:related-work}

\paragraph{Inference-time steering.}
Plug-and-play language models use gradients from an external attribute model to steer hidden states during decoding \citep{dathathri2020plug}. Others edit internal representations directly, adding directions derived from contrastive example sets or linear probes \citep{turner2023activation,li2023inference}. Activation Transport casts steering as an optimal-transport problem, applying affine maps that move activations toward a target distribution while preserving its shape \citep{rodriguez2025controlling}. Systematic evaluations show that these methods trade target strength against utility and are sensitive to layer, strength, and method design \citep{axbench2025}. All of them modify \emph{what the model predicts} at a given step. For autoregressive models this is the only available surface, because the order in which predictions become final is fixed by the left-to-right factorization.

\paragraph{Steering diffusion language models.}
Several recent methods port these ideas to diffusion language models. \citet{shnaidman2025activation} inject contrastive activation directions during masked denoising and localize the greatest leverage to early denoising steps and mid-to-late transformer layers. ILRR aligns denoising activations against a reference sequence \citep{ilrr2026}. DLM-Scope trains sparse autoencoders on diffusion-language-model activations and intervenes on the resulting features \citep{dlmscope2026}. Concurrent work develops an optimal-transport view of activation steering for masked diffusion models, learning affine transport maps between pooled source and target activation distributions \citep{otmdm2026}. Sequential-Monte-Carlo and Feynman--Kac methods instead reweight and resample whole denoising trajectories according to a reward \citep{singhal2025feynman}. We do not contribute a new activation intervention: we use DiffMean and diagonal ACT as two independently constructed steered operating points, and intervene only after provisional predictions have been produced.

\paragraph{Masked-diffusion decoding policies.}
Masked diffusion samplers must choose which unresolved positions to reveal at each step, and this choice strongly affects both quality and speed \citep{sahoo2024simple,nie2025large,you2025lladav}. Confidence, entropy, and margin heuristics are standard, and Dream exposes the position-ranking rule directly as a decoding argument \citep{dream2025}, which is the interface our commitment-order control replaces. ReMDM permits previously revealed tokens to be remasked, enabling inference-time refinement \citep{wang2025remasking}; KLASS uses token-level KL stability across denoising steps to accelerate inference \citep{klass2025}; CoRe identifies context-brittle tokens by perturbing masked context and prioritizes unstable predictions for revision \citep{core2026}; \citet{jazbec2025learning} learn unmasking policies with reinforcement learning, treating token selection as a quality--cost decision problem. These rules all score positions using signals internal to the model---likelihood, stability, or a learned value---and optimize quality, efficiency, or robustness. The trajectory-level methods above do use an external reward, but resample whole completions. We instead condition the within-trajectory commitment decision on an external, user-specified target score, and treat the per-step budget as a second exposed control rather than a fixed schedule.

\section{Problem Formulation}
\label{sec:problem}

\subsection{Masked-diffusion generation}

Let $\mathcal V$ be the vocabulary, $L\in\mathbb N$ the output length, and $\mathcal C$ the space of conditioning inputs. Write $\mathcal X=(\mathcal V\cup\{\texttt{[MASK]}\})^L$ for the space of generation states. At denoising step $t$, the state for a conditioning input $c\in\mathcal C$ is $x_t\in\mathcal X$, and
\[
    M_t=\{i\in\{1,\ldots,L\}: x_{t,i}=\texttt{[MASK]}\}
\]
is the set of unresolved positions, with $M_0=\{1,\ldots,L\}$ and $M_{t+1}\subsetneq M_t$.

A fixed, possibly steered denoiser is a map
\[
    p_\theta:\mathcal X\times\mathcal C\to\Delta(\mathcal V)^L,
\]
where $\Delta(\mathcal V)$ denotes the probability simplex over $\mathcal V$ and $p_i(\cdot\mid x_t,c)$ is the $i$th coordinate of $p_\theta(x_t,c)$. Its provisional prediction and confidence at an unresolved position $i\in M_t$ are
\[
    \hat x_{t,i}=\arg\max_{v\in\mathcal V}p_i(v\mid x_t,c),
    \qquad
    q_{t,i}=\max_{v\in\mathcal V}p_i(v\mid x_t,c),
\]
where $\hat x_{t,i}\in\mathcal V$ and $q_{t,i}\in[0,1]$. Both $p_\theta$ and any activation intervention applied to it are held fixed throughout each commitment sweep, so that the commitment policy is the only quantity that varies.

\subsection{Commitment order and budget}

Producing provisional predictions does not determine the output: the sampler must still decide which of them become permanent, and how many at a time. We formalize that decision as a pair of maps.

\begin{definition}[Commitment order]
\label{def:order}
Let $r:\mathcal X\times\mathcal C\times\{1,\ldots,L\}\to\mathbb R$ be a scoring map and write $r_{t,i}=r(x_t,c,i)$. The \emph{commitment order} induced by $r$ at step $t$ is the total order $\succeq_t$ on $M_t$ given by
\[
    i\succeq_t j
    \iff
    r_{t,i}>r_{t,j}
    \ \ \text{or}\ \
    \bigl(r_{t,i}=r_{t,j}\ \text{and}\ i\le j\bigr),
    \qquad i,j\in M_t .
\]
It determines the sequence in which provisional predictions are made permanent.
\end{definition}

\begin{definition}[Commitment budget]
\label{def:budget}
Let $u:\mathcal X\times\mathcal C\to\mathbb N$ be a budget map and write $u_t=u(x_t,c)$, subject to $1\le u_t\le|M_t|$. The \emph{commitment budget} at step $t$ is $u_t$, and the \emph{committed set} $C_t\subseteq M_t$ is the unique $\succeq_t$-maximal subset of $M_t$ with $|C_t|=u_t$, which we write as
\[
    C_t=\operatorname{Top}_{u_t}\{r_{t,i}: i\in M_t\}.
\]
Predictions at positions in $C_t$ become permanent, so $M_{t+1}=M_t\setminus C_t$; positions in $M_{t+1}$ remain revisable. The budget determines how quickly the sequence becomes irreversible.
\end{definition}

The pair $(r,u)$ is a \emph{commitment policy}. Because $\succeq_t$ is a total order, $C_t$ in Definition~\ref{def:budget} is well defined for every $u_t$. The two components are independent: $r$ fixes the order in which positions are resolved, while $u$ fixes the rate at which the order is consumed.

Standard confidence-based decoding instantiates the policy with $r_{t,i}=q_{t,i}$ and a budget schedule fixed in advance, so neither component depends on any property of the text being generated. Given the definitions above, the open question is whether the scoring map $r$ and the budget map $u$ can instead be made to depend on a user-specified target score evaluated on the current provisional output, and whether doing so exposes controllable operating points in the joint space of target, quality, and denoising cost. Section~\ref{sec:approach} constructs one such map for each component, holding the other fixed, so that the two axes can be characterized separately.

\section{Target-Aware Commitment Controls}
\label{sec:approach}

Let $S(y;\xi)\in\mathbb R$ be a frozen scalar target score for a decoded output $y$, where $\xi$ contains any task-specific context and larger values indicate stronger target satisfaction; token sequences are decoded to text before they are passed to $S$. The two controls act on the two components of Definitions~\ref{def:order} and~\ref{def:budget}. Counterfactual commitment replaces the scoring map $r$ while holding $u$ fixed, isolating which predictions survive. Adaptive budgeting replaces the budget map $u$ while holding $r$ fixed to confidence, isolating how rapidly predictions become irreversible. Both use a weight in $[0,1]$: the weight multiplies confidence and its complement multiplies the target signal. We denote the two weights by $\lambda_{\mathrm{order}}$ and $\lambda_{\mathrm{budget}}$.

\subsection{Counterfactual target-aware commitment}
\label{sec:approach-order}

At step $t$, let $\tilde x_t$ be the full provisional completion obtained by replacing every position $i\in M_t$ with $\hat x_{t,i}$. We delete the decoded text span corresponding to each provisional token and denote the completion with position $i$ removed by $A_i(\tilde x_t)$. Its local target contribution is
\[
    \Delta_{t,i}=S(\tilde x_t;\xi)-S(A_i(\tilde x_t);\xi),
\]
where $\xi$ is held fixed across the comparison. A larger positive $\Delta_{t,i}$ means that the provisional token at position $i$ supports the target more strongly. We min--max normalize these contributions within the current step to $\bar\Delta_{t,i}\in[0,1]$ and take the scoring map of Definition~\ref{def:order} to be
\begin{equation}
    r_{t,i}^{\mathrm{cf}}
    =
    \lambda_{\mathrm{order}}\,q_{t,i}
    +(1-\lambda_{\mathrm{order}})\,\bar\Delta_{t,i},
    \label{eq:counterfactual-score}
\end{equation}
where $\lambda_{\mathrm{order}}\in[0,1]$ mixes model confidence $q_{t,i}$ with normalized target contribution $\bar\Delta_{t,i}$, so that $r_{t,i}^{\mathrm{cf}}\in[0,1]$. At $\lambda_{\mathrm{order}}=1$ the method reduces to confidence ranking; lower values place more weight on target contribution. In the main experiments $u_t\equiv1$, so varying $\lambda_{\mathrm{order}}$ changes only the commitment order and leaves denoising cost fixed.

\begin{figure}[t]
    \centering
    \includegraphics[width=\linewidth]{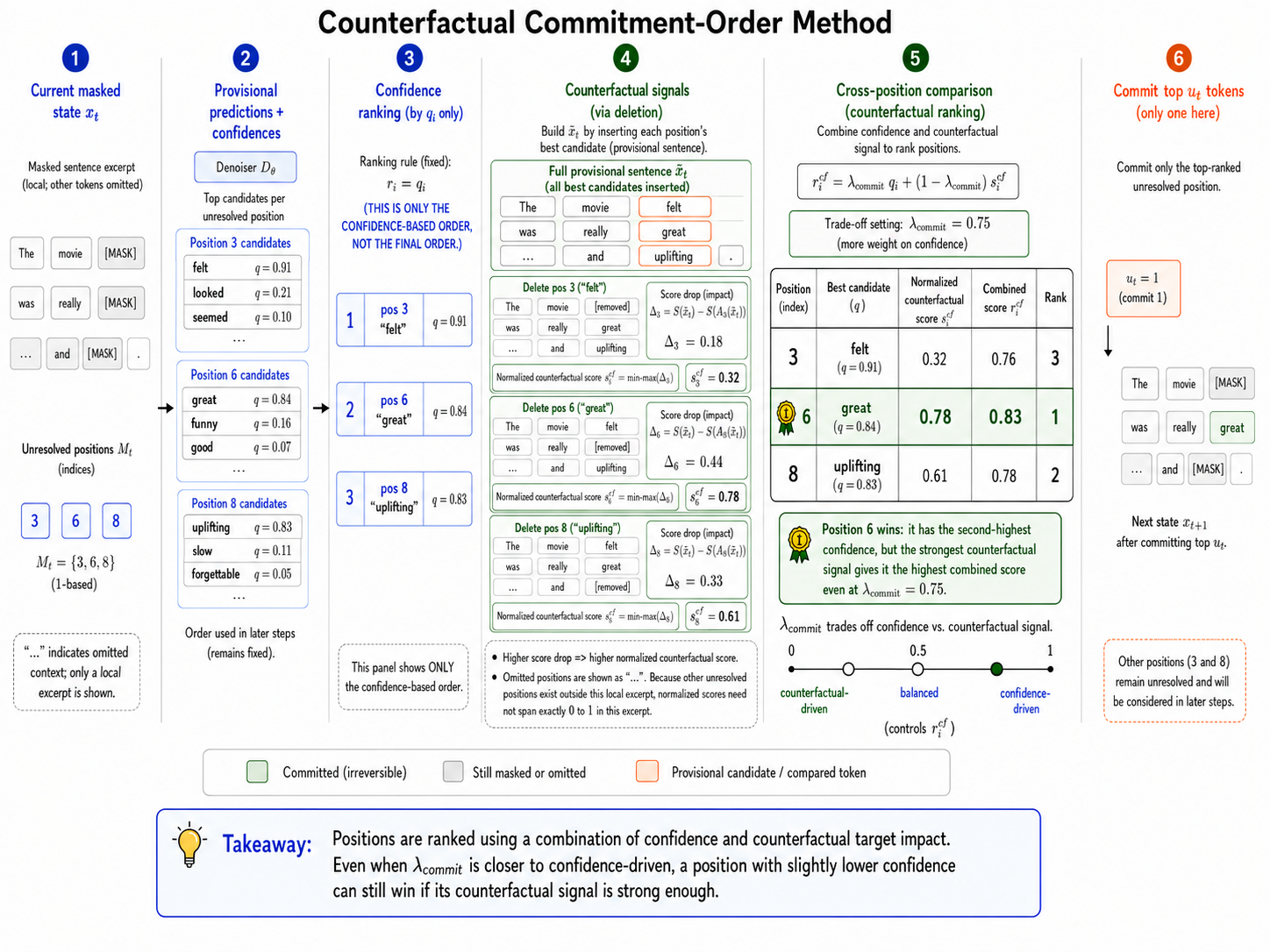}
    \caption{\textbf{Counterfactual target-aware commitment order.} Each unresolved position contributes one provisional candidate. We measure its local target contribution by deleting its decoded span, combine the normalized score drop with model confidence, and commit the highest-scoring position. With $\lambda_{\mathrm{order}}=0.75$, the second-most-confident position can still win when its counterfactual contribution is sufficiently stronger.}
    \label{fig:method-counterfactual}
\end{figure}

The term \emph{counterfactual} refers to the local span-deletion comparison on the current provisional completion, not to the downstream causal effect of committing a token. The deletion variants differ only in the removed position and are scored in one batched call. Exact span construction, normalization, and pseudocode are in Appendix~\ref{app:counterfactual-details}.

\subsection{Adaptive commitment budgets}
\label{sec:approach-budget}

We next hold the scoring map fixed to confidence and replace only the budget map. From the provisional completion we compute target progress $s_t=S(\tilde x_t;\xi)$ and confidence $h_t\in[0,1]$, defined as the mean confidence of the top $U_{\max}$ unresolved candidates. Because their scales and trajectory-dependent distributions differ, we transform both signals into phase-specific empirical quantiles $\hat s_t,\hat h_t\in[0,1]$ using a calibration set disjoint from evaluation. We then set
\begin{equation}
    z_t=(1-\lambda_{\mathrm{budget}})\hat s_t+\lambda_{\mathrm{budget}}\hat h_t,
    \qquad
    u_t=\min\!\left(1+\lfloor U_{\max}z_t\rfloor,\;U_{\max},\;|M_t|\right),
    \label{eq:adaptive-budget}
\end{equation}
where $\lambda_{\mathrm{budget}}\in[0,1]$ mixes calibrated target progress and confidence, $z_t\in[0,1]$ is the resulting mixed signal, $U_{\max}\in\mathbb N$ is the maximum budget, and $\lfloor\cdot\rfloor$ is the floor function. Thus $\lambda_{\mathrm{budget}}=0$ is target-progress-driven and $\lambda_{\mathrm{budget}}=1$ is confidence-driven. We use $U_{\max}=3$, so $u_t\in\{1,2,3\}$ except near completion, while the scoring map remains $r_{t,i}=q_{t,i}$ for every $\lambda_{\mathrm{budget}}$.

\begin{figure}[t]
    \centering
    \includegraphics[width=\linewidth]{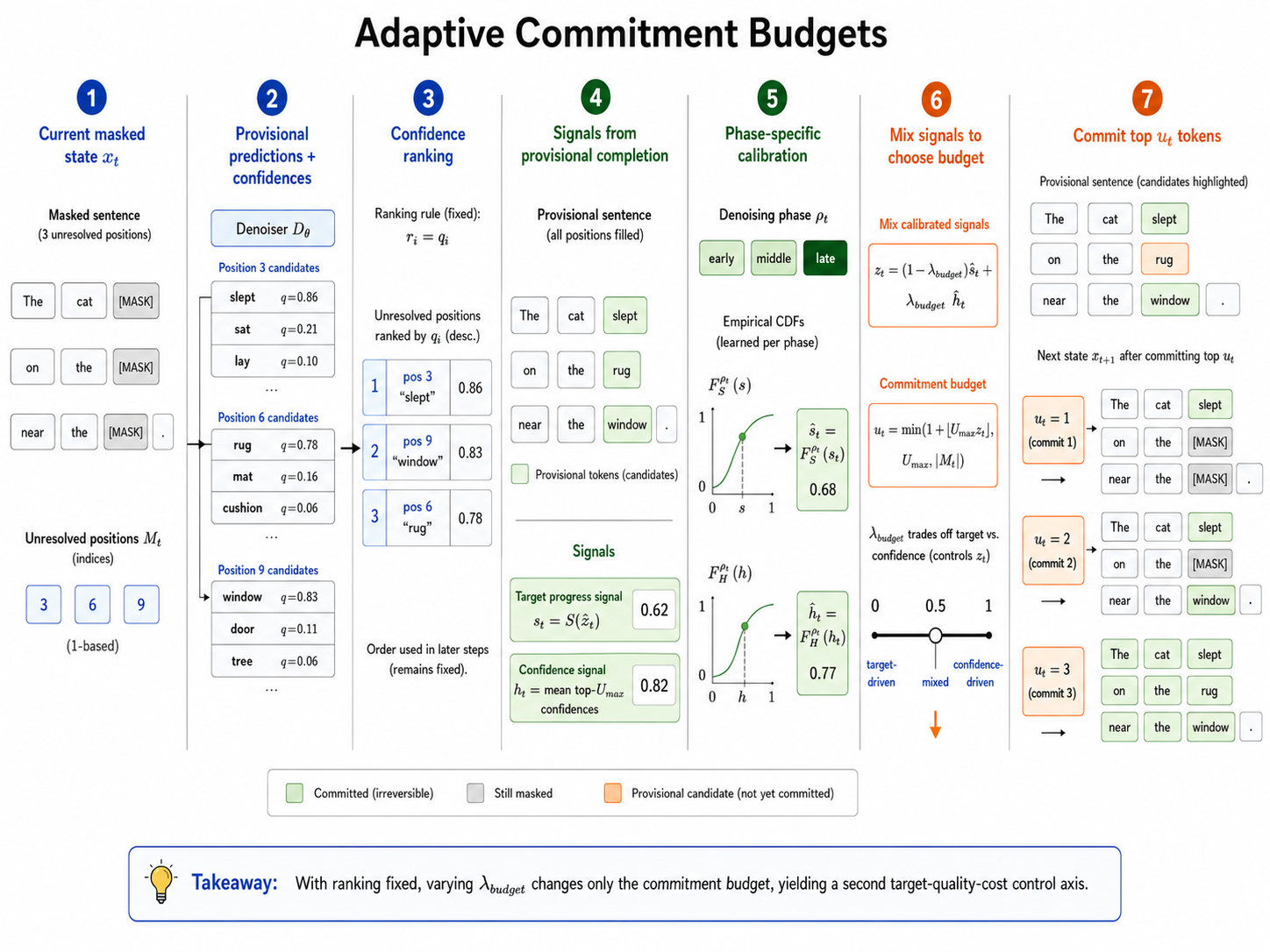}
    \caption{\textbf{Adaptive commitment budgets.} Provisional predictions are always ranked by confidence. Target progress and top-candidate confidence are calibrated within the current denoising phase, mixed through $\lambda_{\mathrm{budget}}$, and mapped to $u_t\in\{1,2,3\}$. Increasing $u_t$ commits more of the same confidence ranking and leaves fewer positions revisable.}
    \label{fig:method-adaptive}
\end{figure}

Nothing in \eqref{eq:adaptive-budget} depends on $U_{\max}=3$: the construction applies to any maximum budget, and we recommend estimating the calibration quantiles on a constant budget near the middle of the intended range. We therefore estimate them once from fixed-$2$ runs, the central member of $\{1,2,3\}$, and reuse them across evaluation configurations rather than tuning them per setting. Adaptive budgeting costs one target-scorer evaluation per step. Phase definitions, calibration, and pseudocode are in Appendix~\ref{app:adaptive-details}.

\section{Experimental Setup}
\label{sec:setup}

\paragraph{Targets and metrics.}
We distinguish \emph{control targets}, such as sentiment and subjectivity, which express a shared output attribute, from \emph{task-success targets}, which depend on the particular input. We instantiate the latter with NLI entailment between a source dialogue and its generated summary and with CLIPScore as a proxy for image--text alignment. Because these targets are example-specific, their scores are recomputed for every provisional completion and no reusable activation direction is required. We report the frozen target score, external-language-model perplexity as a fluency measure, and mean denoising steps as the principal compute measure. NLI entailment and CLIPScore are automatic proxies rather than complete measures of task quality. Exact scorer definitions, checkpoints, validity diagnostics, and metric details are in Appendix~\ref{app:formal-details}.

\begin{table}[t]
\caption{Evaluation settings. Sentiment, subjectivity and caption sentiment are control targets; NLI entailment and CLIPScore are task-success targets and use no activation intervention, isolating commitment control. The last two rows are the second-family confirmation of Appendix~\ref{app:dream}, which covers one control target and one task-success target.}
\label{tab:experiment-summary}
\centering
\scriptsize
\setlength{\tabcolsep}{3.5pt}
\begin{tabular}{lllll}
\toprule
Setting & Model & Data ($N$) & Activation & Target \\
\midrule
Sentiment & LLaDA & Amazon ($300$) & None / DiffMean / ACT & Negative sentiment \\
Subjectivity & LLaDA & Amazon ($300$) & None / DiffMean / ACT & Subjectivity \\
Summarization & LLaDA & SAMSum ($300$) & None & NLI entailment \\
Caption sentiment & LLaDA-V & SentiCap ($100$) & None / DiffMean / ACT & Negative sentiment \\
Image alignment & LLaDA-V & SentiCap ($100$) & None & CLIPScore \\
\midrule
Sentiment & Dream-7B & Amazon ($300$) & None / DiffMean / ACT & Negative sentiment \\
Summarization & Dream-7B & SAMSum ($300$) & None & NLI entailment \\
\bottomrule
\end{tabular}
\end{table}

We use LLaDA-8B-Instruct for text generation and LLaDA-V for captioning \citep{nie2025large,you2025lladav}. Generation is deterministic except for the explicitly identified stochastic budget control, with masked output lengths $60$ and $35$, respectively. Commitment-order experiments fix $u_t\equiv1$ and sweep $\lambda_{\mathrm{order}}\in\{0.5,0.75,0.9,1\}$. Adaptive experiments fix confidence ranking and sweep $\lambda_{\mathrm{budget}}\in\{0,0.5,1\}$.

\paragraph{Baselines and controlled comparisons.}
Fixed budgets $u_t\equiv1,2,3$ show the coarse constant-budget landscape. To separate state-dependent adaptation from average denoising cost, the primary adaptive baseline is a scorer-free stochastic schedule that uses the same confidence ranking but samples each budget from a state-independent distribution. Its neighboring integer budgets and sampling probability are selected post hoc from denoising-step counts alone---never from target score or perplexity---to match the adaptive reference in mean steps. The sampled budget is independent of the scorer, prompt, confidence values, phase, and denoising state. Full construction details are in Appendix~\ref{app:matched-schedule-details}.

For control targets, we test no activation intervention, DiffMean, and diagonal ACT using one layer window, $\ell\in\{9,\ldots,24\}$, held fixed across every target and commitment configuration, and each intervention is held at a single representative strength, so that the commitment weight is the only variable along a sweep. These are representative unsteered and steered operating points, not optimized estimates of either steering method's best achievable performance.

Adaptive signals are calibrated on fixed-$2$ runs over a held-out prompt set that is disjoint from the evaluation prompts: $500$ prompts for the text settings and $125$ for the vision-language settings. No evaluation prompt is used to estimate any calibration statistic. Full prompts, checkpoints, steering strengths, layer choices, matched-schedule construction, calibration details, implementation settings, and runtime protocol are in Appendix~\ref{app:experimental_details}.

\paragraph{Second model family.}
The formulation requires only that the sampler separate provisional prediction from token commitment, so whether the \emph{empirical} behavior depends on the LLaDA family is a separate question. We therefore repeat both mechanisms on Dream-7B \citep{dream2025} for two settings: sentiment control, with all three activation conditions, and NLI entailment on SAMSum without activation steering. Dream is initialized from autoregressive weights rather than trained from scratch with random masking, and differs from LLaDA in tokenizer, depth, and confidence calibration, so it exercises a different denoiser prior while retaining the commitment structure the methods act on. The commitment configuration is unchanged; the steering layer window, the intervention strengths, and the adaptive calibration are re-derived for Dream (Appendix~\ref{app:dream-setup}). Covering both a control target and a task-success target means model-family breadth and target-type breadth are not established on disjoint settings.

\section{Results}
\label{sec:results}

We isolate the two commitment components. Section~\ref{sec:results-order} fixes $u_t\equiv1$ and varies only the scoring map. Section~\ref{sec:results-budget} fixes confidence ranking and varies only the budget map, using the compute-matched stochastic schedule as the principal controlled baseline. Section~\ref{sec:results-task} evaluates the task-success targets. Complete numeric tables and per-configuration plots are in Appendix~\ref{app:full-tables}; the Dream-7B results are in Appendix~\ref{app:dream}.

\subsection{Commitment order gives an empirically monotonic control knob}
\label{sec:results-order}

\begin{figure}[t]
    \centering
    \includegraphics[width=0.68\linewidth]{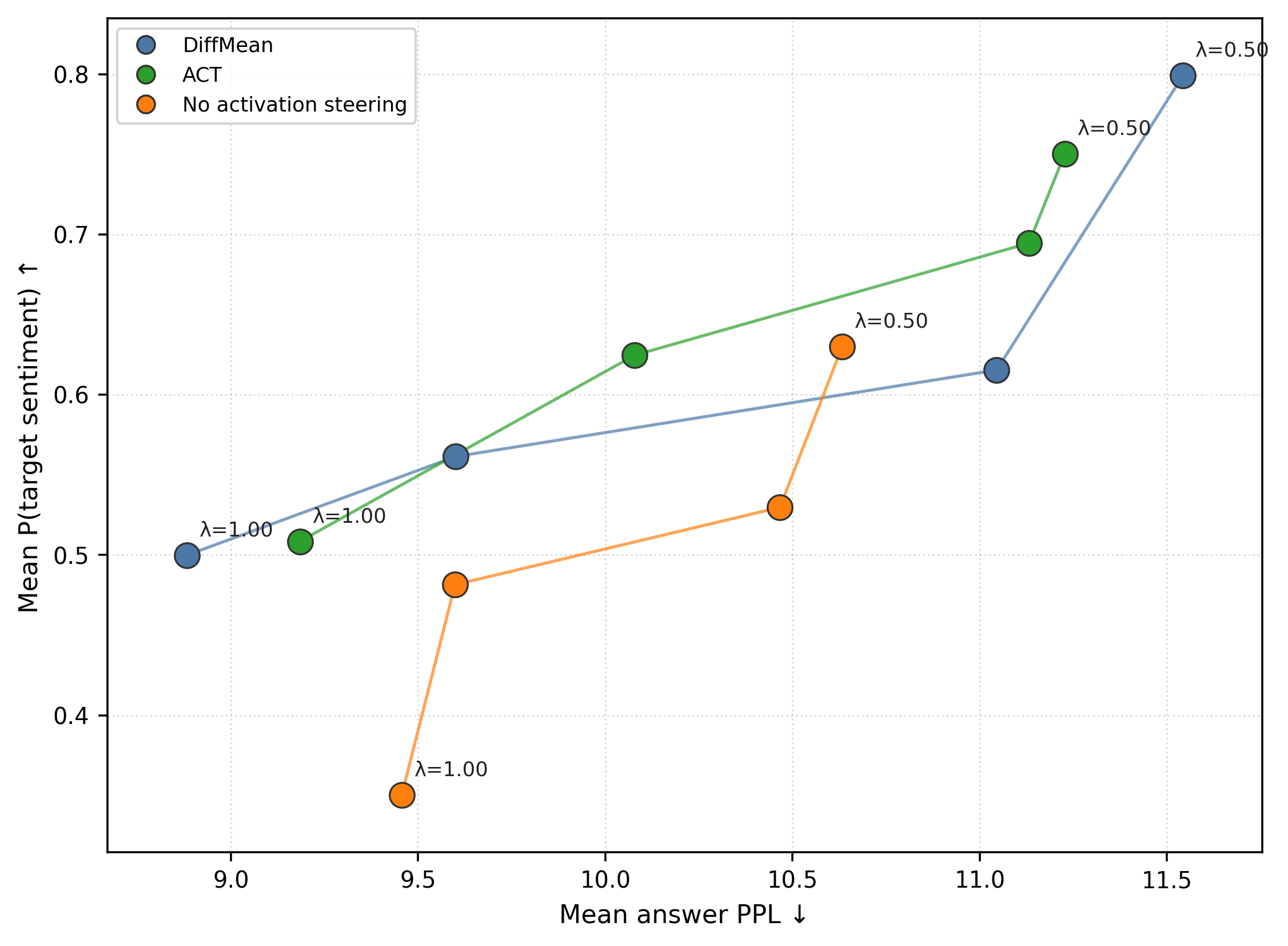}
    \caption{Counterfactual commitment-order trade-offs for LLaDA sentiment. We vary $\lambda_{\mathrm{order}}$ while fixing $u_t\equiv1$, so movement along each curve changes only which provisional token is committed next.}
    \label{fig:counterfactual-sentiment}
\end{figure}

Figure~\ref{fig:counterfactual-sentiment} shows empirically monotonic target--quality movement under all three activation conditions: decreasing $\lambda_{\mathrm{order}}$ consistently increases both the target score and PPL. Because the budget is fixed, every point within an activation condition uses the same number of denoising steps, so the movement is attributable to commitment order rather than extra refinement computation. The no-activation curve further shows that commitment alone can substantially alter the final target score without changing activations or logits.

DiffMean and ACT shift the location and range of the curve, but $\lambda_{\mathrm{order}}$ moves every condition in the same direction: activation steering sets the base operating point, while commitment determines which predictions become irreversible. Table~\ref{tab:main-generalization} shows the same empirical ordering for subjectivity and vision-language caption sentiment. Moving from confidence-only commitment to the strongest tested counterfactual mixture increases the target by $0.15$--$0.23$ for subjectivity and by $0.15$--$0.56$ for caption sentiment. The response magnitude depends on the target and intervention, but the direction is unchanged across all tested control settings.

\begin{table}[t]
\caption{Generalization of counterfactual commitment order across control targets and across model families. We report the full target-score and PPL sequences as $\lambda_{\mathrm{order}}$ moves from stronger counterfactual weighting ($0.5$) toward confidence-only commitment ($1$). \cmark{} indicates that both sequences are non-increasing across the four tested weights.}
\label{tab:main-generalization}
\centering
\scriptsize
\setlength{\tabcolsep}{3.5pt}
\resizebox{\linewidth}{!}{%
\begin{tabular}{llccc}
\toprule
Model / target & Activation &
\makecell{Target\\$\lambda_{\mathrm{order}}=0.5 \rightarrow 0.75 \rightarrow 0.9 \rightarrow 1$} &
\makecell{PPL\\$\lambda_{\mathrm{order}}=0.5 \rightarrow 0.75 \rightarrow 0.9 \rightarrow 1$} &
\makecell{Ordering\\consistent} \\
\midrule
LLaDA / Subjectivity & None
    & $0.64 \rightarrow 0.57 \rightarrow 0.54 \rightarrow 0.49$
    & $12.5 \rightarrow 10.7 \rightarrow 9.4 \rightarrow 8.9$
    & \cmark \\
LLaDA / Subjectivity & DiffMean
    & $0.75 \rightarrow 0.70 \rightarrow 0.63 \rightarrow 0.57$
    & $14.2 \rightarrow 11.0 \rightarrow 10.3 \rightarrow 8.6$
    & \cmark \\
LLaDA / Subjectivity & ACT
    & $0.70 \rightarrow 0.60 \rightarrow 0.54 \rightarrow 0.47$
    & $12.2 \rightarrow 9.8 \rightarrow 9.1 \rightarrow 8.6$
    & \cmark \\
LLaDA-V / Caption sentiment & None
    & $0.84 \rightarrow 0.47 \rightarrow 0.33 \rightarrow 0.32$
    & $13.5 \rightarrow 12.3 \rightarrow 12.0 \rightarrow 11.8$
    & \cmark \\
LLaDA-V / Caption sentiment & DiffMean
    & $0.89 \rightarrow 0.47 \rightarrow 0.38 \rightarrow 0.33$
    & $15.0 \rightarrow 12.1 \rightarrow 11.6 \rightarrow 11.3$
    & \cmark \\
LLaDA-V / Caption sentiment & ACT
    & $0.57 \rightarrow 0.49 \rightarrow 0.45 \rightarrow 0.42$
    & $11.5 \rightarrow 11.1 \rightarrow 11.1 \rightarrow 11.0$
    & \cmark \\
\midrule
Dream-7B / Sentiment & None
    & $0.71 \rightarrow 0.52 \rightarrow 0.41 \rightarrow 0.30$
    & $8.0 \rightarrow 7.8 \rightarrow 7.4 \rightarrow 7.0$
    & \cmark \\
Dream-7B / Sentiment & DiffMean
    & $0.80 \rightarrow 0.64 \rightarrow 0.54 \rightarrow 0.44$
    & $7.8 \rightarrow 7.0 \rightarrow 6.7 \rightarrow 6.4$
    & \cmark \\
Dream-7B / Sentiment & ACT
    & $0.76 \rightarrow 0.65 \rightarrow 0.50 \rightarrow 0.42$
    & $8.4 \rightarrow 7.7 \rightarrow 7.6 \rightarrow 6.9$
    & \cmark \\
\bottomrule
\end{tabular}}
\end{table}

The same ordering holds on Dream-7B, where both sequences are non-increasing under all three activation conditions (Table~\ref{tab:main-generalization}). Dream in fact traverses a wider target range than LLaDA in the unsteered condition, moving from $0.30$ at confidence-only commitment to $0.71$ at $\lambda_{\mathrm{order}}=0.5$, at a perplexity cost of about one point. Resampling the evaluation set with replacement, the target ordering for the three LLaDA sentiment sweeps is reproduced in $88.1$--$91.4\%$ of resamples (Appendix~\ref{app:order-preservation}).

Because counterfactual ordering spends additional scorer computation, we also
ask what the same budget buys if it is spent on repeated sampling instead. At
the expected number of generations that matches the measured counterfactual
overhead, best-of-$N$ selection under the same target scorer reaches a lower
target score at a higher perplexity than the $\lambda_{\mathrm{order}}=0.75$
commitment point in three of four settings, and matches it at $1.1$ lower
perplexity in the fourth (Appendix~\ref{app:best-of-n}).

We ran two additional controls. Random commitment order costs substantially more perplexity than the counterfactual sweep for far less target movement, so the effect does not follow from merely departing from confidence ranking (Appendix~\ref{app:counterfactual-plots}); and re-scoring the same generations with an independent sentiment classifier, on both model families, preserves the ordering in every sweep (Appendix~\ref{app:heldout-scorer}).

\subsection{Adaptive budgets are empirically monotonic and competitive at matched average cost}
\label{sec:results-budget}

\begin{figure}[t]
    \centering
    \includegraphics[width=0.72\linewidth]{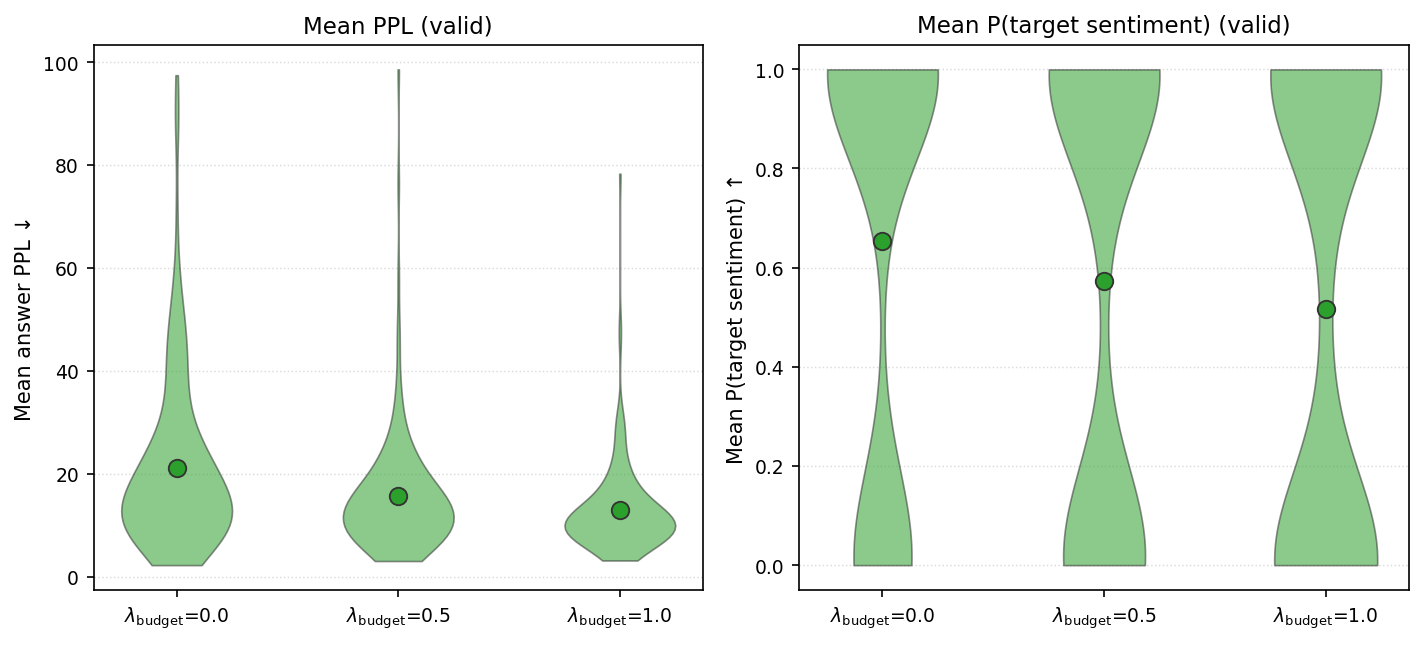}
    \caption{Adaptive commitment budgets for LLaDA sentiment with ACT. Varying $\lambda_{\mathrm{budget}}$ with confidence ranking fixed yields an empirically ordered target--quality sweep.}
    \label{fig:adaptive-main}
\end{figure}

With ranking fixed to confidence, target-progress-driven budgeting ($\lambda_{\mathrm{budget}}=0$) consistently gives the strongest target scores, confidence-driven budgeting ($\lambda_{\mathrm{budget}}=1$) the lowest PPL, and the mixed policy lies between (Fig.~\ref{fig:adaptive-main}). This ordering also holds in all eleven additional control configurations, including the three Dream-7B conditions (Appendices~\ref{app:adaptive-tables} and~\ref{app:dream}). Commitment budget therefore forms a second interpretable axis: $\lambda_{\mathrm{order}}$ controls which predictions survive, $\lambda_{\mathrm{budget}}$ how quickly they become permanent.

\begin{figure}[t]
    \centering
    \includegraphics[width=0.72\linewidth]{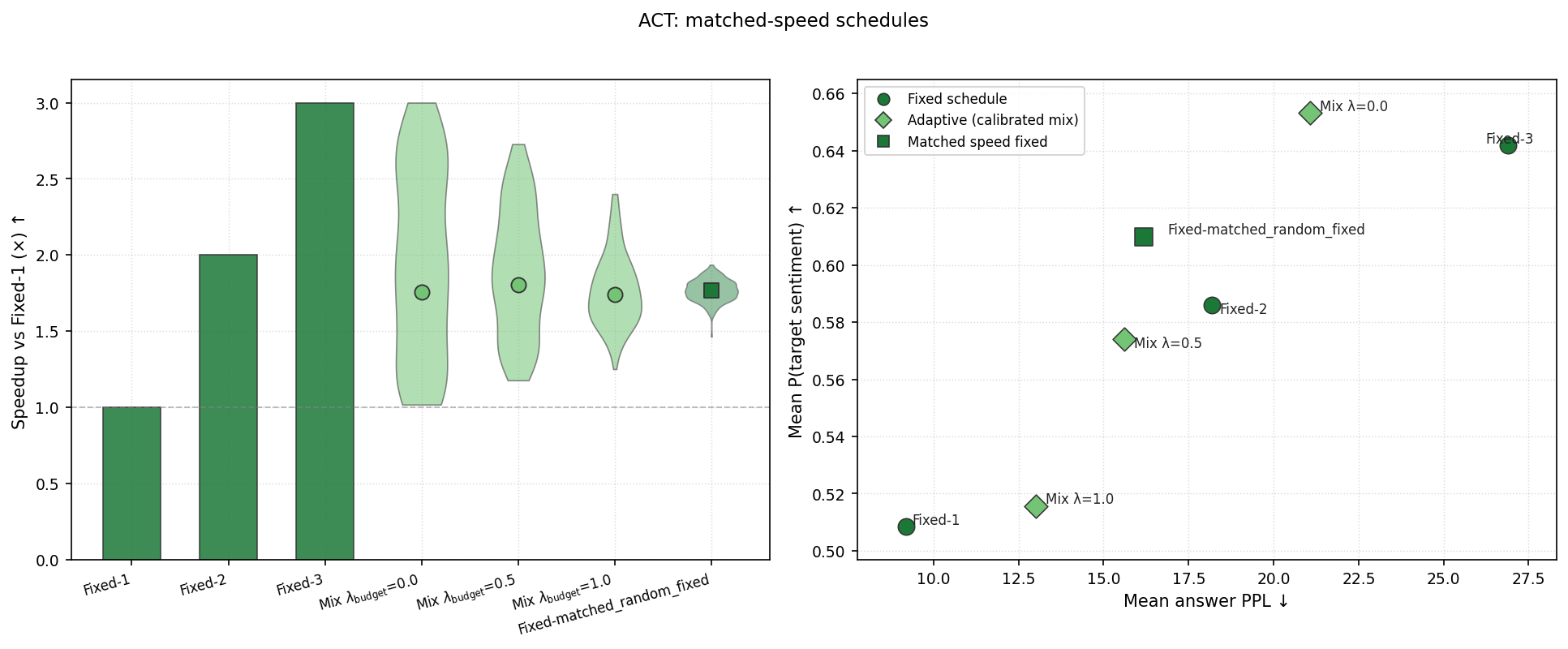}
    \caption{Adaptive budgets versus a scorer-free stochastic fixed schedule matched in average denoising steps, for LLaDA sentiment with ACT. Fixed-$1/2/3$ are shown as contextual reference points. This matched comparison isolates state-dependent adaptation from average denoising cost.}
    \label{fig:adaptive-matched}
\end{figure}

Fixed-$1$, fixed-$2$, and fixed-$3$ expose the coarse regimes produced by constant budgets, but do not separate adaptation from average commitment rate. The compute-matched stochastic schedule provides the decisive comparison: same confidence ranking, essentially the same average denoising-step count, but budgets sampled from a state-independent distribution and no access to the target scorer. Any difference in target--quality position is therefore attributable to conditioning the budget on the current denoising state rather than to average compute (Fig.~\ref{fig:adaptive-matched}).

Table~\ref{tab:matched-stochastic-main} reports adaptive minus matched-stochastic differences. Across all fifteen settings the adaptive point is competitive at matched average denoising cost: it is never jointly dominated in target and PPL. The comparison is intentionally not framed as universal strict dominance. Instead, it shows that state-dependent budgeting reaches a different and generally favorable target--quality operating point even after controlling for mean denoising computation.

\begin{table}[t]
\caption{Adaptive schedules versus compute-matched stochastic fixed schedules, for four representative settings; all fifteen are in Appendices~\ref{app:matched-tables} and~\ref{app:dream}. Deltas are adaptive minus matched stochastic, so positive $\Delta$Target favors adaptation and negative $\Delta$PPL indicates better fluency. ``Competitive'' means the adaptive point is not jointly dominated in target and PPL at matched average denoising cost. The Dream-7B $\lambda_{\mathrm{budget}}=0$ row improves both metrics simultaneously.}
\label{tab:matched-stochastic-main}
\centering
\scriptsize
\setlength{\tabcolsep}{5.0pt}
\begin{tabular}{llcccc}
\toprule
Model / target & Activation &
$\lambda_{\mathrm{budget}}$ &
$\Delta$Target &
$\Delta$PPL &
Competitive \\
\midrule
LLaDA / Sentiment & DiffMean & $0$   & $+0.07$ & $+4.2$ & \cmark \\
                   &          & $0.5$ & $-0.03$ & $-3.0$ & \cmark \\
                   &          & $1$   & $-0.07$ & $-4.2$ & \cmark \\
\midrule
LLaDA-V / Caption sentiment & None & $0$   & $+0.24$ & $+2.5$ & \cmark \\
                   &          & $0.5$ & $-0.01$ & $-1.1$ & \cmark \\
                   &          & $1$   & $-0.15$ & $-2.9$ & \cmark \\
\midrule
LLaDA / NLI entailment & None & $0$   & $+0.03$ & $+2.5$ & \cmark \\
                   &          & $0.5$ & $-0.02$ & $-1.8$ & \cmark \\
                   &          & $1$   & $-0.06$ & $-4.1$ & \cmark \\
\midrule
Dream-7B / Sentiment & DiffMean & $0$   & $+0.03$ & $-0.8$ & \cmark \\
                   &          & $0.5$ & $-0.02$ & $-2.5$ & \cmark \\
                   &          & $1$   & $-0.03$ & $-3.7$ & \cmark \\
\bottomrule
\end{tabular}
\end{table}

In some settings the comparison is stronger still. Beyond not being dominated by its matched control, an adaptive point can also fall outside the reach of the constant-budget family: no interpolation between fixed-$1$, fixed-$2$, and fixed-$3$ attains its target score at its perplexity. This holds for target-driven budgeting ($\lambda_{\mathrm{budget}}=0$) in the LLaDA-V caption-sentiment settings and in LLaDA sentiment with ACT. On Dream-7B sentiment, two of the nine matched comparisons are strict improvements on both metrics at once (Appendix~\ref{app:dream}).

Because all reported entries are means over a finite evaluation set, we also ask how stable the observed orderings are. Resampling the LLaDA sentiment evaluation set with replacement, the target ordering is preserved in $86.7$--$95.3\%$ of resamples across the six sweeps, and the adaptive point remains non-dominated by its matched schedule in at least $91.1\%$ of resamples in every configuration (Appendix~\ref{app:order-preservation}).

\subsection{Commitment extends to task-success targets}
\label{sec:results-task}

\begin{figure}[t]
    \centering
    \begin{minipage}{0.49\linewidth}
        \centering
        \includegraphics[width=\linewidth]{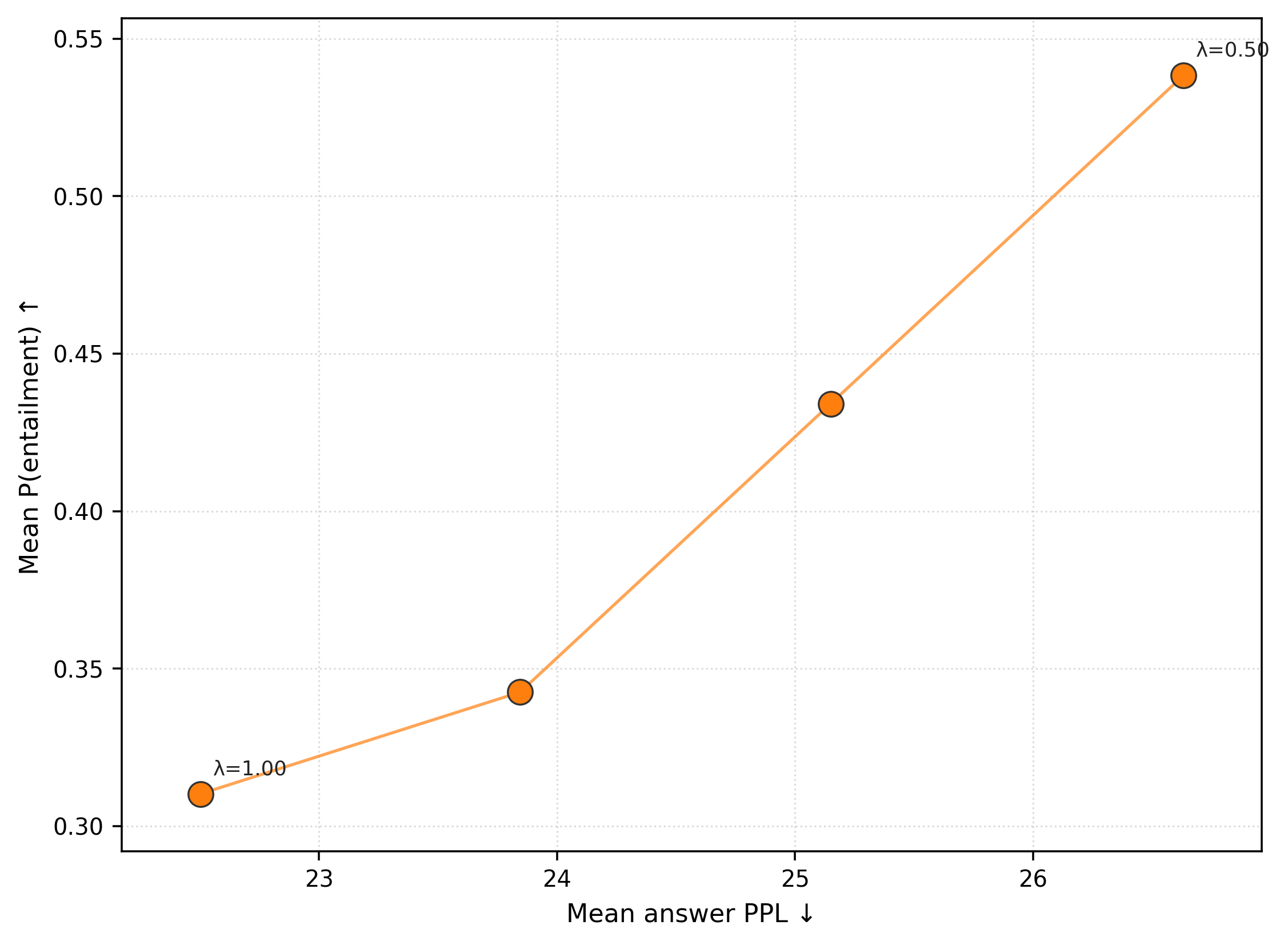}
    \end{minipage}\hfill
    \begin{minipage}{0.49\linewidth}
        \centering
        \includegraphics[width=\linewidth]{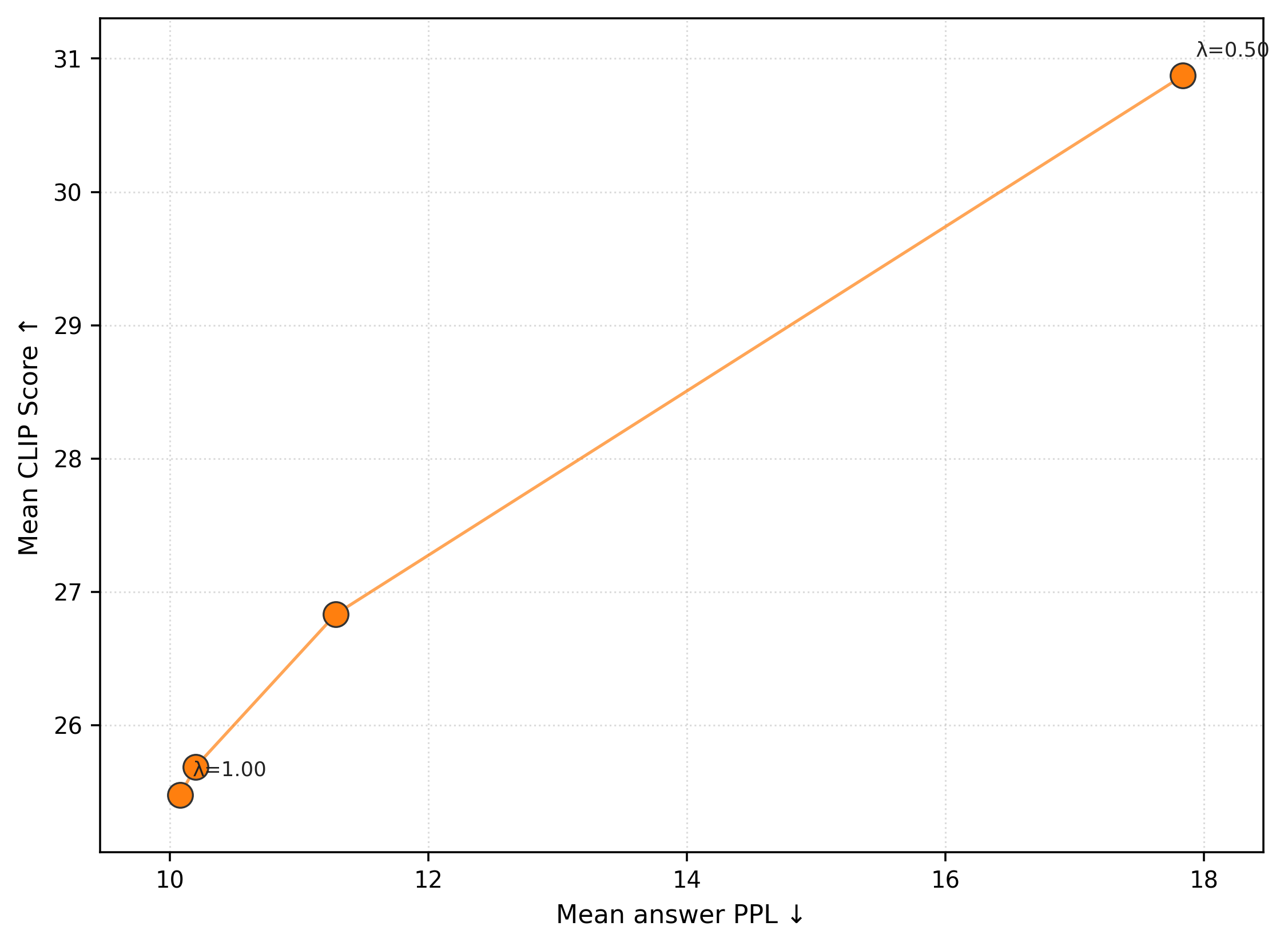}
    \end{minipage}
    \caption{Counterfactual commitment order for task-success objectives. \textbf{Left:} mean NLI entailment probability versus PPL on SAMSum. \textbf{Right:} mean CLIPScore versus PPL on SentiCap, where CLIPScore is nonnegative cosine similarity scaled by $100$. Both use $u_t\equiv1$ and no activation steering.}
    \label{fig:task-counterfactual}
\end{figure}

Sentiment and subjectivity are output attributes. We now ask whether the same controls apply when the target measures task success: whether a summary is entailed by its source dialogue, and whether a caption matches its image. Unlike an attribute, these targets are example-specific and correspond to no single reusable output direction, so we evaluate them without activation steering. Both objectives are frozen automatic proxies for task quality rather than task quality itself, so the claim in this section is that commitment control moves these scores in an ordered and consistent way, not that it improves summarization or captioning in full. For both objectives we additionally verify that the ordering survives re-scoring with an independent model: an entailment model on both families for NLI, and a CLIP model pretrained on a different corpus for CLIPScore (Appendix~\ref{app:heldout-scorer}).

Counterfactual ordering remains empirically monotonic in both. For NLI, moving from
confidence-only commitment to the strongest tested counterfactual mixture raises mean
entailment probability from approximately $0.31$ to $0.54$, while PPL increases from
approximately $22.5$ to $26.6$ (Fig.~\ref{fig:task-counterfactual}); the same sweep on
Dream-7B moves entailment from $0.34$ to $0.60$ at $22.4$ to $27.3$ PPL
(Appendix~\ref{app:dream-nli}). Since $u_t$ remains fixed, the gain comes entirely from
changing which provisional summary tokens are committed first. CLIPScore exhibits the
same ordered behavior in LLaDA-V captioning.

Adaptive budgeting behaves the same way---target-driven strongest on the objective,
confidence-driven lowest on PPL, mixed in between---and all these settings remain
competitive against their compute-matched stochastic schedules
(Table~\ref{tab:matched-stochastic-main}). Full sweeps are in
Appendices~\ref{app:full-tables} and~\ref{app:dream}.
\section{Discussion, Limitations, and Conclusion}

Token commitment is complementary to interventions that alter model predictions. Across the tested conditions, activation steering changes the base target--quality operating point, while commitment order and budget provide additional movement around it. The two controls have distinct roles: order determines which provisional predictions survive at fixed denoising cost, whereas budget changes the rate at which predictions become irreversible. Increasing the activation-steering strength $\alpha$ or $\beta$ also trades target against fluency, so it is natural to ask how commitment control compares to simply steering harder. A fair answer would require optimizing each mechanism independently---sweeping intervention strength as well as commitment weight---and characterizing the resulting two-dimensional operating region; we hold each intervention at one strength here and leave that joint characterization to future work. The Dream-7B experiments do use substantially different strengths ($\alpha=0.7$, $\beta=0.4$) from the LLaDA ones ($\alpha=3.0$, $\beta=0.35$) and reproduce the same commitment behavior, so the effect is not tied to one intervention magnitude; because model and strength vary together there, this does not substitute for a controlled strength sweep.

Our evidence spans two modalities and two model families, and the second family covers both a control target and a task-success target. It does not cover subjectivity or the vision-language settings, which are evaluated on LLaDA and LLaDA-V only. All objectives are defined by frozen external scorers. We verify that the reported orderings are not artifacts of the particular decoding scorer by re-scoring the same generations with independent models: for sentiment and NLI entailment on both model families, and for CLIPScore on LLaDA-V (Appendix~\ref{app:heldout-scorer}). Counterfactual commitment also has materially greater scorer-side overhead: in a representative ACT sentiment benchmark, adaptive budgeting adds $8.5\%$ wall-clock latency relative to a scorer-free schedule at essentially identical average denoising steps, whereas counterfactual ordering adds $48.9\%$ relative to confidence ranking at the same $60$ denoising steps (Appendix~\ref{app:runtime-batching}). That overhead is the relevant quantity to compare against alternative uses of the same budget: at a matched expected number of generations, best-of-$N$ selection under the same target scorer reaches a lower target score at a higher perplexity than the $\lambda_{\mathrm{order}}=0.75$ commitment point in three of the four settings we tested, and the same target score at lower perplexity in the fourth (Appendix~\ref{app:best-of-n}). A promising direction is to amortize the counterfactual computation with a lightweight surrogate that predicts token contributions from the current denoising state while preserving their ranking.

Returning to the question posed in Section~\ref{sec:problem}, our findings are that
(1) with the budget map held fixed, replacing the scoring map with a target-aware
mixture moves generation along an empirically monotonic target--quality trade-off at
identical denoising cost in every setting we evaluate, and neither random ordering nor
scorer-free resampling at a matched generation budget reproduces that movement;
(2) with the scoring map held fixed to confidence, replacing the budget map exposes a
second empirically monotonic trade-off, non-increasing in both target and perplexity in
every setting we evaluate, and the resulting adaptive point is never jointly dominated by
a scorer-free schedule matched in average denoising cost in any of the fifteen settings,
with some target-driven points lying outside the reach of any constant-budget schedule;
and
(3) both behaviors extend from output-attribute targets to example-specific proxies for
task success, with the ordering preserved under independent scorers for NLI entailment on
both model families and for CLIPScore on LLaDA-V, even though these objectives remain
proxies.

Overall, masked diffusion decoding is not merely a downstream implementation detail. Concretely, on LLaDA sentiment with no activation intervention, varying only which provisional token is committed moves the target score from $0.35$ to $0.63$ at an identical $60$ denoising steps; on Dream-7B the same knob moves it from $0.30$ to $0.71$, and it moves SAMSum entailment from $0.31$ to $0.54$ on LLaDA and from $0.34$ to $0.60$ on Dream. Both knobs are empirically monotonic across every setting we evaluate, so each exposes an interpretable operating curve rather than a single intervention point, and the reported orderings survive resampling of the evaluation set in the large majority of cases (Appendix~\ref{app:order-preservation}) as well as substitution of the target scorer (Appendix~\ref{app:heldout-scorer}). By deciding which provisional predictions become permanent and how rapidly commitment proceeds, the sampler exposes a diffusion-specific control surface.

\subsubsection*{Broader Impact Statement}
This work enables inference-time control of masked-diffusion generation without retraining, allowing users to select trade-offs among target objectives, output quality, and compute. The same mechanism could steer content toward harmful objectives or amplify biases and failures in imperfect automatic scorers. Consequential applications should therefore audit target scorers and retain human oversight.

\bibliography{references}
\bibliographystyle{tmlr}

\appendix
\raggedbottom

\section{Full Problem Definitions and Algorithmic Details}
\label{app:formal-details}

\subsection{Target scores, quality, and cost}

Let $S(y;\xi)$ denote a generic target score, where $y$ is the decoded output and $\xi$ is any input-specific scoring context. For sentiment and subjectivity we use the target-class probability of frozen binary classifiers. For NLI,
\[
S_{\mathrm{NLI}}(y,c)
=
P_{\phi_{\mathrm{NLI}}}(\mathrm{entailment}\mid c,y).
\]
Here $c$ is the source dialogue, $y$ is its generated summary, and $\phi_{\mathrm{NLI}}$ denotes the frozen NLI scorer.
For image--text alignment, we use CLIPScore computed with
\texttt{openai/clip-vit-base-patch32}. With normalized image and text embeddings,
\[
S_{\mathrm{CLIP}}(y,I)
=
\max\!\left(
100\,
\frac{f_{\mathrm{text}}(y)^\top f_{\mathrm{image}}(I)}
{\|f_{\mathrm{text}}(y)\|_2\|f_{\mathrm{image}}(I)\|_2},
\,0
\right).
\]
Here $I$ is the input image, while $f_{\mathrm{text}}$ and $f_{\mathrm{image}}$ are the frozen CLIP text and image encoders. Thus, the reported CLIPScore is a nonnegative cosine similarity scaled by $100$.
The methods evaluate these same functions on provisional completions $\tilde x_t$ during denoising.

Let $p_{\mathrm{ref}}$ be an external autoregressive language model. We compute
\[
\operatorname{PPL}_{\mathrm{ref}}(y)
=
\exp\!\left(
-\frac1{L_y}\sum_{\ell=1}^{L_y}
\log p_{\mathrm{ref}}(y_\ell\mid y_{<\ell})
\right).
\]
Here $y_\ell$ is the $\ell$th token of $y$ and $L_y$ is the number of evaluated output tokens. Perplexity is computed on the generated continuation, summary, or caption only. We additionally flag outputs with $\operatorname{PPL}_{\mathrm{ref}}(y)>\tau_{\mathrm{PPL}}$, where $\tau_{\mathrm{PPL}}>0$ is a fixed validity threshold, for the diagnostics in Appendix~\ref{app:invalid-diagnostics}. The principal denoising-cost metric is the number of denoising steps $D(y)$ used to produce $y$, averaged over examples. Unless stated otherwise, reported target and perplexity means are computed over valid generations.

\subsection{Counterfactual commitment details}
\label{app:counterfactual-details}

For each provisional token $i\in M_t$, we identify the character span $[a_{t,i},b_{t,i})$ induced by that token in the decoded provisional completion. We form
\[
A_i(\tilde x_t)
=
\operatorname{clean}\!\left(
\operatorname{text}(\tilde x_t)_{<a_{t,i}}
\Vert
\operatorname{text}(\tilde x_t)_{\ge b_{t,i}}
\right),
\]
where $\operatorname{text}$ decodes a token sequence, $\Vert$ concatenates text spans, and $\operatorname{clean}$ normalizes whitespace and substitutes a minimal nonempty string if deletion produces an empty text. For binary scorers, $\Delta_{t,i}$ is computed from the raw target-class logit to reduce softmax saturation; multi-class scorers such as NLI use the target probability directly.

We normalize within each step:
\[
\bar\Delta_{t,i}
=
\begin{cases}
\dfrac{\Delta_{t,i}-\Delta_t^{\min}}
{\Delta_t^{\max}-\Delta_t^{\min}},
& \Delta_t^{\max}>\Delta_t^{\min},\\[0.8em]
0,&\text{otherwise},
\end{cases}
\]
where $\Delta_t^{\min}=\min_{j\in M_t}\Delta_{t,j}$ and $\Delta_t^{\max}=\max_{j\in M_t}\Delta_{t,j}$.
When all contributions are equal, or only one unresolved position remains, this value does not affect the committed set. Confidence is retained on its native probability scale, while counterfactual contributions are min--max normalized within each denoising step so that the target signal is comparable across denoising states; $\lambda_{\mathrm{order}}$ therefore selects an operating point along this interpolation.

\begin{algorithm}[h]
\caption{Counterfactual target-aware commitment at step $t$}
\let\AND\relax
\begin{algorithmic}[1]
\REQUIRE State $x_t$, unresolved positions $M_t$, provisional predictions $\{\hat x_{t,i}\}$, confidences $\{q_{t,i}\}$, scorer $S$, budget $u_t$, weight $\lambda_{\mathrm{order}}$
\STATE Form $\tilde x_t$ by filling every $i\in M_t$ with $\hat x_{t,i}$
\STATE Compute $S(\tilde x_t;\xi)$
\FOR{$i\in M_t$}
    \STATE Construct $A_i(\tilde x_t)$
    \STATE $\Delta_{t,i}\leftarrow S(\tilde x_t;\xi)-S(A_i(\tilde x_t);\xi)$
\ENDFOR
\STATE Normalize $\{\Delta_{t,i}\}$ to $\{\bar\Delta_{t,i}\}$
\STATE $r_{t,i}^{\mathrm{cf}}\leftarrow
\lambda_{\mathrm{order}}q_{t,i}
+(1-\lambda_{\mathrm{order}})\bar\Delta_{t,i}$
\STATE $C_t\leftarrow\operatorname{Top}_{u_t}\{r_{t,i}^{\mathrm{cf}}:i\in M_t\}$
\RETURN $C_t$
\end{algorithmic}
\end{algorithm}

Before batching, the method produces $|M_t|+1$ scorer inputs at step $t$: the full provisional completion and one deletion variant per unresolved position.

\subsection{Adaptive-budget calibration and algorithm}
\label{app:adaptive-details}

We define the top-$U_{\max}$ confidence signal
\[
h_t=\frac1{|\mathcal K_t|}\sum_{i\in\mathcal K_t}q_{t,i},
\qquad
\mathcal K_t=\operatorname{Top}_{\min(U_{\max},|M_t|)}
\{q_{t,i}:i\in M_t\}.
\]
Here $\mathcal K_t\subseteq M_t$ contains the highest-confidence unresolved positions available at step $t$.
Calibration uses a fixed-$2$ run on a held-out prompt set drawn from the same corpus as the evaluation prompts but disjoint from them: $500$ prompts for the text settings and $125$ for the vision-language settings. No evaluation prompt contributes to any empirical CDF. Fixed-$2$ is the central member of the evaluated family $\{1,2,3\}$ and supplies an intermediate denoising regime around which adaptation may choose smaller or larger budgets. The same calibration procedure is used for every adaptive configuration and is not tuned separately by target. Because the calibration is a per-model empirical quantile transform, it is re-derived rather than transferred when the denoiser changes; no code path or hyperparameter is model-specific.

Let $\pi_t=1-|M_t|/L_{\mathrm{block}}\in[0,1]$ denote denoising progress, where $L_{\mathrm{block}}$ is the number of positions in the generated block, equal to the output length $L$ in all our configurations. We divide denoising into early, middle, and late thirds by $\pi_t$ and construct phase-specific empirical CDFs $F_S^\rho$ and $F_H^\rho$ for target progress and confidence, indexed by phase $\rho\in\{\text{early},\text{middle},\text{late}\}$. At test time,
\[
\hat s_t=F_S^{\rho_t}(s_t),
\qquad
\hat h_t=F_H^{\rho_t}(h_t).
\]
Here $\rho_t$ is the phase containing progress $\pi_t$, while $F_S^{\rho_t}$ and $F_H^{\rho_t}$ are the target-progress and confidence CDFs estimated for that phase.

\begin{algorithm}[h]
\caption{Adaptive commitment budget at step $t$}
\let\AND\relax
\begin{algorithmic}[1]
\REQUIRE State $x_t$, unresolved positions $M_t$, provisional predictions, confidences, scorer $S$, CDFs $F_S^\rho,F_H^\rho$, $U_{\max}$, weight $\lambda_{\mathrm{budget}}$
\STATE Form provisional completion $\tilde x_t$
\STATE $s_t\leftarrow S(\tilde x_t;\xi)$
\STATE Compute $h_t$ from the top-$U_{\max}$ confidences
\STATE Determine phase $\rho_t$ and calibrate $\hat s_t,\hat h_t$
\STATE $z_t\leftarrow(1-\lambda_{\mathrm{budget}})\hat s_t+\lambda_{\mathrm{budget}}\hat h_t$
\STATE $u_t\leftarrow\min(1+\lfloor U_{\max}z_t\rfloor,U_{\max},|M_t|)$
\STATE $C_t\leftarrow\operatorname{Top}_{u_t}\{q_{t,i}:i\in M_t\}$
\RETURN $C_t$
\end{algorithmic}
\end{algorithm}

Adaptive budgeting evaluates one full provisional completion per denoising step and does not require per-token deletion scores.

\subsection{Compute-matched stochastic fixed schedule}
\label{app:matched-schedule-details}

For each (target, activation) setting we construct a \emph{single} matched schedule. The three adaptive configurations $\lambda_{\mathrm{budget}}\in\{0,0.5,1\}$ have very similar mean denoising-step counts, so one scorer-free control suffices for all of them. Let $D^\star$ be the mean denoising-step count over those three configurations. The matched schedule samples between neighboring integer budgets $k$ and $k+1$:
\[
u_t\sim
\begin{cases}
k+1,&\text{with probability }\gamma,\\
k,&\text{with probability }1-\gamma.
\end{cases}
\]
Here $k\in\mathbb N$ and $k+1$ are the neighboring budgets, and $\gamma\in[0,1]$ is a fixed sampling probability.
The pair $(k,k+1)$ brackets the average commitment rate associated with $D^\star$, and a fixed $\gamma$ is chosen so that the schedule's realized mean denoising-step count approximates $D^\star$. Both are selected \emph{post hoc}, from denoising-step counts alone, and never from target scores or perplexity. Once selected, $\gamma$ and the RNG seed are held fixed throughout evaluation. The sampled budget is independent of the target score, calibration statistics, confidence values, phase, prompt, and denoising progress. After sampling $u_t$, the policy commits the top-$u_t$ unresolved positions under confidence ranking. All three adaptive points are therefore compared against the same scorer-free control at essentially the same average denoising cost.

Run metadata records the selected neighboring budgets, the fixed probability, the schedule seed, the target and realized average steps, the absolute matching error, and the per-generation budget sequence.

\subsection{Activation-intervention definitions}
\label{app:activation-definitions}

DiffMean uses
\[
v_\ell=\mu_\ell^{\mathrm{tgt}}-\mu_\ell^{\mathrm{src}},
\qquad
h_\ell\leftarrow h_\ell+\alpha v_\ell.
\]
Here $\mu_\ell^{\mathrm{src}}$ and $\mu_\ell^{\mathrm{tgt}}$ are source- and target-pole mean activations at layer $\ell$, $v_\ell$ is their difference, $h_\ell$ is the current hidden state, and $\alpha\in\mathbb R$ is the intervention strength.
Diagonal ACT uses
\[
T_\ell(h)
=
\mu_\ell^{\mathrm{tgt}}
+
\frac{\sigma_\ell^{\mathrm{tgt}}}
{\sigma_\ell^{\mathrm{src}}+\epsilon}
\odot(h-\mu_\ell^{\mathrm{src}}),
\qquad
h_\ell\leftarrow h_\ell+\beta(T_\ell(h_\ell)-h_\ell).
\]
Here $\sigma_\ell^{\mathrm{src}}$ and $\sigma_\ell^{\mathrm{tgt}}$ are elementwise standard deviations, $\epsilon>0$ prevents division by zero, $\odot$ denotes elementwise multiplication, $T_\ell$ is the resulting diagonal affine transport, and $\beta\in\mathbb R$ is its strength.
Positive strengths are oriented toward the stated target. We use $\ell\in\{9,\ldots,24\}$, spanning the mid-to-late layers where steering interventions in masked diffusion language models have been reported to have the greatest effect \citep{shnaidman2025activation}. The window is fixed once and reused for every target, activation method, and commitment configuration; we do not search layers per setting, and DiffMean and ACT are therefore representative steered operating points rather than estimates of either method's best achievable performance. For Dream-7B the same fractional-depth window is used rather than the same absolute layer indices (Appendix~\ref{app:dream-setup}).
\clearpage
\section{Additional Experimental Details}
\label{app:experimental_details}

This appendix contains reproduction-level details omitted from the main experimental setup.

\Needspace{0.32\textheight}
\subsection{Prompt templates}

We use task-specific prompt templates for product-review continuation, SAMSum summarization, and image captioning. The exact templates are listed in Table~\ref{tab:prompt_templates}. In all cases, prompts instruct the model not to add headings, labels, or commentary. The Dream-7B experiments reuse the corresponding templates unchanged.

\begin{table}[H]
\caption{Prompt templates used in the experiments.}
\label{tab:prompt_templates}
\centering
\small
\begin{tabular}{p{0.22\linewidth} p{0.70\linewidth}}
\toprule
Task & Prompt template \\
\midrule
Product-review continuation &
\texttt{Continue the following product review naturally. Keep the same topic as the original text. Write around 10 complete and long sentences. Do not add headings, labels, or commentary. Review: \{prefix\}} \\
\midrule
SAMSum summarization &
\texttt{Summarize the following conversation in one concise sentence. Do not add headings, labels, or commentary. Conversation: \{dialogue\} Summary:} \\
\midrule
LLaDA-V captioning &
\texttt{Describe the image.} \\
\bottomrule
\end{tabular}
\end{table}

\Needspace{0.56\textheight}
\subsection{Generation, scoring, and commitment hyperparameters}

\begin{table}[H]
\caption{Generation, scoring, and commitment hyperparameters shared across experiments. The held-out scorers are used only for the post-hoc checks in Appendix~\ref{app:heldout-scorer} and never during decoding. Activation-steering strengths are reported separately in Table~\ref{tab:steering_strengths}.}
\label{tab:generation_hyperparams}
\centering
\small
\begin{tabular}{p{0.34\linewidth} p{0.55\linewidth}}
\toprule
Hyperparameter & Value \\
\midrule
LLaDA model & \texttt{GSAI-ML/LLaDA-8B-Instruct} \\
LLaDA-V model & \texttt{GSAI-ML/LLaDA-V} \\
Dream model & \texttt{Dream-org/Dream-v0-Instruct-7B} \\
LLaDA generation length & $60$ \\
LLaDA block length & $60$ \\
LLaDA-V generation length & $35$ \\
Dream generation length & $60$ \\
Temperature & $0$ \\
Random seed & $42$ \\
Termination-token policy & banned in first $50$ generated positions; allowed in final $10$ positions \\
Main steering layers (LLaDA) & $\{9,\ldots,24\}$ of $32$ \\
Main steering layers (Dream) & $\{8,\ldots,21\}$ of $28$ \\
Counterfactual commitment budget & $u_t \equiv 1$ \\
Adaptive maximum budget & $U_{\max}=3$ \\
Fixed-budget baselines & $u_t \equiv 1,2,3$ \\
Counterfactual sweep & $\lambda_{\mathrm{order}}\in\{0.5,0.75,0.9,1.0\}$ \\
Adaptive sweep & $\lambda_{\mathrm{budget}}\in\{0,0.5,1\}$ \\
Adaptive confidence signal & mean of top-$3$ candidate confidences \\
Calibration schedule & fixed-$2$ \\
Calibration prompts & $500$ (text) / $125$ (vision-language), disjoint from evaluation prompts \\
Calibration phases & early/middle/late by fraction of generated block unmasked \\
Sentiment scorer checkpoint & \texttt{distilbert/distilbert-base-} \newline \texttt{uncased-finetuned-sst-2-english} \\
Held-out sentiment scorer & \texttt{cardiffnlp/twitter-roberta-base-sentiment-latest} \\
Subjectivity scorer checkpoint & \texttt{GroNLP/mdebertav3-subjectivity-english} \\
NLI scorer checkpoint & \texttt{cross-encoder/nli-MiniLM2-L6-H768} \\
Held-out NLI scorer & \texttt{cross-encoder/nli-deberta-v3-large} \\
CLIP scorer checkpoint & \texttt{openai/clip-vit-base-patch32} \\
Held-out CLIP scorer & \texttt{laion/CLIP-ViT-B-32-} \newline \texttt{laion2B-s34B-b79K} \\
NLI scorer max length & $512$ tokens, with padding and truncation \\
Reference perplexity model & Qwen2.5-3B \\
Perplexity invalidity threshold & $\tau_{\mathrm{PPL}}=100$ \\
\bottomrule
\end{tabular}
\end{table}

\Needspace{0.32\textheight}
\subsection{Activation-steering strengths}

\begin{table}[H]
\caption{Activation-steering strengths used in the experiments. Each is held at a single value throughout, so that the commitment weight is the only variable along a sweep. By convention, positive $\alpha$ or $\beta$ always steers toward the paper's stated target direction: negative sentiment, subjectivity, or negative caption sentiment. DiffMean applies $h_\ell \leftarrow h_\ell+\alpha v_\ell$, while ACT applies $h_\ell \leftarrow h_\ell+\beta(T_\ell(h_\ell)-h_\ell)$.}
\label{tab:steering_strengths}
\centering
\small
\begin{tabular}{p{0.22\linewidth} p{0.22\linewidth} p{0.28\linewidth} p{0.16\linewidth}}
\toprule
Model & Method & Target & Strength \\
\midrule
LLaDA & DiffMean & Sentiment & $\alpha=3.0$ \\
LLaDA & DiffMean & Subjectivity & $\alpha=2.0$ \\
LLaDA & ACT & Sentiment & $\beta=0.35$ \\
LLaDA & ACT & Subjectivity & $\beta=0.25$ \\
LLaDA-V & DiffMean & Caption sentiment & $\alpha=3.0$ \\
LLaDA-V & ACT & Caption sentiment & $\beta=0.5$ \\
\midrule
Dream-7B & DiffMean & Sentiment & $\alpha=0.7$ \\
Dream-7B & ACT & Sentiment & $\beta=0.4$ \\
\bottomrule
\end{tabular}
\end{table}

\Needspace{0.62\textheight}
\subsection{Vector and statistics construction}

\begin{table}[H]
\caption{Activation-vector and ACT-statistics construction. These hyperparameters determine what each activation intervention steers toward.}
\label{tab:vector_construction}
\centering
\small
\begin{tabular}{p{0.21\linewidth} p{0.21\linewidth} p{0.18\linewidth} p{0.30\linewidth}}
\toprule
Setting & Data source & Pole sizes & Construction details \\
\midrule
LLaDA DiffMean sentiment &
Amazon Polarity train &
$5000$ positive, $5000$ negative &
Mean-difference vector oriented toward the negative-sentiment pole, $v_\ell=\mu_\ell^{\mathrm{neg}}-\mu_\ell^{\mathrm{pos}}$; text truncated to $4000$ characters; tokenized to max length $256$ \\
\midrule
LLaDA DiffMean subjectivity &
Amazon Polarity train, scored by the subjectivity classifier &
top $5000$ objective, bottom $5000$ objective &
Same DiffMean construction as sentiment, using subjective and objective poles; vector is oriented toward the subjective target; text truncated to $4000$ characters; tokenized to max length $256$ \\
\midrule
LLaDA ACT sentiment &
Amazon Polarity train &
$5000$ positive, $5000$ negative &
Diagonal source/target statistics from positive source to negative target; text truncated to $4000$ characters; tokenized to max length $256$ \\
\midrule
LLaDA ACT subjectivity &
Amazon Polarity train, scored by the subjectivity classifier &
top $5000$ objective, bottom $5000$ objective &
Diagonal source/target statistics from objective source to subjective target; text truncated to $4000$ characters; tokenized to max length $256$ \\
\midrule
LLaDA-V DiffMean sentiment &
SentiCap captions &
$1000$ positive, $1000$ negative &
Mean-difference vector oriented toward the negative-caption-sentiment pole; sign is matched to the LLaDA convention so $+\alpha$ steers toward negative sentiment; activations pooled over caption-token positions only, excluding image patch tokens \\
\midrule
LLaDA-V ACT sentiment &
SentiCap captions &
$1000$ positive, $1000$ negative &
Diagonal source/target statistics from positive caption-sentiment source to negative caption-sentiment target; activations pooled over caption-token positions only, excluding image patch tokens \\
\midrule
Dream-7B DiffMean / ACT sentiment &
Amazon Polarity train &
$5000$ positive, $5000$ negative &
Identical construction and orientation to the corresponding LLaDA rows, recomputed with Dream's tokenizer and hidden states over Dream's steering window \\
\bottomrule
\end{tabular}
\end{table}

\Needspace{0.38\textheight}
\subsection{Runtime and batching overhead}
\label{app:runtime-batching}

We measure end-to-end inference overhead in a representative LLaDA sentiment
setting with ACT ($\beta=0.35$), using the same $300$ prompts, generation
configuration, and layer range as the corresponding main experiment.
Generation uses batch size one, while the external sentiment scorer runs on
GPU. Models are loaded once before measurement, and several untimed warm-up
examples are executed before recording results. GPU operations are synchronized
at timing boundaries. Reported generation times exclude model loading,
post-hoc target and perplexity evaluation, and file I/O.

We construct two matched comparisons. To measure adaptive-budget overhead, we
run the adaptive policy with mean-top-$3$ confidence and
$\lambda_{\mathrm{budget}}=0.5$, and select a scorer-free stochastic fixed
schedule whose average denoising-step count matches that specific run; this
timing schedule is therefore matched to $\lambda_{\mathrm{budget}}=0.5$ alone,
rather than to the family mean used for the main results
(Appendix~\ref{app:matched-schedule-details}). To measure
counterfactual-order overhead, we compare
$\lambda_{\mathrm{order}}=0.5$ against scorer-free confidence ranking while
holding $u_t\equiv1$; both policies therefore use exactly $60$ denoising steps.
For counterfactual commitment, the full provisional completion and all
token-deletion variants from one denoising step are evaluated together in one
batched classifier call.

\begin{table}[H]
\caption{\textbf{Representative runtime overhead.}
``Other'' includes provisional-text decoding, deletion-variant construction,
calibration and normalization, and commitment-policy operations. Overhead is
measured relative to the scorer-free baseline with either matched average
denoising cost or the same fixed commitment budget.}
\label{tab:runtime-overhead}
\centering
\small
\setlength{\tabcolsep}{4.5pt}
\begin{tabular}{@{}lrrrrr@{}}
\toprule
Policy &
\makecell{Average\\steps} &
\makecell{Mean\\s/gen} &
\makecell{ms/\\step} &
\makecell{Runtime share (\%)\\LLaDA / scorer / other} &
Overhead \\
\midrule
Matched stochastic
& $33.280$ & $0.6584$ & $19.78$
& $98.5\,/\,0.0\,/\,1.5$ & reference \\

Adaptive, $\lambda_{\mathrm{budget}}=0.5$
& $33.273$ & $0.7141$ & $21.46$
& $90.5\,/\,6.2\,/\,3.3$ & $+8.5\%$ \\
\midrule
Confidence ranking, $u_t=1$
& $60.000$ & $1.1871$ & $19.79$
& $98.5\,/\,0.0\,/\,1.5$ & reference \\

Counterfactual, $\lambda_{\mathrm{order}}=0.5$
& $60.000$ & $1.7676$ & $29.46$
& $66.4\,/\,18.4\,/\,15.2$ & $+48.9\%$ \\
\bottomrule
\end{tabular}
\end{table}

\paragraph{Adaptive-budget overhead.}
The adaptive and matched stochastic fixed schedules use $33.273$ and $33.280$
denoising steps per generation, a difference of only $0.007$ steps, and their
base-denoiser times are correspondingly close ($0.6461$ versus $0.6484$ seconds
per generation). The remaining difference therefore isolates the cost of
evaluating the provisional completion and computing the adaptive budget.
Adaptive budgeting increases mean end-to-end latency from $0.6584$ to $0.7141$
seconds per generation, an overhead of $8.5\%$; the median gives the same
conclusion ($0.6471$ versus $0.6874$ seconds). Target scoring itself accounts
for $0.0446$ seconds per generation. Across the full benchmark the policy makes
$9{,}982$ scorer calls on $9{,}982$ scorer inputs, consistent with one target
evaluation per adaptive denoising step.

\paragraph{Counterfactual-order overhead.}
Counterfactual and confidence-based commitment both perform exactly $60$
LLaDA forward passes per generation, and their measured base-denoiser times are
correspondingly similar ($1.1729$ and $1.1696$ seconds), confirming that the
comparison isolates the additional commitment-order computation.
Counterfactual ranking increases mean latency from $1.1871$ to $1.7676$
seconds per generation, or $48.9\%$. The base LLaDA denoiser remains
the largest component at $66.4\%$ of total counterfactual runtime; batched
scorer inference accounts for $18.4\%$, and text decoding, token-span deletion,
normalization, and ranking for the remaining $15.2\%$.

With length-$60$ generation and $u_t\equiv1$, each example requires $60$
batched scorer calls and
\[
    \sum_{m=1}^{60}(m+1)=1{,}890
\]
scorer inputs, where $m$ is the number of unresolved positions at a denoising
step: one full completion and one deletion variant for every remaining
position at each step. Across $300$ examples, this yields $18{,}000$ scorer
calls and $567{,}000$ scorer inputs, exactly matching the observed counts.
Because all variants from a step are evaluated together, the average call
contains $31.5$ inputs rather than requiring a separate model invocation for
each deletion.

\paragraph{Interpretation.}
Adaptive budgeting adds $8.5\%$ mean wall-clock latency beyond matched denoising
computation in this representative setting, whereas counterfactual ordering is
materially more expensive: it constructs and scores a shrinking set of
leave-one-token-out variants at every step. Nevertheless, batching keeps its
end-to-end latency at $1.49\times$ the confidence-ranking baseline rather than
scaling with the raw number of counterfactual inputs. These measurements are
implementation- and scorer-dependent and should be read as a representative cost
characterization rather than a universal latency estimate. Future work could
replace repeated leave-one-token-out scoring with a learned surrogate trained to
preserve the induced counterfactual ranking.

\Needspace{0.40\textheight}
\subsection{Best-of-$N$ at matched generation budget}
\label{app:best-of-n}

Counterfactual ordering spends extra target-scorer computation inside a single
trajectory. The natural alternative is to spend a comparable budget on repeated
sampling and keep the best completion under the same scorer. We therefore
compare the commitment-order curve against best-of-$N$ selection at a matched
expected number of generations.

\paragraph{Protocol.}
Best-of-$N$ requires stochastic generation, so these runs use temperature $1.0$;
each example is generated twice, with a different RNG seed per draw. Best-of-$2$
selects the draw with the higher target score, and the perplexity we report for
that example is the perplexity of the selected draw, not the better of the two
perplexities. Best-of-$1$ is the mean over the two draws, that is, the expected
score of a single draw. Both use the same prompts, generation length, and target
scorer as the corresponding main experiment.

For each setting we form the operating point at the expected generation count
that matches that setting's measured counterfactual-ordering overhead. Rather
than sampling $N$ per example, which would discard half the available signal, we
mix the two estimates analytically:
\[
    \text{point at } \mathbb{E}[N]=1+\gamma
    \;=\;
    (1-\gamma)\cdot\text{best-of-}1 \;+\; \gamma\cdot\text{best-of-}2 .
\]
Here $\gamma\in[0,1]$ is the mixture weight, and the interpolation is applied componentwise to target score and perplexity.
For LLaDA sentiment, counterfactual ordering costs $1.7676$ against $1.1871$
seconds per generation (Table~\ref{tab:runtime-overhead}), a factor of $1.489$,
so $\gamma=0.49$ and $\mathbb{E}[N]=1.49$. Best-of-$1$ attains target $0.414$ at
perplexity $10.81$ and best-of-$2$ attains $0.554$ at $11.03$, giving
$0.51(0.414)+0.49(0.554)=0.483$ and $0.51(10.81)+0.49(11.03)=10.92$. The
remaining settings are constructed the same way from their own measured
overhead.

\begin{table}[H]
\caption{\textbf{Commitment-order curves against best-of-$N$ at a matched
generation budget.} The first two columns reproduce the commitment-order
sequences of Table~\ref{tab:counterfactual-generalization} for the four settings
evaluated without activation steering; the last column is the best-of-$N$
operating point at the expected generation count matching that setting's
counterfactual-ordering overhead.}
\label{tab:best-of-n}
\centering
\scriptsize
\setlength{\tabcolsep}{3.2pt}
\resizebox{\linewidth}{!}{%
\begin{tabular}{@{}lccc@{}}
\toprule
Model / target &
\makecell{Target\\
$\lambda_{\mathrm{order}}=0.5\to0.75\to0.9\to1$} &
\makecell{PPL\\
$\lambda_{\mathrm{order}}=0.5\to0.75\to0.9\to1$} &
\makecell{Best-of-$N$ at matched budget\\target / PPL} \\
\midrule

LLaDA / Sentiment
& $0.63\!\to\!0.53\!\to\!0.48\!\to\!0.35$
& $10.6\!\to\!10.5\!\to\!9.6\!\to\!9.5$
& $0.483$ / $10.92$ \\

Dream-7B / Sentiment
& $0.71\!\to\!0.52\!\to\!0.41\!\to\!0.30$
& $8.0\!\to\!7.8\!\to\!7.4\!\to\!7.0$
& $0.515$ / $7.95$ \\

\midrule

LLaDA / NLI entailment
& $0.54\!\to\!0.43\!\to\!0.34\!\to\!0.31$
& $26.6\!\to\!25.1\!\to\!23.8\!\to\!22.5$
& $0.404$ / $25.51$ \\

Dream-7B / NLI entailment
& $0.60\!\to\!0.47\!\to\!0.39\!\to\!0.34$
& $27.3\!\to\!25.0\!\to\!23.5\!\to\!22.4$
& $0.470$ / $26.12$ \\

\bottomrule
\end{tabular}}
\end{table}

\paragraph{Reading the comparison.}
Taking $\lambda_{\mathrm{order}}=0.75$ in Table~\ref{tab:best-of-n} as the
comparison point, commitment ordering
attains a higher target score at a lower perplexity than best-of-$N$ on LLaDA
sentiment ($0.53$ / $10.5$ against $0.483$ / $10.92$), Dream-7B sentiment
($0.52$ / $7.8$ against $0.515$ / $7.95$), and LLaDA NLI ($0.43$ / $25.1$
against $0.404$ / $25.51$); on Dream-7B NLI the two reach the same entailment
score ($0.47$ against $0.470$) and the commitment point does so at $1.1$ lower
perplexity, which we read as a match rather than an improvement.

The budget match is derived from wall-clock measurement: $\mathbb{E}[N]$ is set
to the measured latency ratio between counterfactual ordering and confidence
ranking (Table~\ref{tab:runtime-overhead}). It therefore assumes that generation
cost scales linearly in the number of draws, and it does not charge best-of-$N$
for the two target-scorer calls used to select between draws. The second
assumption is conservative rather than favorable: billing those calls would
reduce the number of draws the budget affords and move the best-of-$N$ point
down, not up. The magnitude is small---in the adaptive run of
Table~\ref{tab:runtime-overhead}, scorer inference accounts for $6.2\%$ of
runtime spread over roughly thirty-three calls per generation, so two calls cost
well under $1\%$ of a single generation---but it runs against best-of-$N$ rather
than for it.

Best-of-$N$ also requires sampling diversity and so cannot be run at temperature
$0$, at which both draws would be identical and best-of-$2$ would collapse to
best-of-$1$; the two regimes therefore differ in sampling as well as in selection
rule. We do not read this as favoring either side. Raising the temperature
increases the target score at a perplexity cost, tracing a trade-off of the same
shape as the commitment sweep: on Dream-7B sentiment the temperature-$1$ single
draw attains $0.432$ at perplexity $7.97$, which $\lambda_{\mathrm{order}}=0.5$
exceeds at essentially identical perplexity ($0.705$ at $7.98$), while on LLaDA
sentiment the single draw ($0.414$ at $10.81$) is below
$\lambda_{\mathrm{order}}=0.9$ ($0.48$ at $9.60$) on both axes. The comparison
is best read as one between two trade-off curves rather than between two
isolated points.
\clearpage
\section{Full Results}
\label{app:full-tables}

This appendix reports the numeric values underlying the commitment-order,
adaptive-budget, and compute-matched comparisons, together with the
corresponding per-configuration plots. Each subsection pairs one table with the
plots for the same set of runs. The main paper shows one representative
configuration per mechanism and summarizes the rest here. All entries are means
over valid generations.

\Needspace{0.38\textheight}
\subsection{Counterfactual commitment-order sweeps}
\label{app:counterfactual-tables}
\label{app:counterfactual-plots}

\begin{table}[H]
\caption{\textbf{Full counterfactual commitment-order sequences.}
Target scores and PPL are reported as
$\lambda_{\mathrm{order}}$ moves from stronger counterfactual weighting
($0.5$) toward confidence-only commitment ($1$).
The final column indicates whether the target and PPL sequences are
non-increasing over the tested commitment weights. Full-precision Dream-7B
values are in Tables~\ref{tab:dream-order} and~\ref{tab:dream-nli-order}.}
\label{tab:counterfactual-generalization}
\centering
\scriptsize
\setlength{\tabcolsep}{3.2pt}
\resizebox{\linewidth}{!}{%
\begin{tabular}{@{}llccc@{}}
\toprule
Model / target & Activation &
\makecell{Target\\
$\lambda_{\mathrm{order}}=0.5\to0.75\to0.9\to1$} &
\makecell{PPL\\
$\lambda_{\mathrm{order}}=0.5\to0.75\to0.9\to1$} &
\makecell{Ordering consistent\\target / PPL} \\
\midrule

LLaDA / Sentiment & None
& $0.63\!\to\!0.53\!\to\!0.48\!\to\!0.35$
& $10.6\!\to\!10.5\!\to\!9.6\!\to\!9.5$
& \cmark / \cmark \\

LLaDA / Sentiment & DiffMean
& $0.80\!\to\!0.62\!\to\!0.56\!\to\!0.50$
& $11.5\!\to\!11.0\!\to\!9.6\!\to\!8.9$
& \cmark / \cmark \\

LLaDA / Sentiment & ACT
& $0.75\!\to\!0.69\!\to\!0.62\!\to\!0.51$
& $11.2\!\to\!11.1\!\to\!10.1\!\to\!9.2$
& \cmark / \cmark \\

\midrule

LLaDA / Subjectivity & None
& $0.64\!\to\!0.57\!\to\!0.54\!\to\!0.49$
& $12.5\!\to\!10.7\!\to\!9.4\!\to\!8.9$
& \cmark / \cmark \\

LLaDA / Subjectivity & DiffMean
& $0.75\!\to\!0.70\!\to\!0.63\!\to\!0.57$
& $14.2\!\to\!11.0\!\to\!10.3\!\to\!8.6$
& \cmark / \cmark \\

LLaDA / Subjectivity & ACT
& $0.70\!\to\!0.60\!\to\!0.54\!\to\!0.47$
& $12.2\!\to\!9.8\!\to\!9.1\!\to\!8.6$
& \cmark / \cmark \\

\midrule

LLaDA-V / Caption sentiment & None
& $0.84\!\to\!0.47\!\to\!0.33\!\to\!0.32$
& $13.5\!\to\!12.3\!\to\!12.0\!\to\!11.8$
& \cmark / \cmark \\

LLaDA-V / Caption sentiment & DiffMean
& $0.89\!\to\!0.47\!\to\!0.38\!\to\!0.33$
& $15.0\!\to\!12.1\!\to\!11.6\!\to\!11.3$
& \cmark / \cmark \\

LLaDA-V / Caption sentiment & ACT
& $0.57\!\to\!0.49\!\to\!0.45\!\to\!0.42$
& $11.5\!\to\!11.1\!\to\!11.1\!\to\!11.0$
& \cmark / \cmark \\

\midrule

LLaDA / NLI entailment & None
& $0.54\!\to\!0.43\!\to\!0.34\!\to\!0.31$
& $26.6\!\to\!25.1\!\to\!23.8\!\to\!22.5$
& \cmark / \cmark \\

LLaDA-V / CLIPScore & None
& $30.9\!\to\!26.8\!\to\!25.7\!\to\!25.5$
& $17.8\!\to\!11.3\!\to\!10.2\!\to\!10.1$
& \cmark / \cmark \\

\midrule

Dream-7B / Sentiment & None
& $0.71\!\to\!0.52\!\to\!0.41\!\to\!0.30$
& $8.0\!\to\!7.8\!\to\!7.4\!\to\!7.0$
& \cmark / \cmark \\

Dream-7B / Sentiment & DiffMean
& $0.80\!\to\!0.64\!\to\!0.54\!\to\!0.44$
& $7.8\!\to\!7.0\!\to\!6.7\!\to\!6.4$
& \cmark / \cmark \\

Dream-7B / Sentiment & ACT
& $0.76\!\to\!0.65\!\to\!0.50\!\to\!0.42$
& $8.4\!\to\!7.7\!\to\!7.6\!\to\!6.9$
& \cmark / \cmark \\

Dream-7B / NLI entailment & None
& $0.60\!\to\!0.47\!\to\!0.39\!\to\!0.34$
& $27.3\!\to\!25.0\!\to\!23.5\!\to\!22.4$
& \cmark / \cmark \\

\bottomrule
\end{tabular}}
\end{table}

\Needspace{0.38\textheight}

The LLaDA sentiment sweep is the canonical setting and appears as
Fig.~\ref{fig:counterfactual-sentiment} in the main text; the NLI and CLIPScore
sweeps appear as Fig.~\ref{fig:task-counterfactual}. The remaining control
settings are shown below.

\begin{figure}[H]
    \centering
    \includegraphics[width=0.76\linewidth]{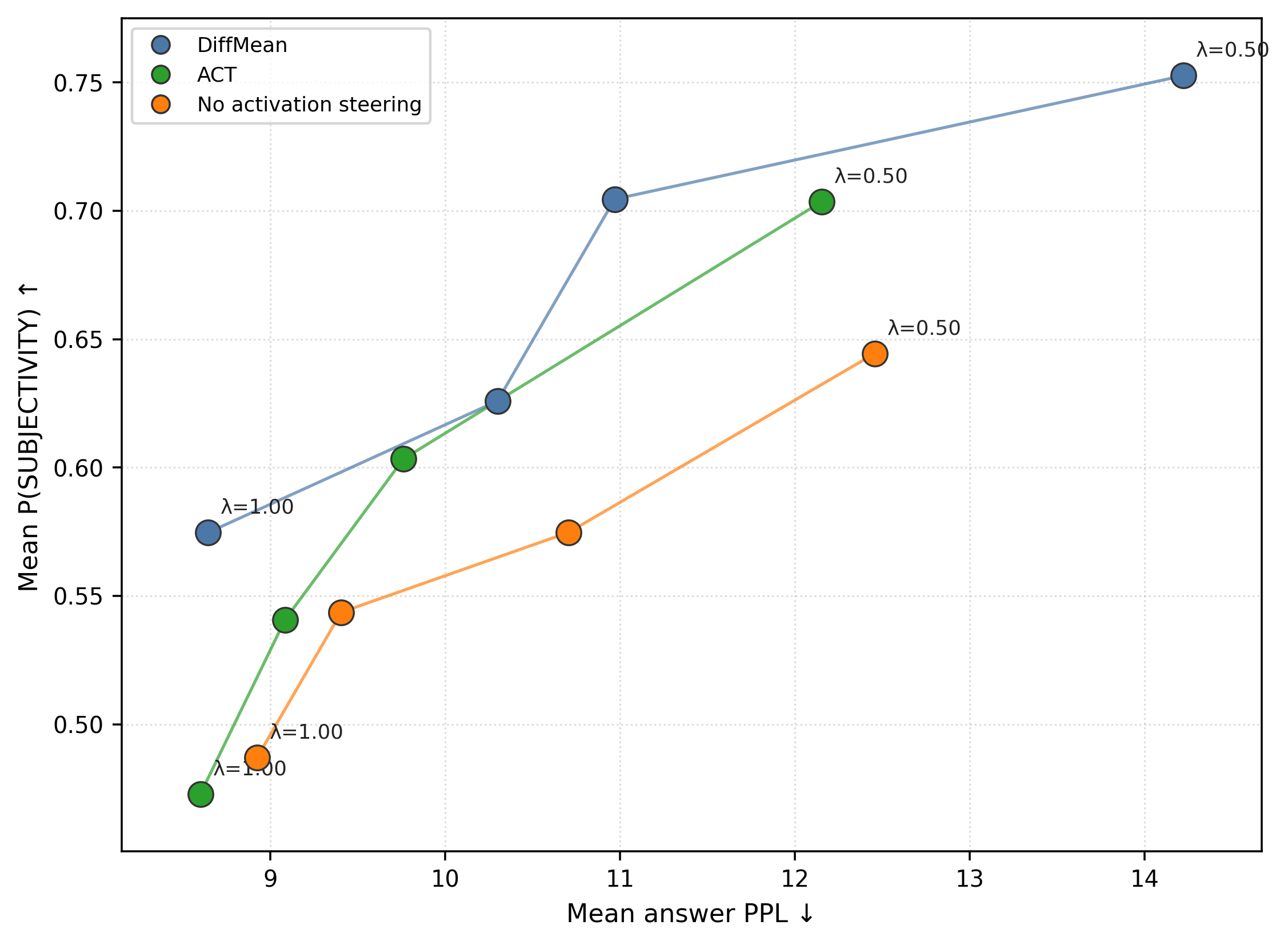}
    \caption{Counterfactual commitment-order trade-off plot for LLaDA subjectivity control.}
    \label{fig:app-counterfactual-objectivity}
\end{figure}
\FloatBarrier

\begin{figure}[H]
    \centering
    \includegraphics[width=0.76\linewidth]{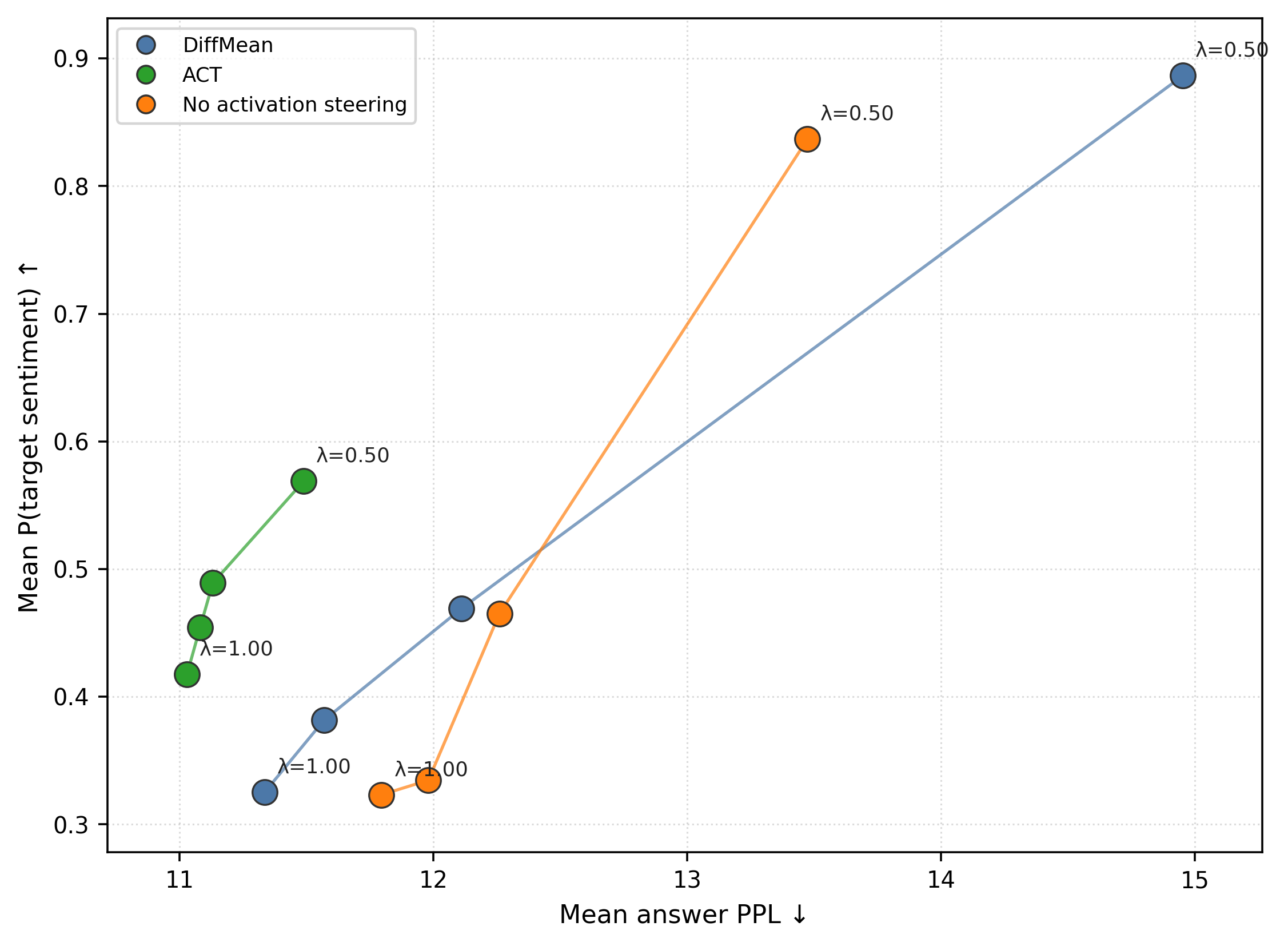}
    \caption{Counterfactual commitment-order trade-off plot for LLaDA-V caption sentiment control.}
    \label{fig:app-counterfactual-lladav-sentiment}
\end{figure}
\FloatBarrier

\Needspace{0.45\textheight}
\paragraph{Extreme commitment-order diagnostics.}
As a control on the commitment-order result, we evaluate one representative
setting---LLaDA sentiment with no activation steering---under pure counterfactual
ordering ($\lambda_{\mathrm{order}}=0$) and under random commitment order. These
are diagnostic anchors rather than practical operating points: pure
counterfactual ordering discards model confidence entirely, while random
ordering ignores both confidence and target information.

The contrast separates two explanations of the main sweep. If any departure from
confidence ranking produced the observed target movement, random ordering would
reproduce it. Instead, random ordering pays the largest perplexity cost of any
policy shown ($13.1$, against $9.5$ for confidence ranking) while reaching a
target score of only $0.41$, barely above the $0.35$ confidence baseline and far
below the $0.63$ that $\lambda_{\mathrm{order}}=0.5$ attains at a perplexity of
$10.6$. Pure counterfactual ordering reaches $0.70$, above
$\lambda_{\mathrm{order}}=0.5$, but at a perplexity of $12.5$; discarding
confidence entirely therefore buys better target but has substantial costs in fluency,
which is why the tested sweep retains a confidence component throughout.
Table~\ref{tab:app-counterfactual-extremes} collects the four policies.

\begin{table}[H]
\caption{\textbf{Extreme commitment-order diagnostics for LLaDA sentiment
without activation steering.} All four policies use $u_t\equiv1$ and therefore
the same $60$ denoising steps. Random ordering incurs the largest perplexity
cost of any policy shown while barely moving the target; the intermediate rows
are reproduced from Table~\ref{tab:counterfactual-generalization} for
reference.}
\label{tab:app-counterfactual-extremes}
\centering
\small
\setlength{\tabcolsep}{9pt}
\begin{tabular}{@{}lcc@{}}
\toprule
Commitment policy & Target & PPL \\
\midrule
Confidence ranking ($\lambda_{\mathrm{order}}=1$) & $0.35$ & $9.5$ \\
Mixed ($\lambda_{\mathrm{order}}=0.5$)            & $0.63$ & $10.6$ \\
Pure counterfactual ($\lambda_{\mathrm{order}}=0$) & $0.70$ & $12.5$ \\
Random commitment order                             & $0.41$ & $13.1$ \\
\bottomrule
\end{tabular}
\end{table}
\FloatBarrier

\clearpage
\Needspace{0.38\textheight}
\subsection{Adaptive-budget sweeps}
\label{app:adaptive-tables}
\label{app:adaptive-dist-plots}

\begin{table}[H]
\caption{\textbf{Full adaptive-budget sequences.}
Target scores and PPL are reported for
$\lambda_{\mathrm{budget}}\in\{0,0.5,1\}$.
The scoring map is fixed to confidence throughout, so movement across each
sequence changes only the adaptive commitment budget. Full-precision Dream-7B
values are in Tables~\ref{tab:dream-budget} and~\ref{tab:dream-nli-budget}.}
\label{tab:adaptive-generalization}
\centering
\scriptsize
\setlength{\tabcolsep}{3.4pt}
\resizebox{\linewidth}{!}{%
\begin{tabular}{@{}llccc@{}}
\toprule
Model / target & Activation &
\makecell{Target\\
$\lambda_{\mathrm{budget}}=0\to0.5\to1$} &
\makecell{PPL\\
$\lambda_{\mathrm{budget}}=0\to0.5\to1$} &
\makecell{Ordering consistent\\target / PPL} \\
\midrule

LLaDA / Sentiment & None
& $0.51\!\to\!0.39\!\to\!0.33$
& $18.7\!\to\!15.6\!\to\!12.5$
& \cmark / \cmark \\

LLaDA / Sentiment & DiffMean
& $0.65\!\to\!0.55\!\to\!0.51$
& $21.3\!\to\!14.1\!\to\!12.9$
& \cmark / \cmark \\

LLaDA / Sentiment & ACT
& $0.65\!\to\!0.57\!\to\!0.52$
& $21.1\!\to\!15.6\!\to\!13.0$
& \cmark / \cmark \\

\midrule

LLaDA / Subjectivity & None
& $0.61\!\to\!0.55\!\to\!0.51$
& $18.4\!\to\!13.2\!\to\!11.6$
& \cmark / \cmark \\

LLaDA / Subjectivity & DiffMean
& $0.61\!\to\!0.60\!\to\!0.58$
& $19.7\!\to\!13.8\!\to\!10.5$
& \cmark / \cmark \\

LLaDA / Subjectivity & ACT
& $0.62\!\to\!0.57\!\to\!0.52$
& $17.5\!\to\!13.3\!\to\!10.5$
& \cmark / \cmark \\

\midrule

LLaDA-V / Caption sentiment & None
& $0.65\!\to\!0.40\!\to\!0.27$
& $18.7\!\to\!15.1\!\to\!13.3$
& \cmark / \cmark \\

LLaDA-V / Caption sentiment & DiffMean
& $0.76\!\to\!0.57\!\to\!0.35$
& $18.9\!\to\!15.4\!\to\!13.2$
& \cmark / \cmark \\

LLaDA-V / Caption sentiment & ACT
& $0.56\!\to\!0.44\!\to\!0.31$
& $16.3\!\to\!13.0\!\to\!12.2$
& \cmark / \cmark \\

\midrule

LLaDA / NLI entailment & None
& $0.40\!\to\!0.35\!\to\!0.31$
& $31.0\!\to\!26.7\!\to\!24.4$
& \cmark / \cmark \\

LLaDA-V / CLIPScore & None
& $29.4\!\to\!28.6\!\to\!28.0$
& $19.5\!\to\!14.8\!\to\!13.6$
& \cmark / \cmark \\

\midrule

Dream-7B / Sentiment & None
& $0.50\!\to\!0.45\!\to\!0.43$
& $13.8\!\to\!11.4\!\to\!9.3$
& \cmark / \cmark \\

Dream-7B / Sentiment & DiffMean
& $0.69\!\to\!0.64\!\to\!0.62$
& $11.3\!\to\!9.7\!\to\!8.4$
& \cmark / \cmark \\

Dream-7B / Sentiment & ACT
& $0.64\!\to\!0.59\!\to\!0.52$
& $12.9\!\to\!10.5\!\to\!8.7$
& \cmark / \cmark \\

Dream-7B / NLI entailment & None
& $0.43\!\to\!0.34\!\to\!0.26$
& $33.3\!\to\!28.6\!\to\!24.8$
& \cmark / \cmark \\

\bottomrule
\end{tabular}}
\end{table}

\Needspace{0.38\textheight}

\paragraph{LLaDA sentiment.}

\begin{figure}[H]
    \centering
    {\small\bfseries No activation}\par\vspace{0.15em}
    \includegraphics[width=0.80\linewidth]{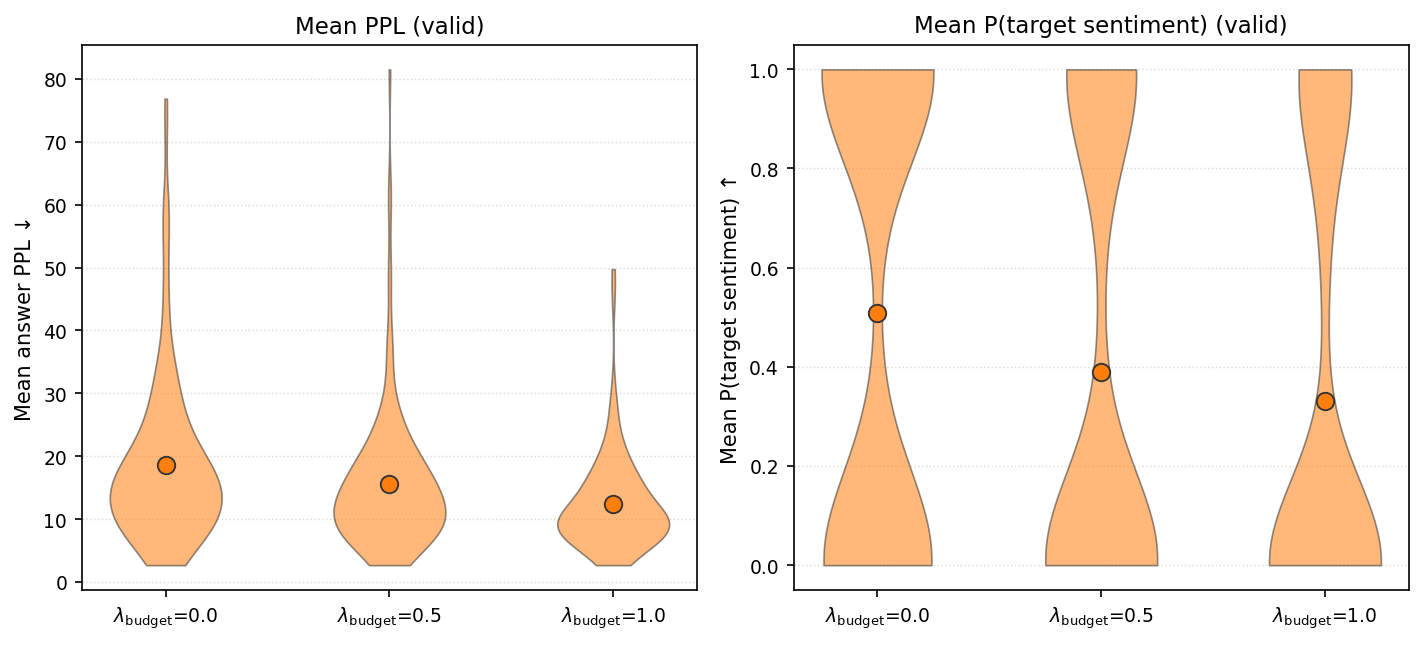}
    \vspace{0.45em}

    {\small\bfseries DiffMean}\par\vspace{0.15em}
    \includegraphics[width=0.80\linewidth]{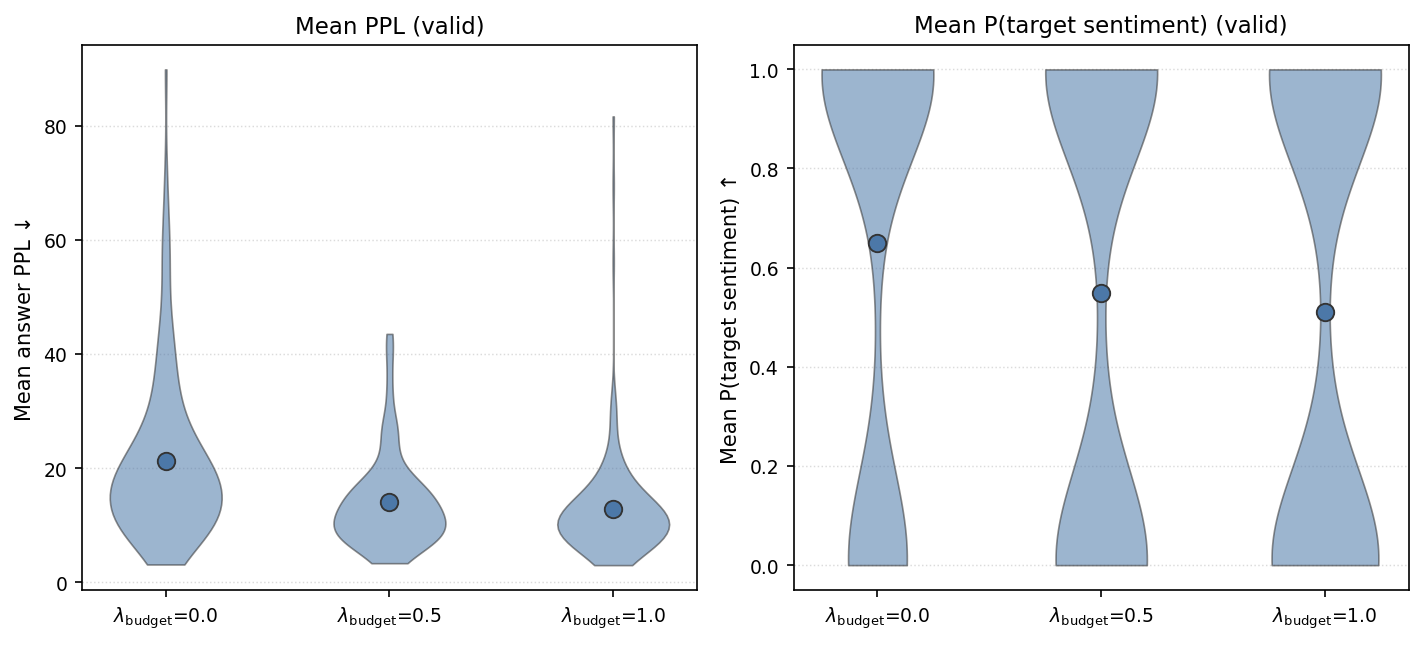}
    \vspace{0.45em}

    {\small\bfseries ACT}\par\vspace{0.15em}
    \includegraphics[width=0.80\linewidth]{images/pareto/sentiment/adaptive_dist_sentiment_act_clean.png}
    \vspace{0.45em}
    \caption{Adaptive-budget sweeps for LLaDA sentiment control under no activation steering, DiffMean, and ACT.}
    \label{fig:app-adaptive-sentiment}
\end{figure}
\FloatBarrier

\paragraph{LLaDA subjectivity.}

\begin{figure}[H]
    \centering
    {\small\bfseries No activation}\par\vspace{0.15em}
    \includegraphics[width=0.80\linewidth]{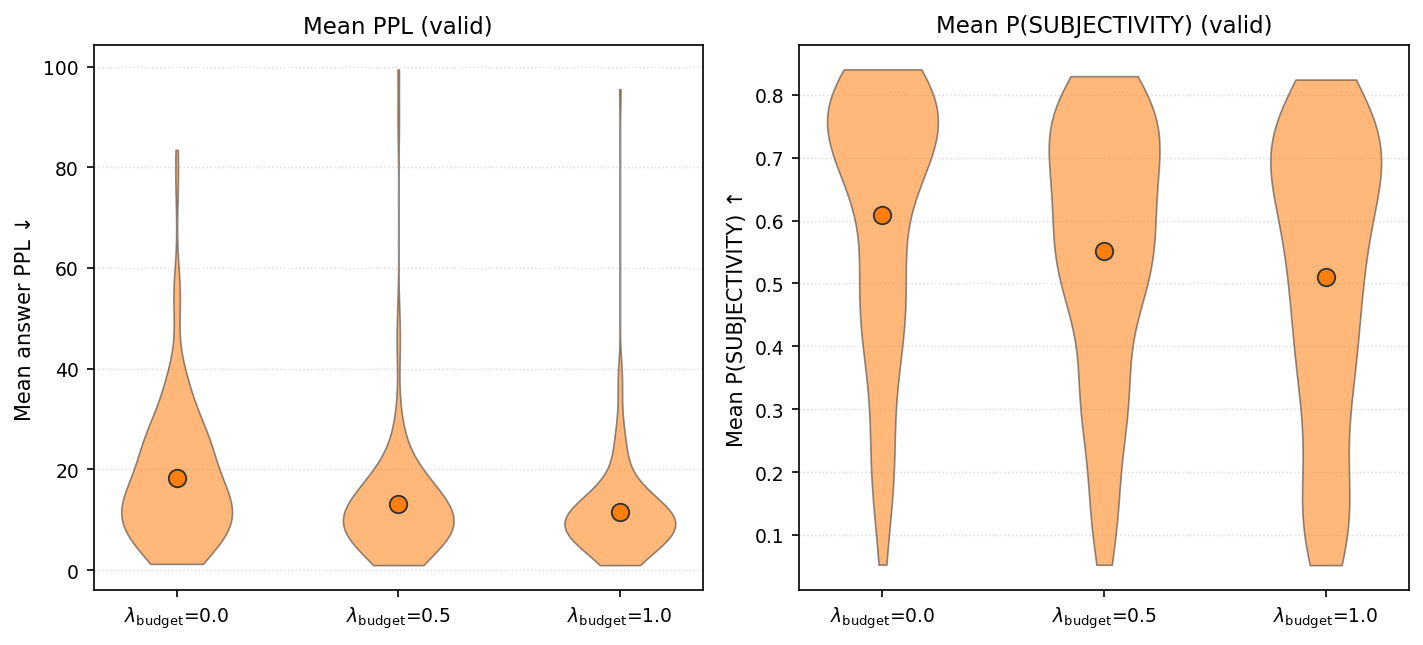}
    \vspace{0.45em}

    {\small\bfseries DiffMean}\par\vspace{0.15em}
    \includegraphics[width=0.80\linewidth]{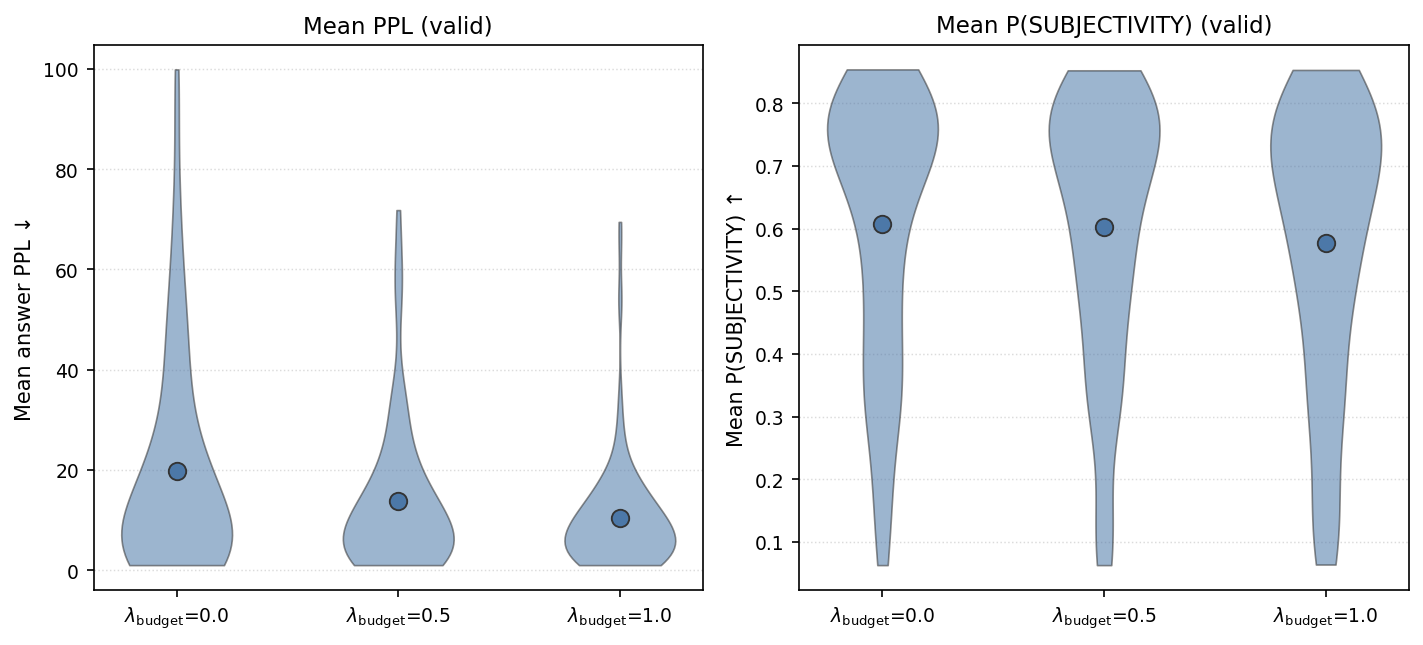}
    \vspace{0.45em}

    {\small\bfseries ACT}\par\vspace{0.15em}
    \includegraphics[width=0.80\linewidth]{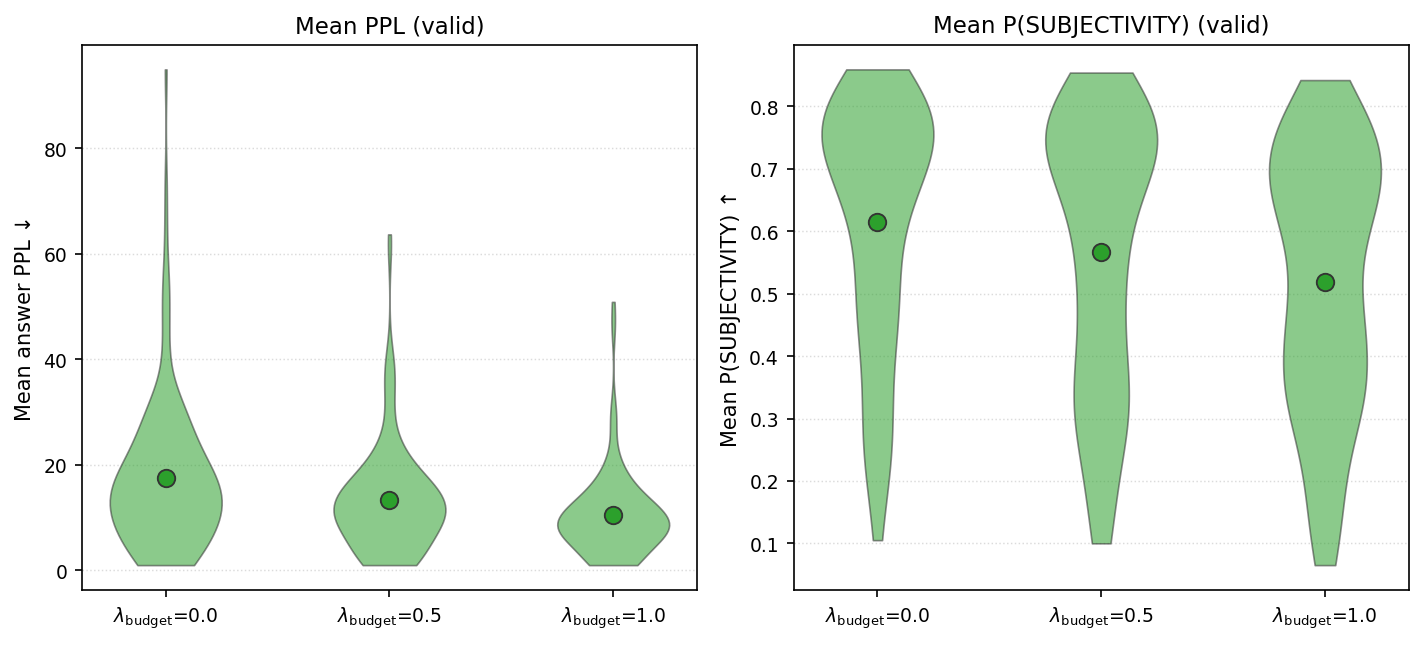}
    \vspace{0.45em}
    \caption{Adaptive-budget sweeps for LLaDA subjectivity control under no activation steering, DiffMean, and ACT.}
    \label{fig:app-adaptive-objectivity}
\end{figure}
\FloatBarrier

\paragraph{LLaDA-V caption sentiment.}
Sentiment control on a caption is only meaningful if the caption still describes
its image. Just as reference perplexity guards fluency in the text settings,
CLIPScore guards image grounding here, so we report it alongside sentiment and
perplexity throughout the caption-sentiment experiments. If commitment control
were driving captions toward negatively associated content untethered from the
image, image--text alignment would degrade; across these sweeps it does not
change substantially, indicating that the captions continue to describe their
images while their sentiment changes. The vision-language setting that
\emph{optimizes} CLIPScore is separate and is discussed in
Section~\ref{sec:results-task}.

\begin{figure}[H]
    \centering
    {\small\bfseries No activation}\par\vspace{0.15em}
    \includegraphics[width=0.80\linewidth]{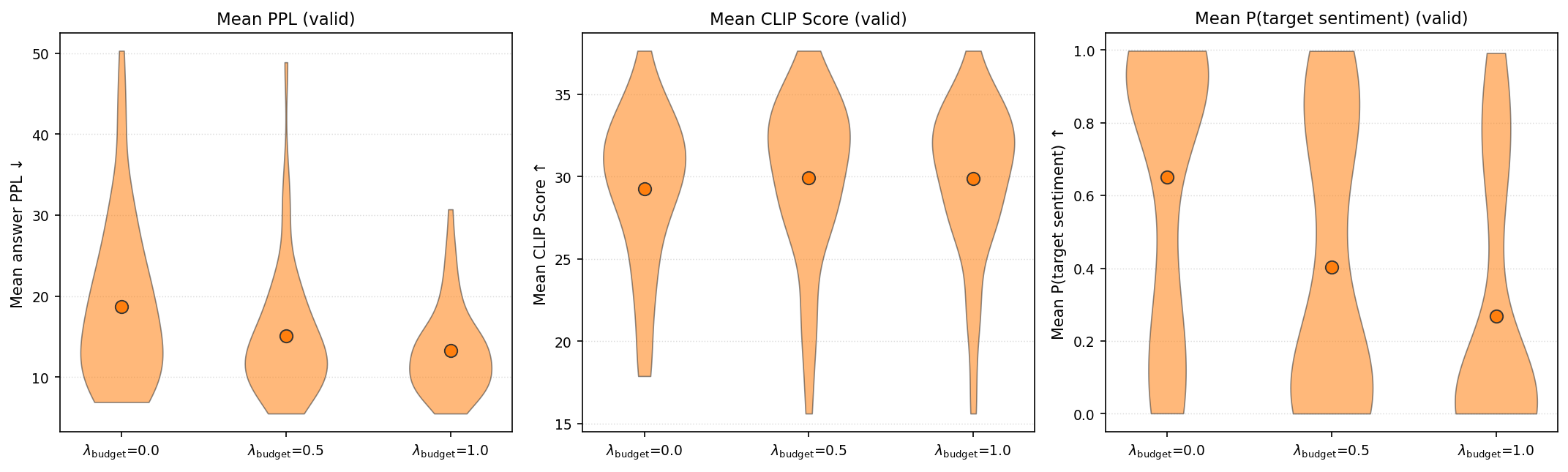}
    \vspace{0.45em}

    {\small\bfseries DiffMean}\par\vspace{0.15em}
    \includegraphics[width=0.80\linewidth]{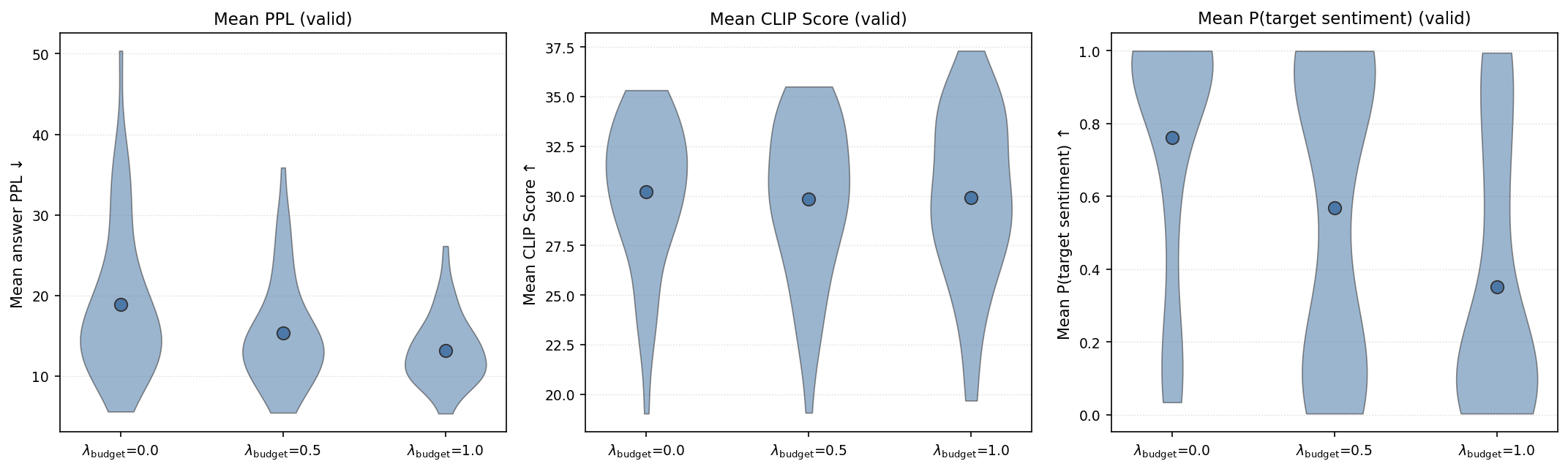}
    \vspace{0.45em}

    {\small\bfseries ACT}\par\vspace{0.15em}
    \includegraphics[width=0.80\linewidth]{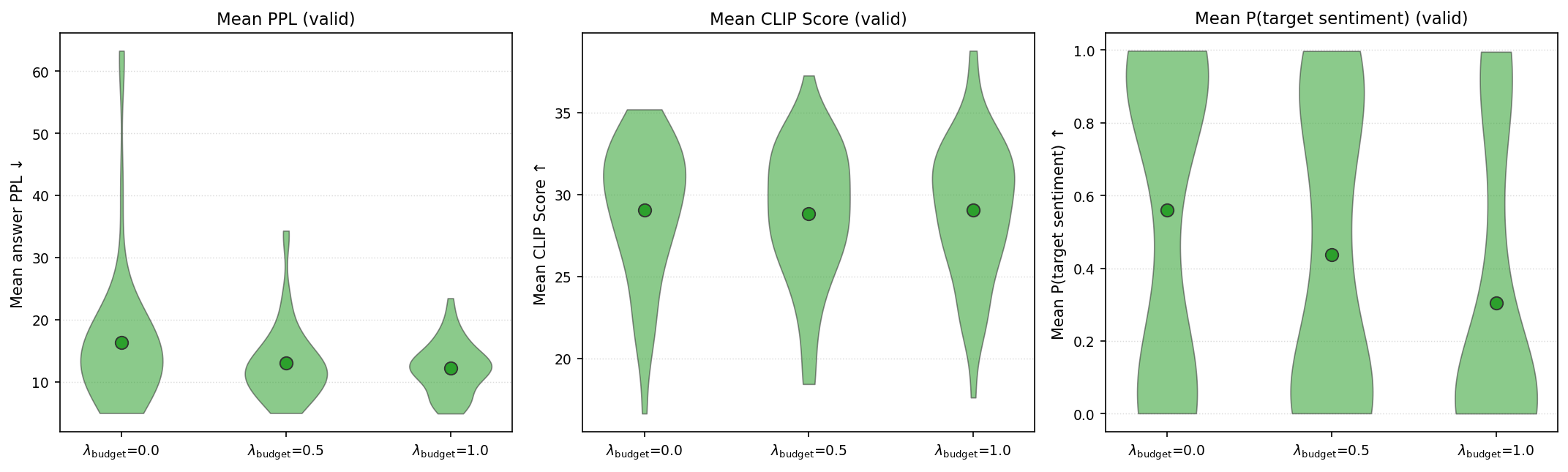}
    \vspace{0.45em}
    \caption{Adaptive-budget sweeps for LLaDA-V caption sentiment under no activation steering, DiffMean, and ACT. The CLIPScore panel is the image-grounding check described above.}
    \label{fig:app-adaptive-lladav-sentiment}
\end{figure}
\FloatBarrier

\paragraph{Task-success targets.}

\begin{figure}[H]
    \centering
    \includegraphics[width=0.82\linewidth]{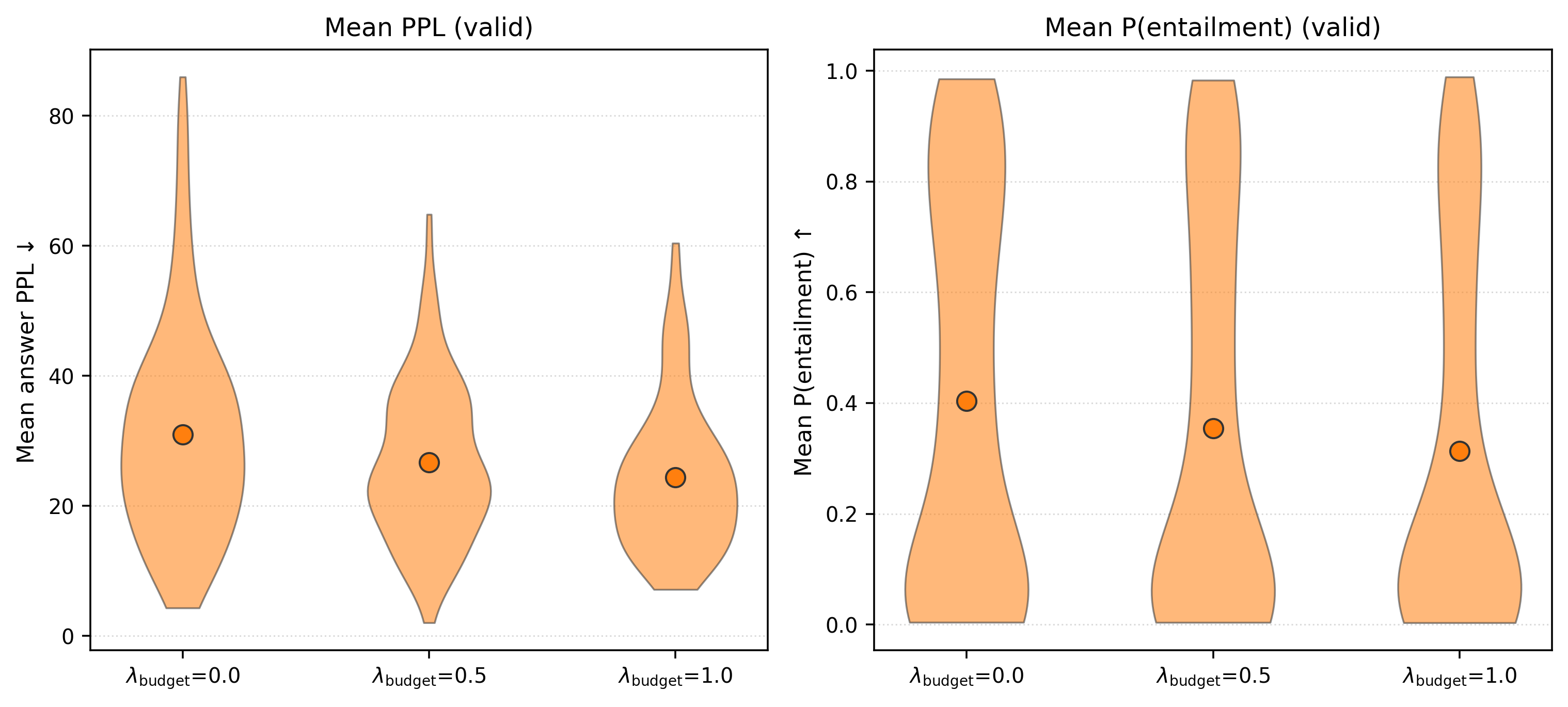}
    \caption{Adaptive-budget sweep for NLI entailment.}
    \label{fig:app-adaptive-nli}
\end{figure}
\FloatBarrier

\begin{figure}[H]
    \centering
    \includegraphics[width=0.82\linewidth]{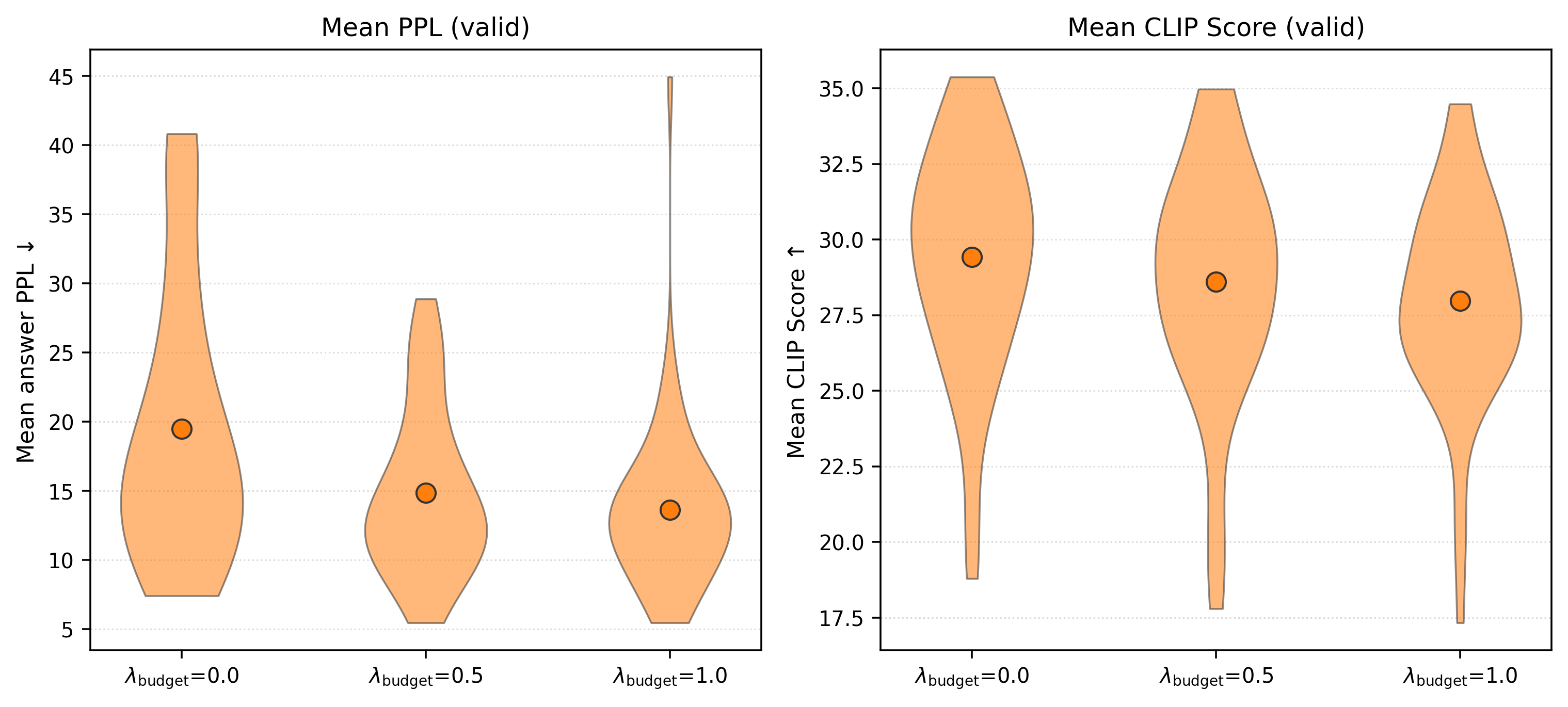}
    \caption{Adaptive-budget sweep for CLIPScore image--text alignment.}
    \label{fig:app-adaptive-clip}
\end{figure}
\FloatBarrier

\clearpage
\Needspace{0.38\textheight}
\subsection{Compute-matched stochastic comparisons}
\label{app:matched-tables}
\label{app:matched-cost-plots}

\begin{table}[H]
\caption{\textbf{Full compute-matched stochastic comparisons.}
For each setting, entries from left to right correspond to
$\lambda_{\mathrm{budget}}\in\{0,0.5,1\}$.
Each adaptive policy is compared with the matched scorer-free,
non-adaptive stochastic schedule for that setting
(Appendix~\ref{app:matched-schedule-details}).
Deltas are adaptive minus matched stochastic: positive $\Delta$Target favors
adaptation, while negative $\Delta$PPL indicates lower perplexity.
``Competitive'' indicates that the adaptive point is not jointly dominated in
target and PPL by its compute-matched stochastic baseline. Two Dream-7B
sentiment entries---DiffMean at $\lambda_{\mathrm{budget}}=0$ and no activation
at $\lambda_{\mathrm{budget}}=0.5$---improve both metrics simultaneously.}
\label{tab:matched-stochastic-full}
\centering
\scriptsize
\setlength{\tabcolsep}{3.3pt}
\resizebox{\linewidth}{!}{%
\begin{tabular}{@{}llccc@{}}
\toprule
Model / target & Activation &
\makecell{$\Delta$Target\\
$\lambda_{\mathrm{budget}}=0\to0.5\to1$} &
\makecell{$\Delta$PPL\\
$\lambda_{\mathrm{budget}}=0\to0.5\to1$} &
\makecell{Competitive\\$0\;/\;0.5\;/\;1$} \\
\midrule

LLaDA / Sentiment & None
& $+0.14\!\to\!+0.02\!\to\!-0.04$
& $+1.6\!\to\!-1.5\!\to\!-4.6$
& \cmark / \cmark / \cmark \\

LLaDA / Sentiment & DiffMean
& $+0.07\!\to\!-0.03\!\to\!-0.07$
& $+4.2\!\to\!-3.0\!\to\!-4.2$
& \cmark / \cmark / \cmark \\

LLaDA / Sentiment & ACT
& $+0.04\!\to\!-0.04\!\to\!-0.09$
& $+4.9\!\to\!-0.6\!\to\!-3.2$
& \cmark / \cmark / \cmark \\

\midrule

LLaDA / Subjectivity & None
& $+0.03\!\to\!-0.03\!\to\!-0.07$
& $+2.8\!\to\!-2.4\!\to\!-4.0$
& \cmark / \cmark / \cmark \\

LLaDA / Subjectivity & DiffMean
& $+0.02\!\to\!+0.02\!\to\!-0.01$
& $+3.3\!\to\!-2.6\!\to\!-6.0$
& \cmark / \cmark / \cmark \\

LLaDA / Subjectivity & ACT
& $+0.05\!\to\!+0.00\!\to\!-0.04$
& $+4.1\!\to\!-0.1\!\to\!-2.9$
& \cmark / \cmark / \cmark \\

\midrule

LLaDA-V / Caption sentiment & None
& $+0.24\!\to\!-0.01\!\to\!-0.15$
& $+2.5\!\to\!-1.1\!\to\!-2.9$
& \cmark / \cmark / \cmark \\

LLaDA-V / Caption sentiment & DiffMean
& $+0.21\!\to\!+0.02\!\to\!-0.20$
& $+2.8\!\to\!-0.8\!\to\!-3.0$
& \cmark / \cmark / \cmark \\

LLaDA-V / Caption sentiment & ACT
& $+0.08\!\to\!-0.05\!\to\!-0.18$
& $+1.6\!\to\!-1.7\!\to\!-2.5$
& \cmark / \cmark / \cmark \\

\midrule

LLaDA / NLI entailment & None
& $+0.03\!\to\!-0.02\!\to\!-0.06$
& $+2.5\!\to\!-1.8\!\to\!-4.1$
& \cmark / \cmark / \cmark \\

LLaDA-V / CLIPScore & None
& $+0.88\!\to\!+0.05\!\to\!-0.57$
& $+2.7\!\to\!-1.9\!\to\!-3.1$
& \cmark / \cmark / \cmark \\

\midrule

Dream-7B / Sentiment & None
& $+0.06\!\to\!+0.01\!\to\!-0.01$
& $+0.1\!\to\!-2.2\!\to\!-4.3$
& \cmark / \cmark / \cmark \\

Dream-7B / Sentiment & DiffMean
& $+0.03\!\to\!-0.02\!\to\!-0.03$
& $-0.8\!\to\!-2.5\!\to\!-3.7$
& \cmark / \cmark / \cmark \\

Dream-7B / Sentiment & ACT
& $+0.03\!\to\!-0.02\!\to\!-0.09$
& $+0.2\!\to\!-2.3\!\to\!-4.0$
& \cmark / \cmark / \cmark \\

Dream-7B / NLI entailment & None
& $+0.09\!\to\!-0.01\!\to\!-0.08$
& $+1.8\!\to\!-2.9\!\to\!-6.7$
& \cmark / \cmark / \cmark \\

\bottomrule
\end{tabular}}
\end{table}

The plots in this subsection compare adaptive budgeting directly with the scorer-free compute-matched stochastic fixed schedule. Fixed-$1/2/3$ remain visible as contextual anchors, while the matched stochastic point provides the controlled comparison used to assess state-dependent adaptation at essentially identical average denoising cost.

\begin{figure}[H]
    \centering
    {\small\bfseries No activation}\par\vspace{0.15em}
    \includegraphics[width=0.80\linewidth]{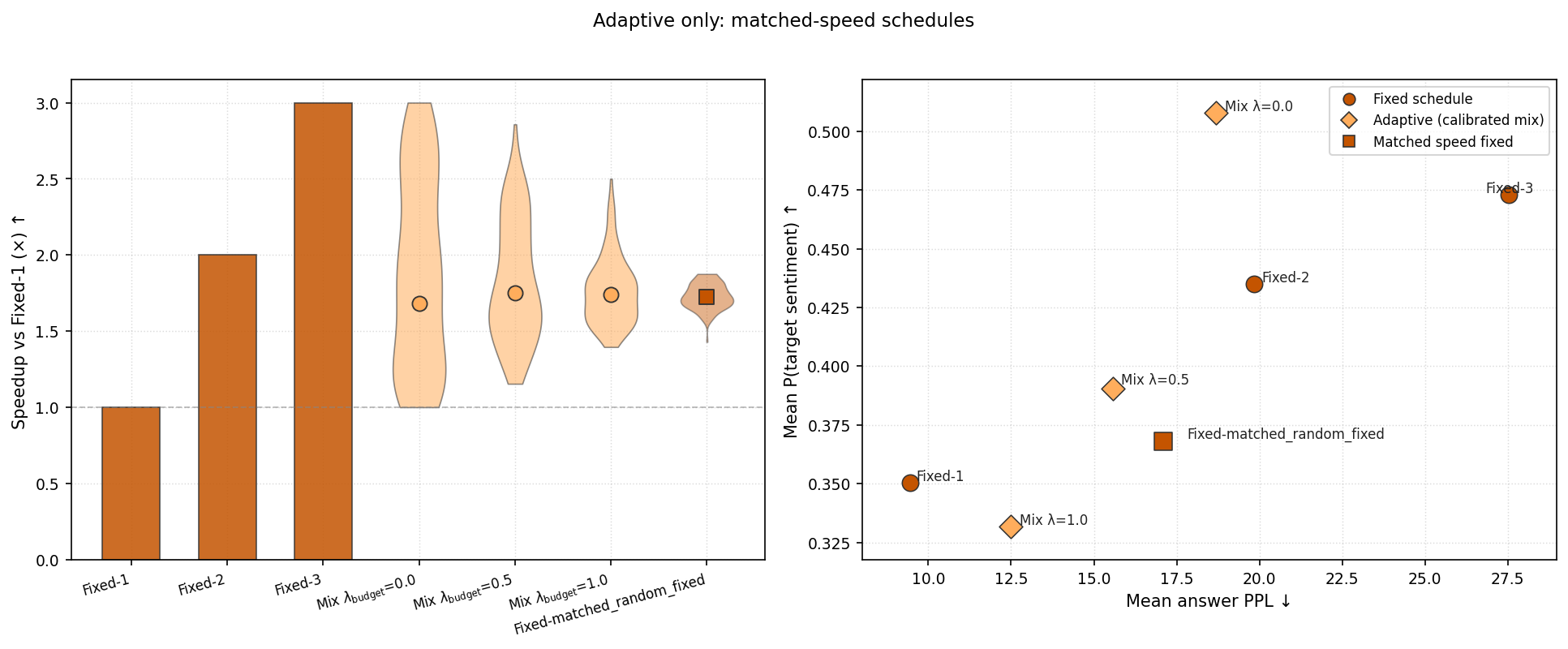}
    \vspace{0.45em}

    {\small\bfseries DiffMean}\par\vspace{0.15em}
    \includegraphics[width=0.80\linewidth]{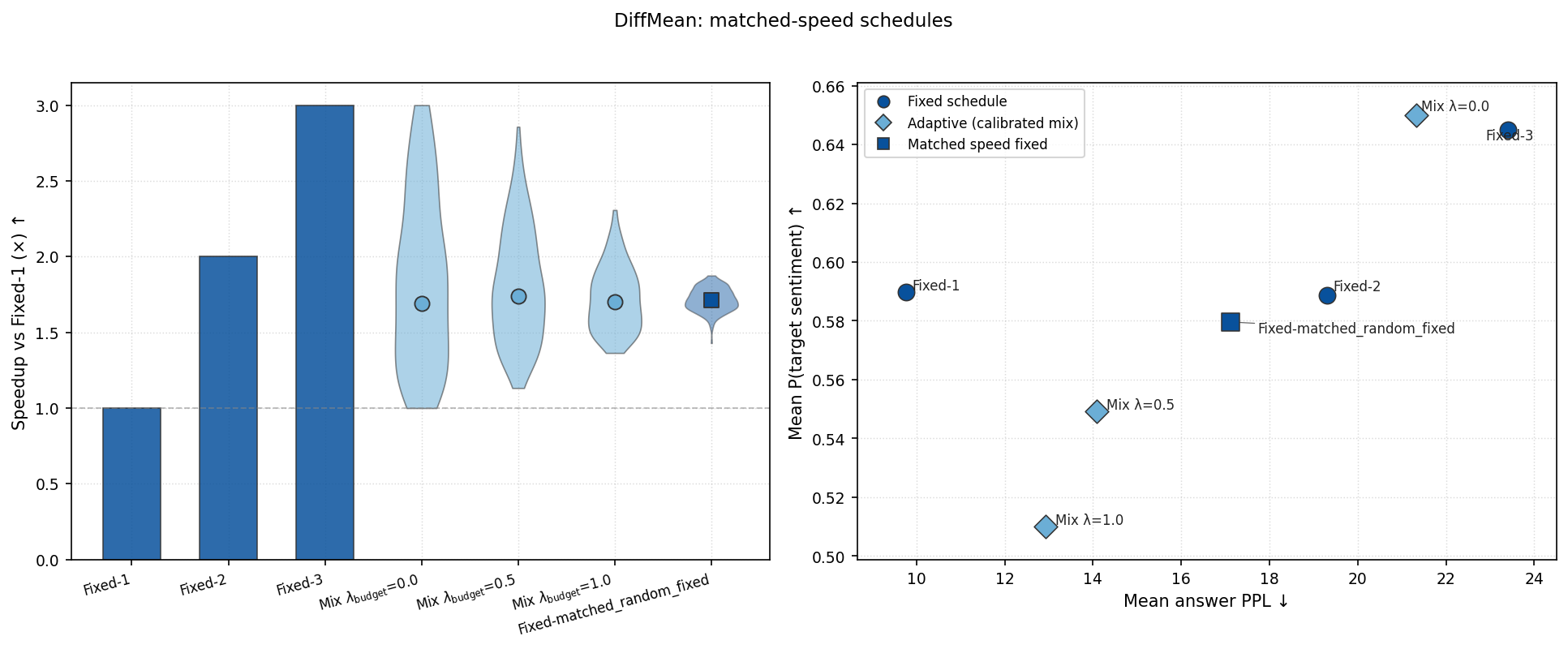}
    \vspace{0.45em}

    {\small\bfseries ACT}\par\vspace{0.15em}
    \includegraphics[width=0.80\linewidth]{images/pareto/sentiment/matched_speed_sentiment_act_clean.png}
    \vspace{0.45em}
    \caption{Adaptive versus compute-matched stochastic fixed schedules for LLaDA sentiment under no activation steering, DiffMean, and ACT. The matched baseline is scorer-free, non-adaptive, confidence-ranked, and matched in average denoising steps.}
    \label{fig:app-matched-sentiment}
\end{figure}
\FloatBarrier

\begin{figure}[H]
    \centering
    {\small\bfseries No activation}\par\vspace{0.15em}
    \includegraphics[width=0.80\linewidth]{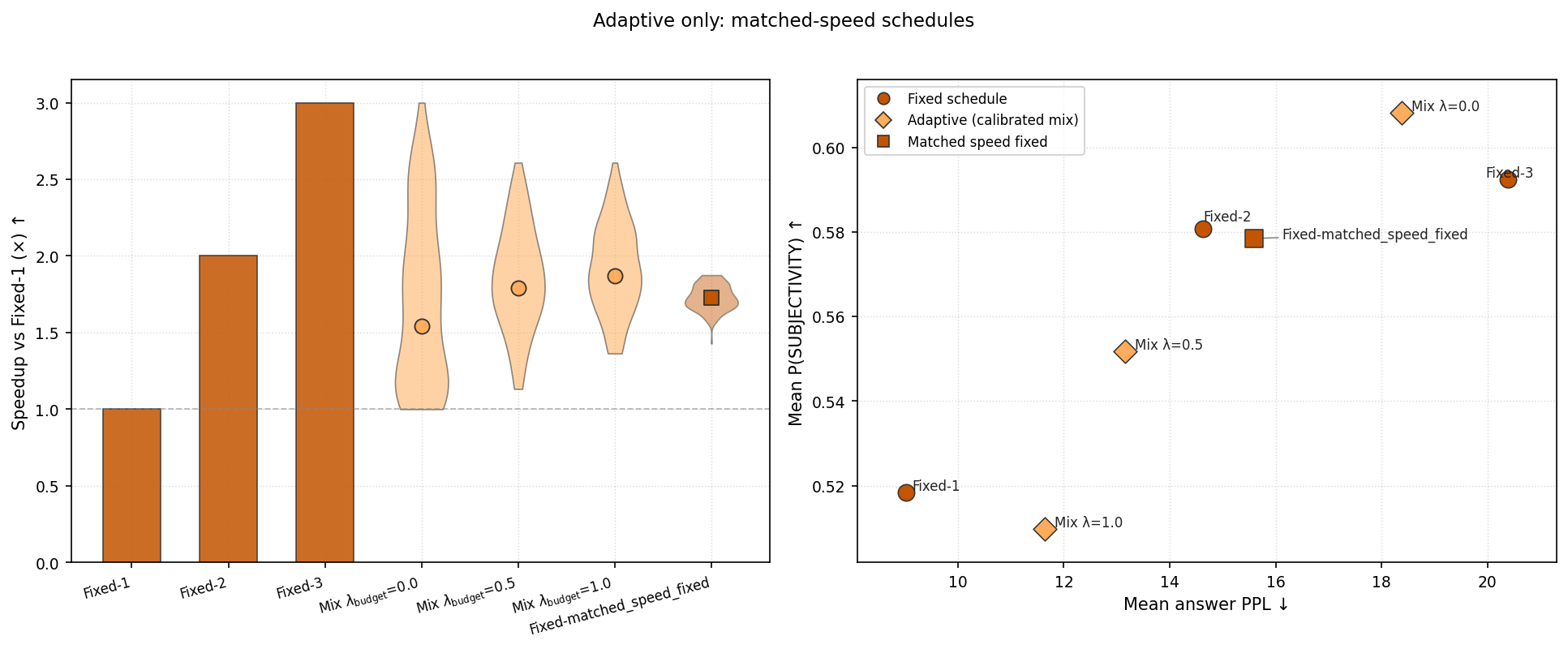}
    \vspace{0.45em}

    {\small\bfseries DiffMean}\par\vspace{0.15em}
    \includegraphics[width=0.80\linewidth]{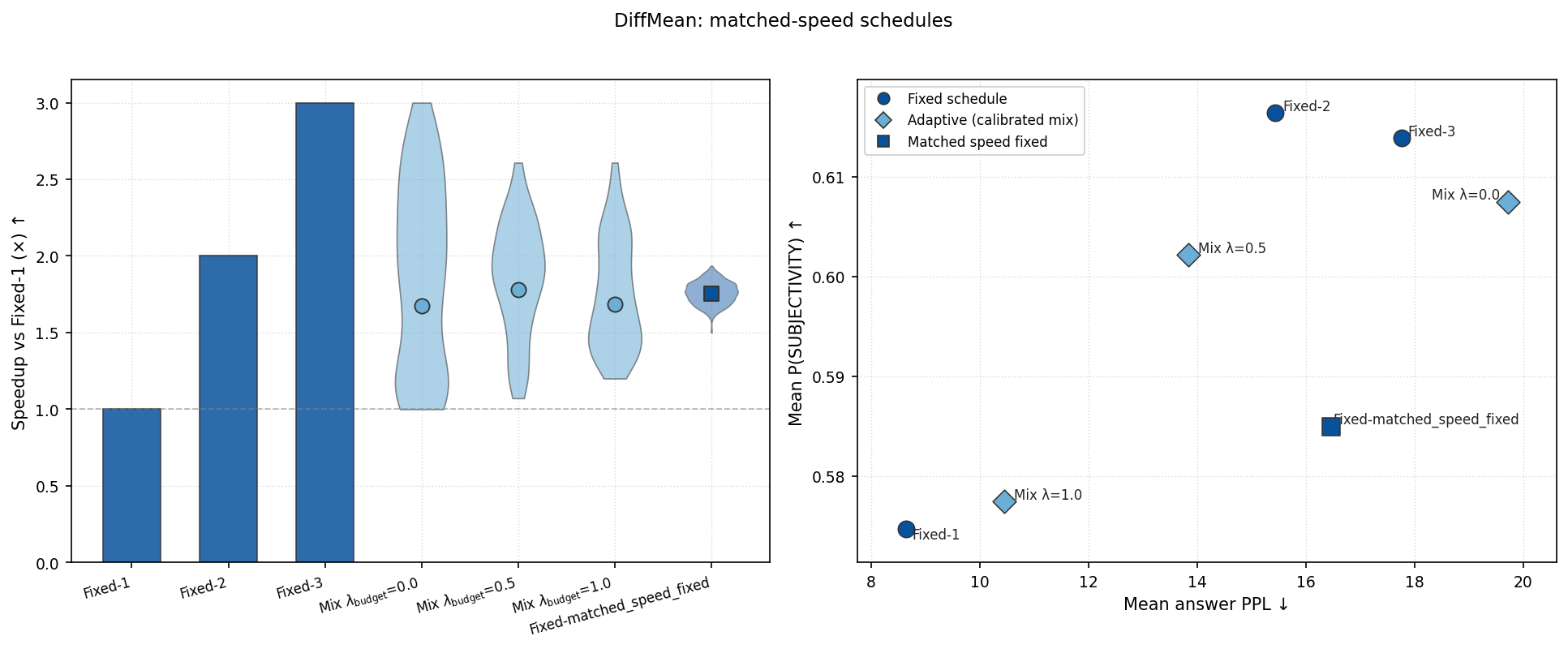}
    \vspace{0.45em}

    {\small\bfseries ACT}\par\vspace{0.15em}
    \includegraphics[width=0.80\linewidth]{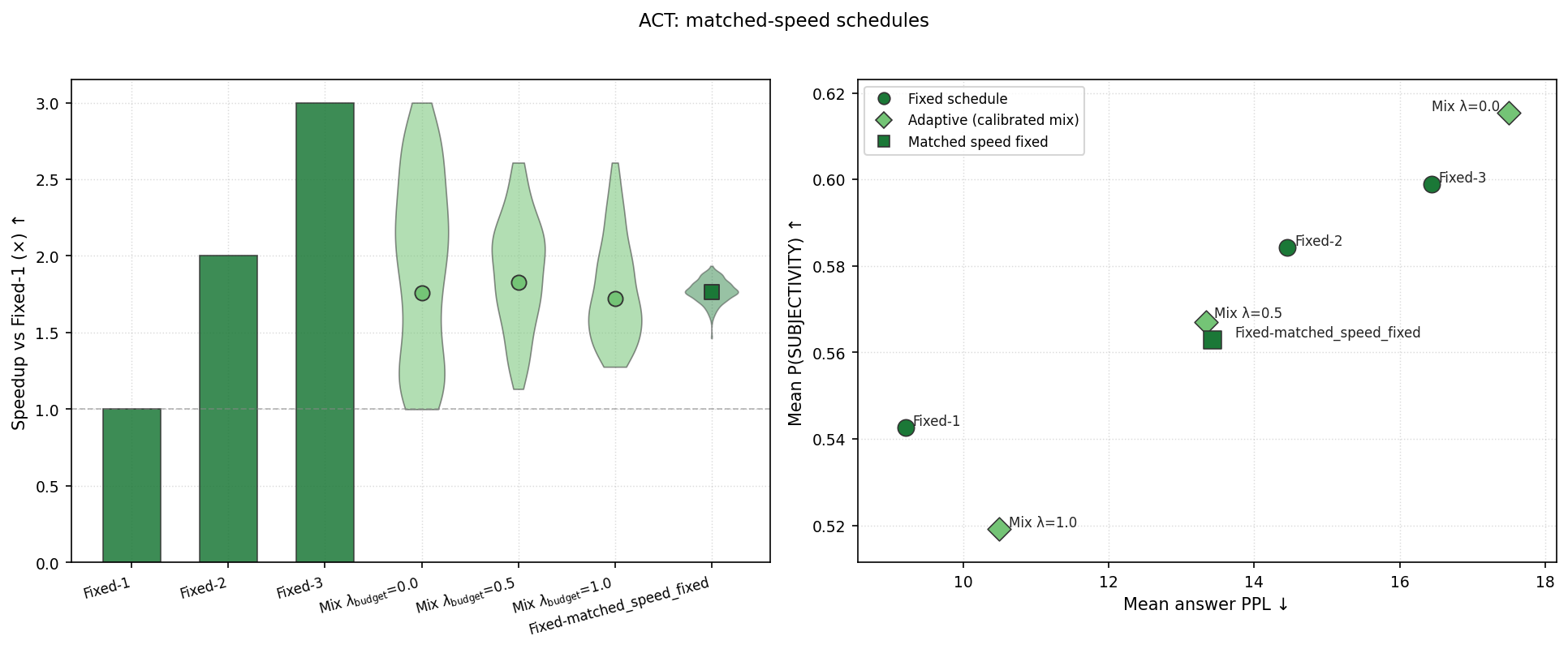}
    \vspace{0.45em}
    \caption{Adaptive versus compute-matched stochastic fixed schedules for LLaDA subjectivity under no activation steering, DiffMean, and ACT.}
    \label{fig:app-matched-objectivity}
\end{figure}
\FloatBarrier

\begin{figure}[H]
    \centering
    {\small\bfseries No activation}\par\vspace{0.15em}
    \includegraphics[width=0.80\linewidth]{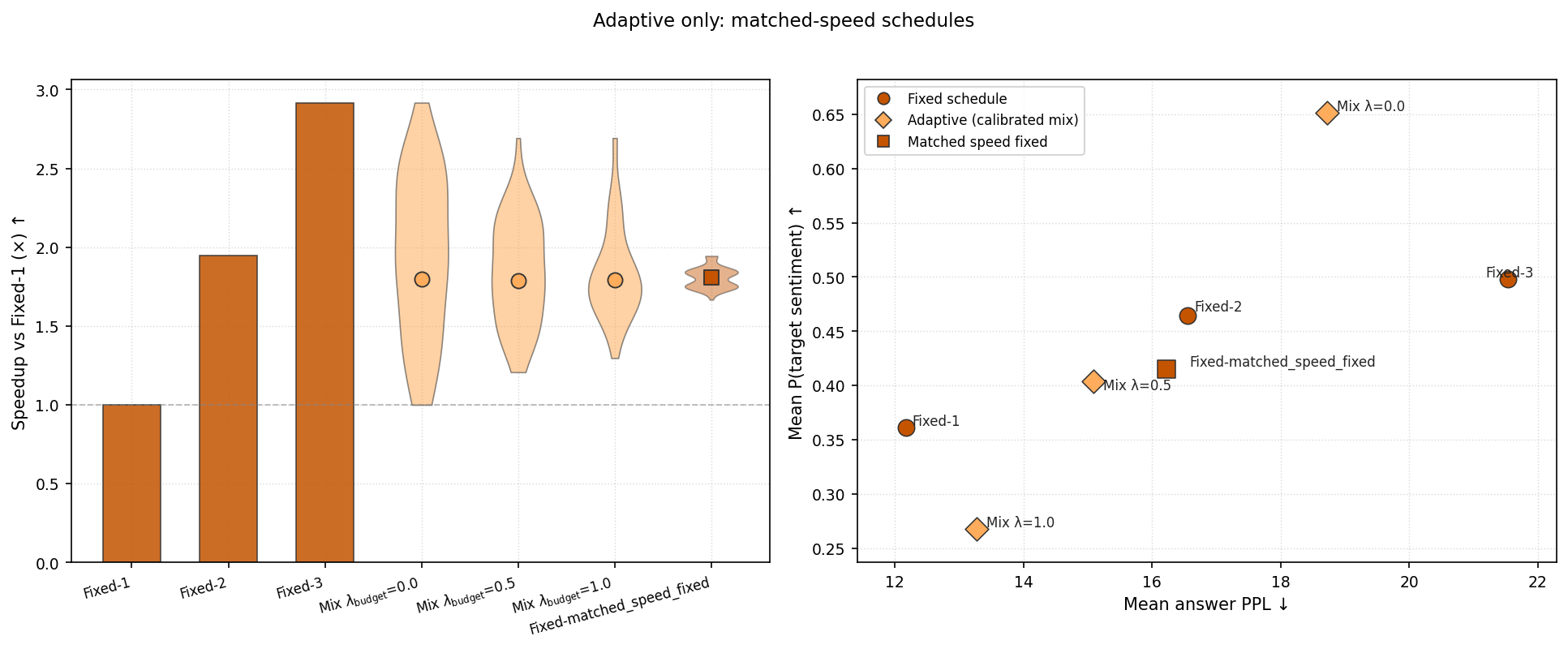}
    \vspace{0.45em}

    {\small\bfseries DiffMean}\par\vspace{0.15em}
    \includegraphics[width=0.80\linewidth]{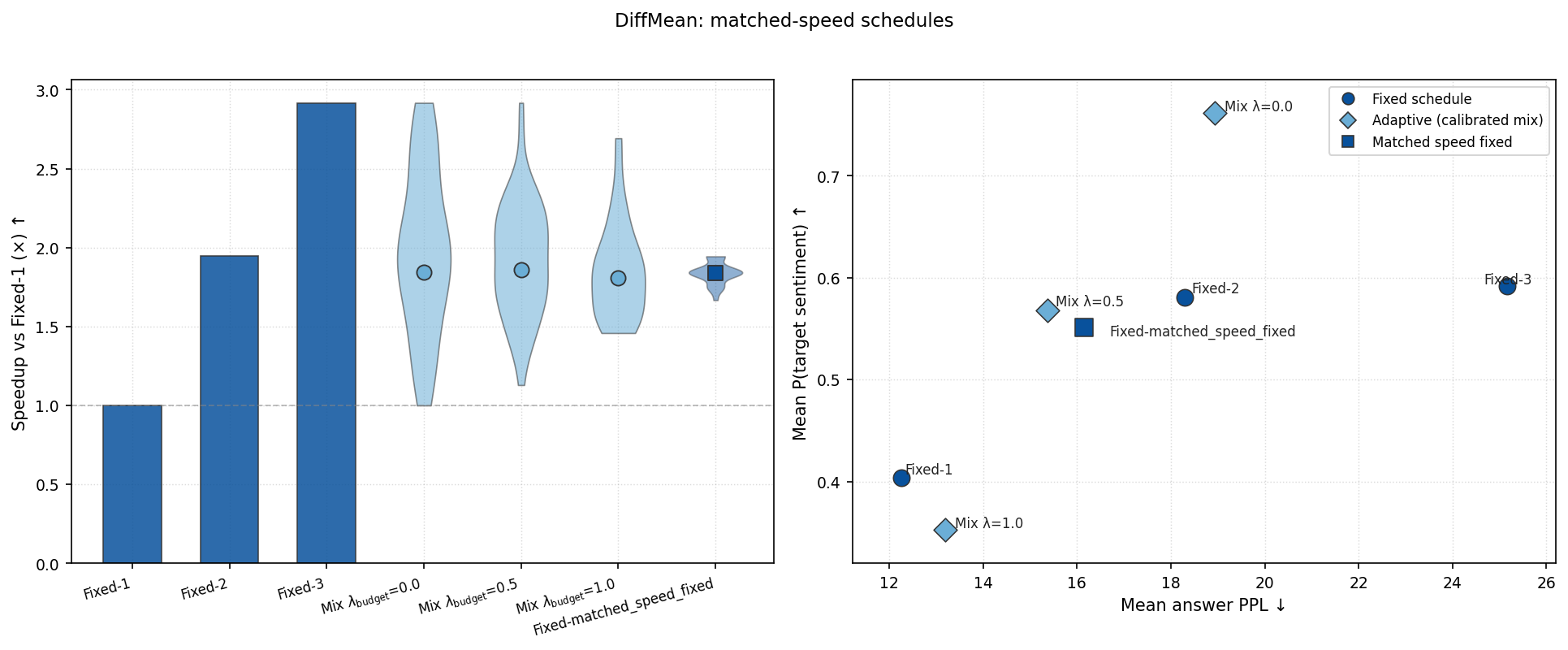}
    \vspace{0.45em}

    {\small\bfseries ACT}\par\vspace{0.15em}
    \includegraphics[width=0.80\linewidth]{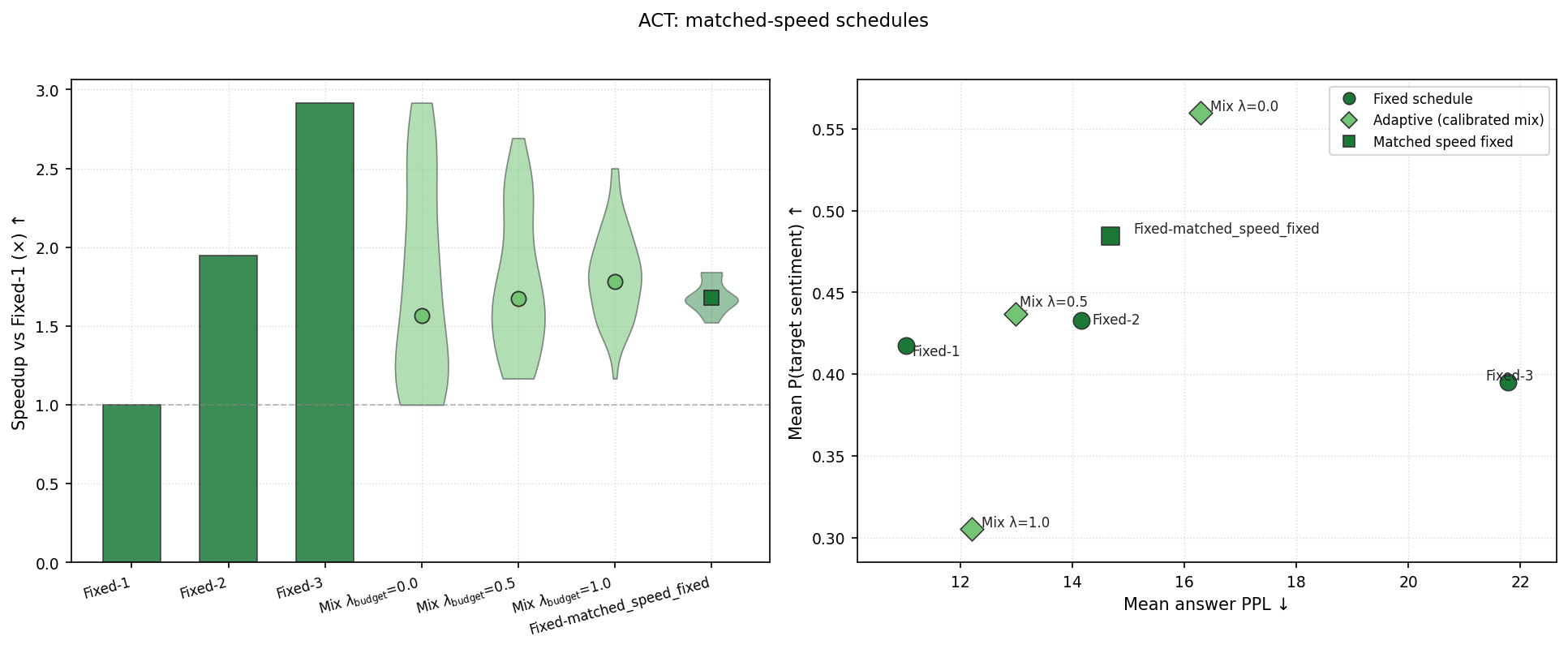}
    \vspace{0.45em}
    \caption{Adaptive versus compute-matched stochastic fixed schedules for LLaDA-V caption sentiment under no activation steering, DiffMean, and ACT.}
    \label{fig:app-matched-lladav-sentiment}
\end{figure}
\FloatBarrier

\begin{figure}[H]
    \centering
    \begin{minipage}{0.49\linewidth}
        \centering
        \includegraphics[width=\linewidth]{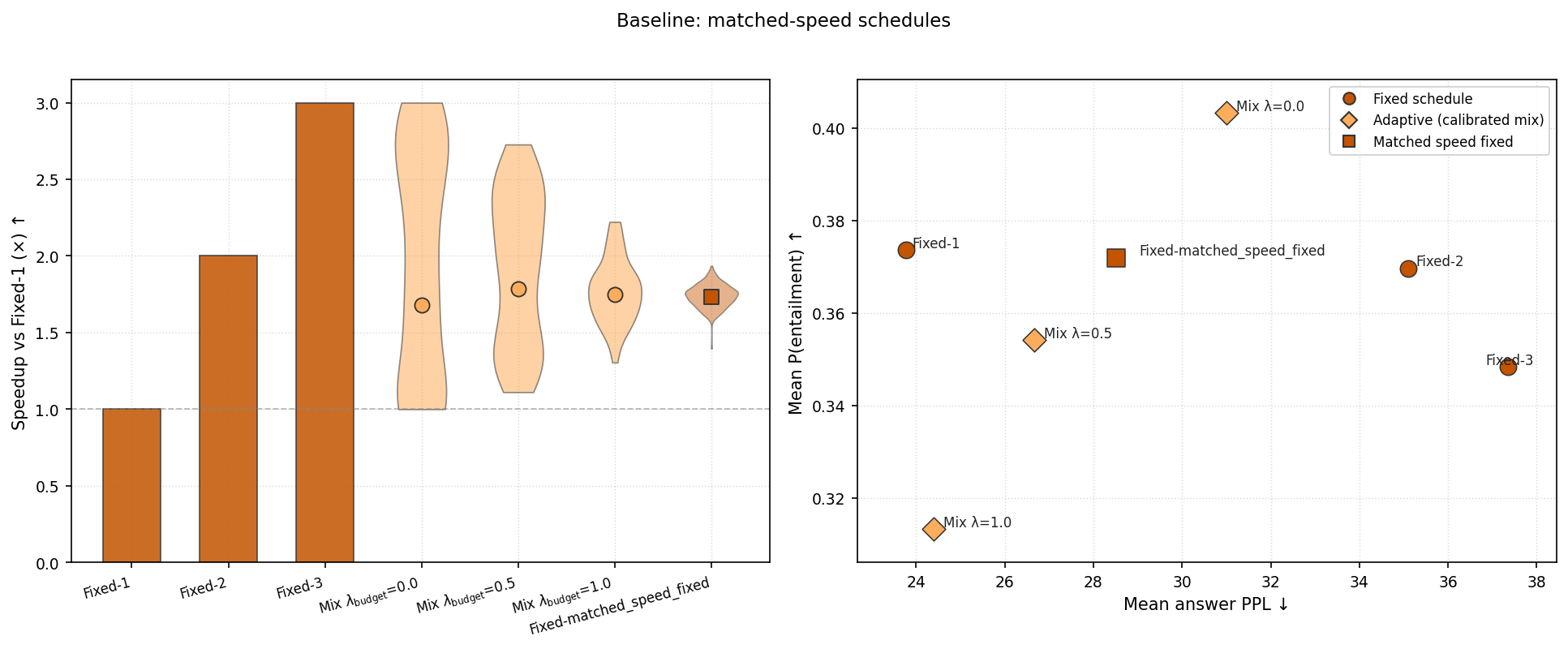}
    \end{minipage}\hfill
    \begin{minipage}{0.49\linewidth}
        \centering
        \includegraphics[width=\linewidth]{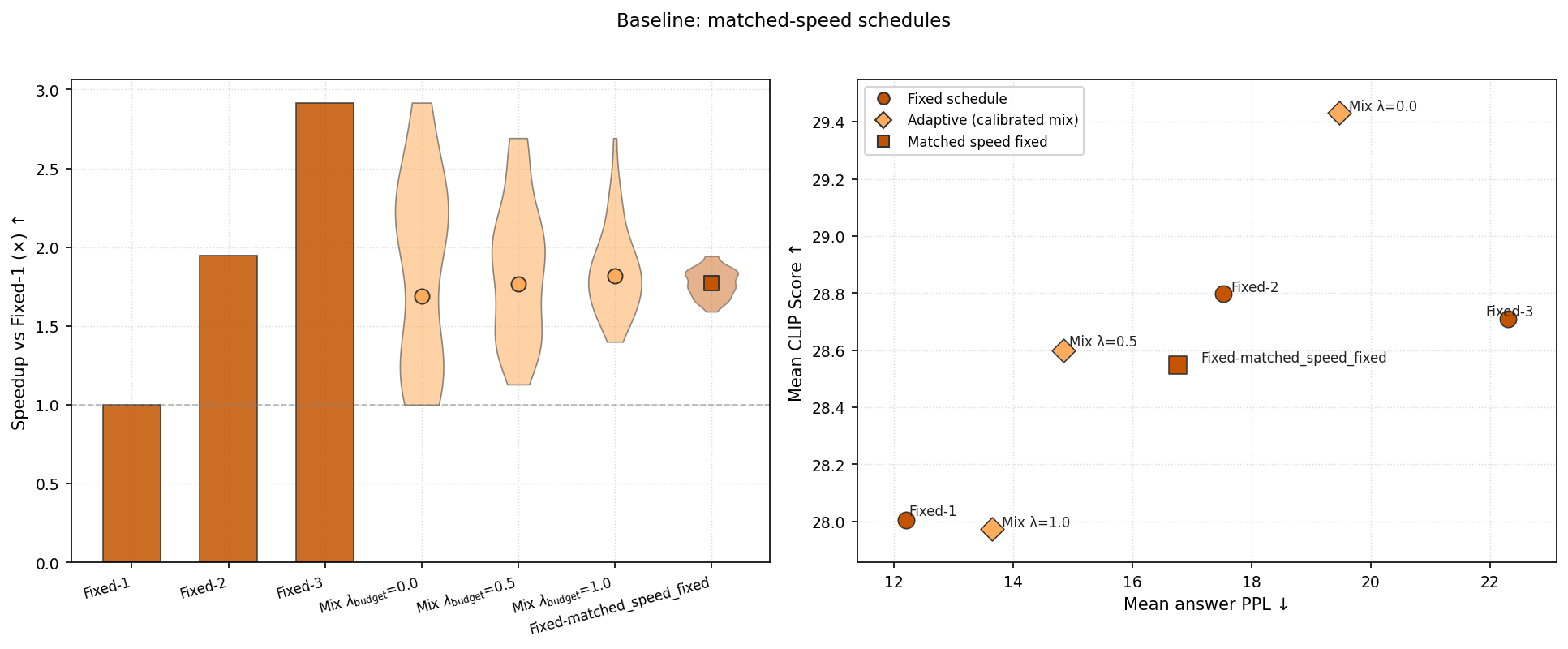}
    \end{minipage}
    \caption{Adaptive versus compute-matched stochastic fixed schedules for the NLI entailment and CLIPScore task-success objectives. Both settings use no activation steering.}
    \label{fig:app-matched-task-level}
\end{figure}
\FloatBarrier

\clearpage
\Needspace{0.42\textheight}
\subsection{Ordering stability under resampling: LLaDA sentiment}
\label{app:order-preservation}

Every entry in Appendices~\ref{app:counterfactual-tables}--\ref{app:matched-tables}
is a mean over a finite evaluation set, so an observed ordering could in
principle be an artifact of which examples happen to be in that set. We
therefore quantify how robust the reported orderings are to the choice of
evaluation examples, for the canonical setting used throughout the paper: LLaDA
sentiment. This is a depth check on one setting, complementary to the
held-out-scorer checks of Appendix~\ref{app:heldout-scorer}, which instead vary
the scorer while holding the evaluation set fixed; we do not repeat the
resampling analysis for the remaining targets, for the vision-language settings,
or for Dream-7B.

\paragraph{Procedure.}
We resample the evaluation examples with replacement, $10{,}000$ times, and
recompute all reported means within each resample. Every condition in a
comparison is evaluated on the same resampled example set, so the comparison
stays paired. We then report the fraction of resamples in which the ordering
observed on the full evaluation set is reproduced. No generation is repeated and
no decoding parameter changes; only the set of examples averaged over varies.
A value near $100\%$ indicates that the reported ordering does not depend on the
particular examples evaluated; a value near $50\%$ would indicate that it does.

\begin{table}[H]
\caption{\textbf{Ordering stability for LLaDA sentiment.} Means are shown at
full precision as $\lambda$ increases across the tested weights
($\lambda_{\mathrm{order}}=0.5\to0.75\to0.9\to1$ for commitment order,
$\lambda_{\mathrm{budget}}=0\to0.5\to1$ for adaptive budget), and correspond to
the rounded entries in Tables~\ref{tab:counterfactual-generalization}
and~\ref{tab:adaptive-generalization}. The last two columns give the fraction of
$10{,}000$ resamples of the evaluation set in which the observed ordering is
reproduced.}
\label{tab:order-preservation}
\centering
\scriptsize
\setlength{\tabcolsep}{3.4pt}
\resizebox{\linewidth}{!}{%
\begin{tabular}{@{}llccrr@{}}
\toprule
Control & Activation &
\makecell{Target means\\(increasing $\lambda$)} &
\makecell{PPL means\\(increasing $\lambda$)} &
\makecell{Target order\\preserved} &
\makecell{PPL order\\preserved} \\
\midrule

Commitment order & DiffMean
& $0.799\!\to\!0.615\!\to\!0.562\!\to\!0.500$
& $11.54\!\to\!11.05\!\to\!9.60\!\to\!8.88$
& $88.1\%$ & $83.7\%$ \\

Commitment order & ACT
& $0.751\!\to\!0.695\!\to\!0.625\!\to\!0.509$
& $11.23\!\to\!11.13\!\to\!10.08\!\to\!9.18$
& $91.4\%$ & $85.0\%$ \\

Commitment order & None
& $0.630\!\to\!0.533\!\to\!0.485\!\to\!0.350$
& $10.63\!\to\!10.47\!\to\!9.60\!\to\!9.46$
& $91.0\%$ & $77.5\%$ \\

\midrule

Adaptive budget & DiffMean
& $0.650\!\to\!0.549\!\to\!0.510$
& $21.32\!\to\!14.08\!\to\!12.92$
& $86.7\%$ & $95.9\%$ \\

Adaptive budget & ACT
& $0.653\!\to\!0.574\!\to\!0.516$
& $21.07\!\to\!15.60\!\to\!13.02$
& $91.0\%$ & $96.7\%$ \\

Adaptive budget & None
& $0.508\!\to\!0.390\!\to\!0.332$
& $18.68\!\to\!15.58\!\to\!12.48$
& $95.3\%$ & $97.4\%$ \\

\bottomrule
\end{tabular}}
\end{table}

The target ordering is reproduced in $86.7$--$95.3\%$ of resamples across all six
sweeps, so the direction of the effect---the property the paper's claims rest
on---is not an artifact of the particular $300$ examples evaluated. The PPL
ordering is reproduced somewhat less often for commitment order
($77.5$--$85.0\%$) than for the adaptive budget ($95.9$--$97.4\%$). This is
expected rather than anomalous: the commitment-order sweeps traverse a PPL range
of only about $1.2$--$2.7$ points, whereas the adaptive-budget sweeps traverse
$6$--$8$ points, and a smaller separation between adjacent means is harder to
resolve at a fixed evaluation-set size. The fluency cost of commitment order is
correspondingly small, and we read its four-point ordering as indicative rather
than firmly established.

\begin{table}[H]
\caption{\textbf{Stability of the compute-matched comparison, LLaDA sentiment.}
Each entry is the fraction of $10{,}000$ resamples in which the adaptive policy
is \emph{not} jointly dominated by its compute-matched stochastic schedule, that
is, in which it is not simultaneously worse in target and worse in perplexity.
The quantity is a floor rather than a measure of improvement: it can fall toward
zero when the adaptive policy is worse on both metrics at once, but a high value
indicates only that the adaptive point lies on a sensible trade-off curve
relative to the scorer-free control, not that adaptation buys a particular
amount.}
\label{tab:matched-stability}
\centering
\small
\setlength{\tabcolsep}{9pt}
\begin{tabular}{@{}lccc@{}}
\toprule
Activation &
$\lambda_{\mathrm{budget}}=0$ &
$\lambda_{\mathrm{budget}}=0.5$ &
$\lambda_{\mathrm{budget}}=1$ \\
\midrule
DiffMean      & $94.7\%$ & $99.9\%$ & $100.0\%$ \\
ACT           & $94.7\%$ & $92.4\%$ & $99.8\%$ \\
No activation & $99.9\%$ & $91.1\%$ & $100.0\%$ \\
\bottomrule
\end{tabular}
\end{table}

The adaptive point remains non-dominated in at least $91.1\%$ of resamples in
every configuration, so the ``competitive'' entries of
Table~\ref{tab:matched-stochastic-full} for this setting are not driven by a
small number of influential examples.

\clearpage
\subsection{Held-out-scorer checks}
\label{app:heldout-scorer}

Both commitment mechanisms query a frozen scorer during decoding and we then
report that same scorer's value. To check that the reported movement reflects a
change in the generated text rather than adaptation to one specific scorer, we
re-score identical generations with independent models. No generation is
repeated and no decoding parameter changes; only the scoring model changes. We
run this check for sentiment and NLI entailment on both model families, and for
CLIPScore on LLaDA-V. Sentiment generations are re-scored with
\texttt{cardiffnlp/twitter-roberta-base-sentiment-latest}, which differs from
the decoding scorer in architecture, pretraining corpus, fine-tuning data, and
label space; SAMSum summaries are re-scored with
\texttt{cross-encoder/nli-deberta-v3-large}, a substantially larger entailment
model than the decoding scorer. Captions are re-scored with
\texttt{laion/CLIP-ViT-B-32-laion2B-s34B-b79K}, which shares the ViT-B/32
architecture of the decoding scorer but is pretrained on a different corpus
(LAION-2B rather than WIT).

Absolute levels are not comparable across scorers. The held-out sentiment
classifier is three-class, so its $P(\mathrm{negative})$ is systematically lower
than that of the binary decoding scorer, because probability mass assigned to
\emph{neutral} is unavailable to the negative class. The held-out entailment
model likewise occupies a compressed and higher operating range than the
decoding scorer across all texts we have examined, generated or otherwise, so
its sweeps span a narrower interval. The held-out CLIP model is calibrated
differently again, and there the effect runs the other way: its commitment-order
sweep spans a wider interval than the decoding scorer's. The comparable quantity
in every case is therefore the ordering induced by the commitment weight, not
the magnitude of the movement.

\begin{table}[H]
\caption{\textbf{Held-out-scorer check, LLaDA sentiment.} The ordering induced
by $\lambda_{\mathrm{order}}$ and $\lambda_{\mathrm{budget}}$ is preserved in
all six sweeps.}
\label{tab:heldout-scorer}
\centering
\scriptsize
\setlength{\tabcolsep}{3.4pt}
\resizebox{\linewidth}{!}{%
\begin{tabular}{@{}llccc@{}}
\toprule
Sweep & Activation &
\makecell{Decoding scorer (SST-2)} &
\makecell{Held-out scorer (cardiffnlp)} &
\makecell{Ordering\\preserved} \\
\midrule
$\lambda_{\mathrm{order}}=0.5\to0.75\to0.9\to1$ & None
& $0.63\!\to\!0.53\!\to\!0.48\!\to\!0.35$
& $0.40\!\to\!0.38\!\to\!0.33\!\to\!0.26$ & \cmark \\
& DiffMean
& $0.80\!\to\!0.62\!\to\!0.56\!\to\!0.50$
& $0.57\!\to\!0.39\!\to\!0.38\!\to\!0.36$ & \cmark \\
& ACT
& $0.75\!\to\!0.69\!\to\!0.62\!\to\!0.51$
& $0.55\!\to\!0.52\!\to\!0.47\!\to\!0.41$ & \cmark \\
\midrule
$\lambda_{\mathrm{budget}}=0\to0.5\to1$ & None
& $0.51\!\to\!0.39\!\to\!0.33$
& $0.36\!\to\!0.28\!\to\!0.24$ & \cmark \\
& DiffMean
& $0.65\!\to\!0.55\!\to\!0.51$
& $0.42\!\to\!0.41\!\to\!0.38$ & \cmark \\
& ACT
& $0.65\!\to\!0.57\!\to\!0.52$
& $0.47\!\to\!0.45\!\to\!0.38$ & \cmark \\
\bottomrule
\end{tabular}}
\end{table}

\begin{table}[H]
\caption{\textbf{Held-out-scorer check, Dream-7B sentiment.} The same check
applied to the second model family. The ordering is preserved in all six
sweeps.}
\label{tab:heldout-scorer-dream}
\centering
\scriptsize
\setlength{\tabcolsep}{3.4pt}
\resizebox{\linewidth}{!}{%
\begin{tabular}{@{}llccc@{}}
\toprule
Sweep & Activation &
\makecell{Decoding scorer (SST-2)} &
\makecell{Held-out scorer (cardiffnlp)} &
\makecell{Ordering\\preserved} \\
\midrule
$\lambda_{\mathrm{order}}=0.5\to0.75\to0.9\to1$ & None
& $0.71\!\to\!0.52\!\to\!0.41\!\to\!0.30$
& $0.40\!\to\!0.32\!\to\!0.27\!\to\!0.22$ & \cmark \\
& DiffMean
& $0.80\!\to\!0.64\!\to\!0.54\!\to\!0.44$
& $0.59\!\to\!0.48\!\to\!0.43\!\to\!0.37$ & \cmark \\
& ACT
& $0.76\!\to\!0.65\!\to\!0.50\!\to\!0.42$
& $0.47\!\to\!0.41\!\to\!0.39\!\to\!0.34$ & \cmark \\
\midrule
$\lambda_{\mathrm{budget}}=0\to0.5\to1$ & None
& $0.50\!\to\!0.45\!\to\!0.43$
& $0.33\!\to\!0.30\!\to\!0.27$ & \cmark \\
& DiffMean
& $0.69\!\to\!0.64\!\to\!0.62$
& $0.54\!\to\!0.51\!\to\!0.49$ & \cmark \\
& ACT
& $0.64\!\to\!0.59\!\to\!0.52$
& $0.47\!\to\!0.44\!\to\!0.39$ & \cmark \\
\bottomrule
\end{tabular}}
\end{table}

\begin{table}[H]
\caption{\textbf{Held-out-scorer check, NLI entailment on both model families.}
The ordering induced by both commitment weights is preserved in all four
sweeps. The held-out model operates over a compressed range, so the sweeps span
a narrower interval than under the decoding scorer while retaining the same
direction.}
\label{tab:heldout-scorer-nli}
\centering
\scriptsize
\setlength{\tabcolsep}{3.4pt}
\resizebox{\linewidth}{!}{%
\begin{tabular}{@{}llccc@{}}
\toprule
Model & Sweep &
\makecell{Decoding scorer\\(nli-MiniLM2)} &
\makecell{Held-out scorer\\(nli-deberta-v3-large)} &
\makecell{Ordering\\preserved} \\
\midrule
LLaDA & $\lambda_{\mathrm{order}}=0.5\to0.75\to0.9\to1$
& $0.54\!\to\!0.43\!\to\!0.34\!\to\!0.31$
& $0.91\!\to\!0.88\!\to\!0.85\!\to\!0.84$ & \cmark \\
LLaDA & $\lambda_{\mathrm{budget}}=0\to0.5\to1$
& $0.40\!\to\!0.35\!\to\!0.31$
& $0.88\!\to\!0.85\!\to\!0.83$ & \cmark \\
\midrule
Dream-7B & $\lambda_{\mathrm{order}}=0.5\to0.75\to0.9\to1$
& $0.60\!\to\!0.47\!\to\!0.39\!\to\!0.34$
& $0.96\!\to\!0.91\!\to\!0.87\!\to\!0.85$ & \cmark \\
Dream-7B & $\lambda_{\mathrm{budget}}=0\to0.5\to1$
& $0.43\!\to\!0.34\!\to\!0.26$
& $0.89\!\to\!0.85\!\to\!0.82$ & \cmark \\
\bottomrule
\end{tabular}}
\end{table}

\begin{table}[H]
\caption{\textbf{Held-out-scorer check, LLaDA-V CLIPScore.} Both commitment
weights induce the same ordering under a CLIP model pretrained on a different
corpus. Absolute values are lower throughout because the two models are
calibrated differently; the held-out sweep also spans a wider interval than the
decoding scorer's.}
\label{tab:heldout-scorer-clip}
\centering
\scriptsize
\setlength{\tabcolsep}{3.4pt}
\resizebox{\linewidth}{!}{%
\begin{tabular}{@{}lccc@{}}
\toprule
Sweep &
\makecell{Decoding scorer\\(OpenAI ViT-B/32)} &
\makecell{Held-out scorer\\(LAION-2B ViT-B/32)} &
\makecell{Ordering\\preserved} \\
\midrule
$\lambda_{\mathrm{order}}=0.5\to0.75\to0.9\to1$
& $30.9\!\to\!26.8\!\to\!25.7\!\to\!25.5$
& $28.1\!\to\!22.9\!\to\!21.0\!\to\!20.3$ & \cmark \\
$\lambda_{\mathrm{budget}}=0\to0.5\to1$
& $29.4\!\to\!28.6\!\to\!28.0$
& $27.1\!\to\!25.7\!\to\!24.9$ & \cmark \\
\bottomrule
\end{tabular}}
\end{table}

The compute-matched comparison also survives the change of scorer. Under the
held-out classifier the adaptive point remains non-dominated by its matched
stochastic schedule in every corresponding comparison, on both model families
and for both the sentiment and NLI settings.

\subsection{Invalid-rate diagnostics}
\label{app:invalid-diagnostics}

We treat invalid rate as a supporting diagnostic rather than a primary
evaluation metric. An output is flagged as invalid when its reference-model
perplexity exceeds $\tau_{\mathrm{PPL}}=100$
(Appendix~\ref{app:experimental_details}). Invalid rate therefore provides a
coarse thresholded view of the same fluency degradation measured continuously by
PPL.

For counterfactual commitment-order experiments, invalid rates do not exhibit a
consistent monotonic relationship with $\lambda_{\mathrm{order}}$ across
settings. This contrasts with the clear ordering observed for the target score
and mean PPL, and we therefore do not interpret invalid-rate variation as an
additional effect of commitment order. For adaptive-budget experiments,
invalid-rate behavior broadly follows the corresponding PPL trends: within
adaptive sweeps, configurations with higher mean PPL generally also exhibit
higher invalid rates, and the same relationship holds in the compute-matched
comparisons. In several settings, particularly LLaDA-V captioning and the
Dream-7B experiments, invalid generations are rare or absent, further limiting
the informativeness of this diagnostic. We therefore report invalid-rate
behavior qualitatively; the continuous PPL results provide the more informative
characterization of generation quality.

\subsection{Qualitative examples}
\label{app:qualitative}

We provide representative generations to illustrate what different values of the
frozen target scores correspond to qualitatively. These examples are
illustrative and are not used for quantitative evaluation.

\paragraph{Sentiment and subjectivity.}
Table~\ref{tab:qualitative-text} shows examples for the two text control
objectives. For sentiment, increasing target score corresponds to increasingly
negative language. For subjectivity, higher scores correspond to increasingly
evaluative or emotive language.

\begin{table}[H]
\caption{\textbf{Qualitative examples for the text control objectives.}
Scores denote the corresponding frozen target-class probability:
negative sentiment or subjectivity.}
\label{tab:qualitative-text}
\centering
\scriptsize
\setlength{\tabcolsep}{4pt}
\begin{tabular}{lllp{0.61\linewidth}}
\toprule
Target & Level & Score & Generated continuation \\
\midrule
Sentiment
& Low
& 0.000
& \textit{``The author is a master of the genre, and the book is a
masterclass in comedic writing. The book is a must-read for anyone who loves
a good laugh \ldots''} \\

& Mid.
& 0.531
& \textit{``\ldots it didn't work. I got the second one and it didn't work
either. I don't know if it's the bread machine or the seller, but I'm not
going to waste my money on this particular bread machine again.''} \\

& High
& 1.000
& \textit{``This movie was a waste of time. The plot was weak, the acting
was bad, and the special effects were terrible. Don't waste your time or your
money on this movie.''} \\
\midrule

Subjectivity
& Low
& 0.248
& \textit{``\ldots message through the use of visuals, sound, music,
dialogue, acting, directing, writing, producing, camera, lighting, editing,
sound, makeup, costumes, props, location, casting, financing, marketing,
distribution, and more.''} \\

& Mid.
& 0.665
& \textit{``However, it's important to remember that the use of such methods
is highly controversial and often associated with ethical concerns. It's
crucial to approach these topics with caution and respect \ldots''} \\

& High
& 0.801
& \textit{``\ldots ancient power and mystery. \ldots they must use their
own wits and courage to uncover the true secrets of the temple.''} \\
\bottomrule
\end{tabular}
\end{table}

\begin{figure}[H]
\centering

\begin{minipage}[t]{0.32\linewidth}
\centering
\includegraphics[width=\linewidth]{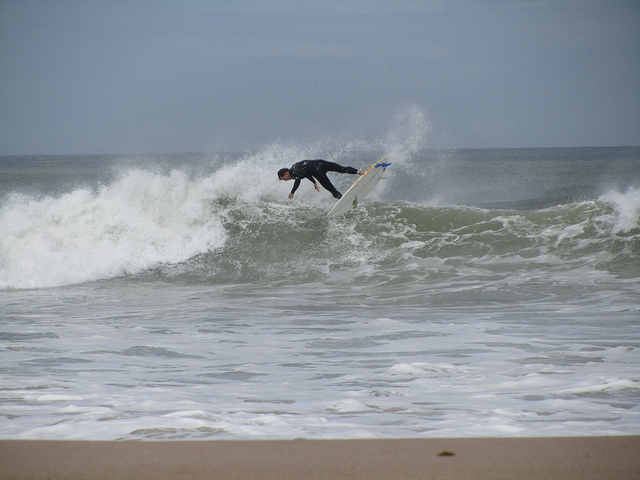}
\scriptsize

\textbf{Low: 0.002}

\textit{``A surfer is skillfully riding a wave on a surfboard. The surfboard
is positioned on the crest of the wave, closer to the center of the image.''}
\end{minipage}
\hfill
\begin{minipage}[t]{0.32\linewidth}
\centering
\includegraphics[width=\linewidth]{images/pareto/bear.jpg}
\scriptsize

\textbf{Intermediate: 0.457}

\textit{``A bear is present in the image, positioned towards the center-right.
It appears to be walking across the path, moving from the left to the right of
the frame.''}
\end{minipage}
\hfill
\begin{minipage}[t]{0.32\linewidth}
\centering
\includegraphics[width=\linewidth]{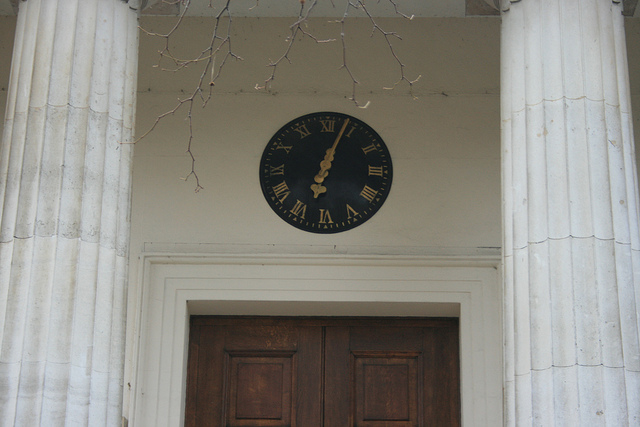}
\scriptsize

\textbf{High: 0.9996}

\textit{``There's a small, insignificant, dry, dirty, and possibly dead plant,
growing on the wall near the base of the column on the left side of the
image.''}
\end{minipage}

\caption{\textbf{Qualitative caption-sentiment examples.}
Images and generated captions are shown at increasing frozen negative-sentiment
scores.}
\label{fig:qualitative-caption-sentiment}
\end{figure}

\paragraph{Caption sentiment.}
Figure~\ref{fig:qualitative-caption-sentiment} shows low-, intermediate-, and
high-scoring caption-sentiment examples. The low-scoring caption contains
positive evaluative language (``skillfully''), the intermediate example is
largely neutral, and the high-scoring example accumulates negative evaluative
modifiers while still describing an element that is present in the image. This
is why we report CLIPScore alongside caption sentiment throughout: it verifies
that captions continue to describe their images as their sentiment changes, and
across the caption-sentiment sweeps image--text alignment does not degrade
substantially (Fig.~\ref{fig:app-adaptive-lladav-sentiment}).

\paragraph{NLI entailment.}
The examples below illustrate how the entailment score varies with factual
consistency between the source dialogue and generated summary.

\begin{table}[H]
\caption{\textbf{Qualitative NLI-entailment examples.}
Higher scores correspond to stronger agreement between the generated summary
and its source dialogue. Profanity in the source dialogue is masked.}
\label{tab:qualitative-nli}
\centering
\scriptsize
\setlength{\tabcolsep}{4pt}
\begin{tabular}{llp{0.72\linewidth}}
\toprule
Level & Score & Dialogue and generated summary \\
\midrule

Low & 0.127 &
\textbf{Dialogue:} \textit{Joshua:} look out the window ~/~ \textit{Noah:}
what's there? XD ~/~ \textit{Joshua:} do it :P ~/~ \textit{Noah:} oh wow ~/~
\textit{Noah:} so much snow :D when did this happen lol ~/~ \textit{Joshua:}
:D ~/~ \textit{Noah:} it's definitely time to build a snowman! ~/~
\textit{Joshua:} build an Olaf :d ~/~ \textit{Noah:} ofc i will, he's my
idol! ~/~ \textit{Joshua:} XD

\textbf{Summary:} \textit{``Joshua and Noah are excited about the unexpected
snow outside and plan to build a snowman, with Joshua suggesting they create
an Olaf-themed snowman and Noah agreeing, as he admires fan favorite Olaf
from the Wreck-It Ralph movie.''} The final clause is a hallucination: Olaf is
not from that film, and the dialogue says nothing about its origin. \\[2pt]

Intermediate & 0.487 &
\textbf{Dialogue:} \textit{Max:} I'm so sorry Lucas. I don't know what got
into me. ~/~ \textit{Lucas:} \ldots{} ~/~ \textit{Lucas:} I don't know either.
~/~ \textit{Mason:} that was really f***ed up Max ~/~ \textit{Max:} I know.
I'm so sorry :(. ~/~ \textit{Lucas:} I don't know, man. ~/~ \textit{Mason:}
what were you thinking?? ~/~ \textit{Max:} I wasn't. ~/~ \textit{Max:} Can we
please meet and talk this through? Please. ~/~ \textit{Lucas:} Ok. I'll think
about it and let you know. ~/~ \textit{Max:} Thanks\ldots

\textbf{Summary:} \textit{``Max apologizes to Lucas and Mason, expressing
confusion over his actions, while Lucas and Mason also express confusion and
concern, ultimately agreeing to meet and discuss the situation, with Lucas
promising to think about it and let Max know when he's ready.''} The gist is
correct, but Max apologizes to Lucas rather than to both, Mason expresses
judgement rather than confusion, and Lucas agrees only to consider meeting. \\[2pt]

High & 0.947 &
\textbf{Dialogue:} \textit{Mike:} hey have you played RDR2 already? ~/~
\textit{Jack:} of course ~/~ \textit{Jack:} I think I'm halfway through ~/~
\textit{Mike:} ooooh ~/~ \textit{Mike:} and whaddya think? ~/~ \textit{Mike:}
coz I've just started ~/~ \textit{Jack:} what can I say, it's freaking great
~/~ \textit{Jack:} the storytelling, the acting, the graphics, immersiveness
(is that a word? hehe) ~/~ \textit{Jack:} you know, when I was a kid, I used to
play with cowboy-themed lego sets ~/~ \textit{Jack:} so it's like a dream
coming true for me :D ~/~ \textit{Mike:} haha ~/~ \textit{Mike:} totally

\textbf{Summary:} \textit{``Jack, who has already played RDR2 and is halfway
through, expresses his immense enthusiasm for the game, praising its
storytelling, acting, graphics, and immersiveness, and likens it to his
childhood fascination with cowboy-themed lego sets.''} Every proposition is
supported by the dialogue, including the specific list of praised attributes
and the childhood comparison. \\

\bottomrule
\end{tabular}
\end{table}

\paragraph{Image--text alignment.}
Finally, Figure~\ref{fig:qualitative-clip} illustrates low, intermediate, and
high CLIPScore generations. Higher scores qualitatively correspond to better
agreement between the generated caption and the visual content.

\begin{figure}[H]
\centering

\begin{minipage}[t]{0.32\linewidth}
\centering
\includegraphics[width=\linewidth]{images/pareto/frisbee.jpg}
\scriptsize

\textbf{Low: 25.59}

\textit{``A frisbee is lying on the ground \ldots Nearby, there's a cellular
telephone positioned closer to the center of the image.''}
\end{minipage}
\hfill
\begin{minipage}[t]{0.32\linewidth}
\centering
\includegraphics[width=\linewidth]{images/pareto/beach.jpg}
\scriptsize

\textbf{Intermediate: 29.47}

\textit{``A person is wearing a backpack, positioned towards the right side
of the image. There are two surfboards on the beach \ldots''}
\end{minipage}
\hfill
\begin{minipage}[t]{0.32\linewidth}
\centering
\includegraphics[width=\linewidth]{images/pareto/children.jpg}
\scriptsize

\textbf{High: 33.62}

\textit{``The image depicts a group of children in a classroom, posing for a
picture. They are seated on benches in rows, with a brick wall in the
background.''}
\end{minipage}

\caption{\textbf{Qualitative CLIPScore examples.}
Generated captions at increasing image--text alignment scores.}
\label{fig:qualitative-clip}
\end{figure}
\FloatBarrier
\clearpage
\section{Dream-7B Results}
\label{app:dream}

This appendix reports the second-family results in full: sentiment control under
all three activation conditions, and NLI entailment on SAMSum without activation
steering. Dream-7B is a masked diffusion language model initialized from
autoregressive weights rather than trained from scratch with random masking
\citep{dream2025}, so it differs from LLaDA in training lineage, tokenizer,
depth, and confidence calibration while retaining the
provisional-prediction/commitment split the methods act on.

\subsection{Setup}
\label{app:dream-setup}

We use \texttt{Dream-org/Dream-v0-Instruct-7B} with generation length $60$,
temperature $0$, seed $42$, and $N=300$ evaluation prompts per setting, matching
the corresponding LLaDA configurations. Prompts are the product-review
continuation and SAMSum summarization templates of
Table~\ref{tab:prompt_templates}, unchanged. The commitment configuration is
unchanged: $u_t\equiv1$ with
$\lambda_{\mathrm{order}}\in\{0.5,0.75,0.9,1.0\}$ for commitment order,
confidence ranking with $U_{\max}=3$ and
$\lambda_{\mathrm{budget}}\in\{0,0.5,1\}$ for the adaptive budget, and the same
compute-matched stochastic fixed schedule construction
(Appendix~\ref{app:matched-schedule-details}).

Three quantities are re-derived rather than transferred, because they are
properties of the denoiser rather than of the method.

\begin{itemize}
\item \textbf{Steering window.} LLaDA-8B has $32$ layers and is steered over
$\ell\in\{9,\ldots,24\}$, a fractional depth of $0.28$--$0.75$. Dream-7B has
$28$ layers, so the matched window is $\ell\in\{8,\ldots,21\}$, a fractional
depth of $0.29$--$0.75$. We match fractional depth rather than absolute layer
index.

\item \textbf{Intervention strengths.} We use $\alpha=0.7$ for DiffMean and
$\beta=0.4$ for ACT, held fixed across every commitment configuration
(Table~\ref{tab:steering_strengths}).

\item \textbf{Adaptive calibration.} The phase-specific empirical CDFs of
Appendix~\ref{app:adaptive-details} are recomputed from a Dream fixed-$2$ run
for each setting.
\end{itemize}

The steering vectors and ACT statistics are rebuilt from the same Amazon
Polarity pole sets with Dream's tokenizer and hidden states
(Table~\ref{tab:vector_construction}). The counterfactual span-deletion
procedure is tokenizer-agnostic---it maps each provisional token to a character
span in the decoded completion---and required no modification. The NLI setting
uses no activation intervention, so it requires neither steering vectors nor a
layer window.

\subsection{Sentiment: commitment order}

\begin{table}[H]
\caption{\textbf{Counterfactual commitment-order sequences, Dream-7B sentiment.}
Both the target score and PPL are non-increasing across the four tested weights
under all three activation conditions.}
\label{tab:dream-order}
\centering
\scriptsize
\setlength{\tabcolsep}{3.2pt}
\resizebox{\linewidth}{!}{%
\begin{tabular}{@{}lccc@{}}
\toprule
Activation &
\makecell{Target\\
$\lambda_{\mathrm{order}}=0.5\to0.75\to0.9\to1$} &
\makecell{PPL\\
$\lambda_{\mathrm{order}}=0.5\to0.75\to0.9\to1$} &
\makecell{Ordering consistent\\target / PPL} \\
\midrule

None
& $0.705\!\to\!0.525\!\to\!0.406\!\to\!0.297$
& $7.98\!\to\!7.79\!\to\!7.44\!\to\!7.04$
& \cmark / \cmark \\

DiffMean
& $0.796\!\to\!0.637\!\to\!0.539\!\to\!0.439$
& $7.85\!\to\!7.00\!\to\!6.73\!\to\!6.40$
& \cmark / \cmark \\

ACT
& $0.764\!\to\!0.651\!\to\!0.498\!\to\!0.423$
& $8.44\!\to\!7.69\!\to\!7.58\!\to\!6.95$
& \cmark / \cmark \\

\bottomrule
\end{tabular}}
\end{table}

The unsteered condition traverses a wider target range than its LLaDA
counterpart, from $0.297$ at confidence-only commitment to $0.705$ at
$\lambda_{\mathrm{order}}=0.5$, against $0.35$ to $0.63$ for LLaDA, and does so
at a smaller perplexity cost: about $0.9$ points on Dream against about $1.1$
points on LLaDA, from a lower base. We do not have a controlled explanation for
the difference in magnitude, and we do not claim commitment control is more
effective on autoregressively initialized denoisers in general; the relevant
observation is that the direction and monotonicity are unchanged.

\begin{figure}[H]
    \centering
    \includegraphics[width=0.76\linewidth]{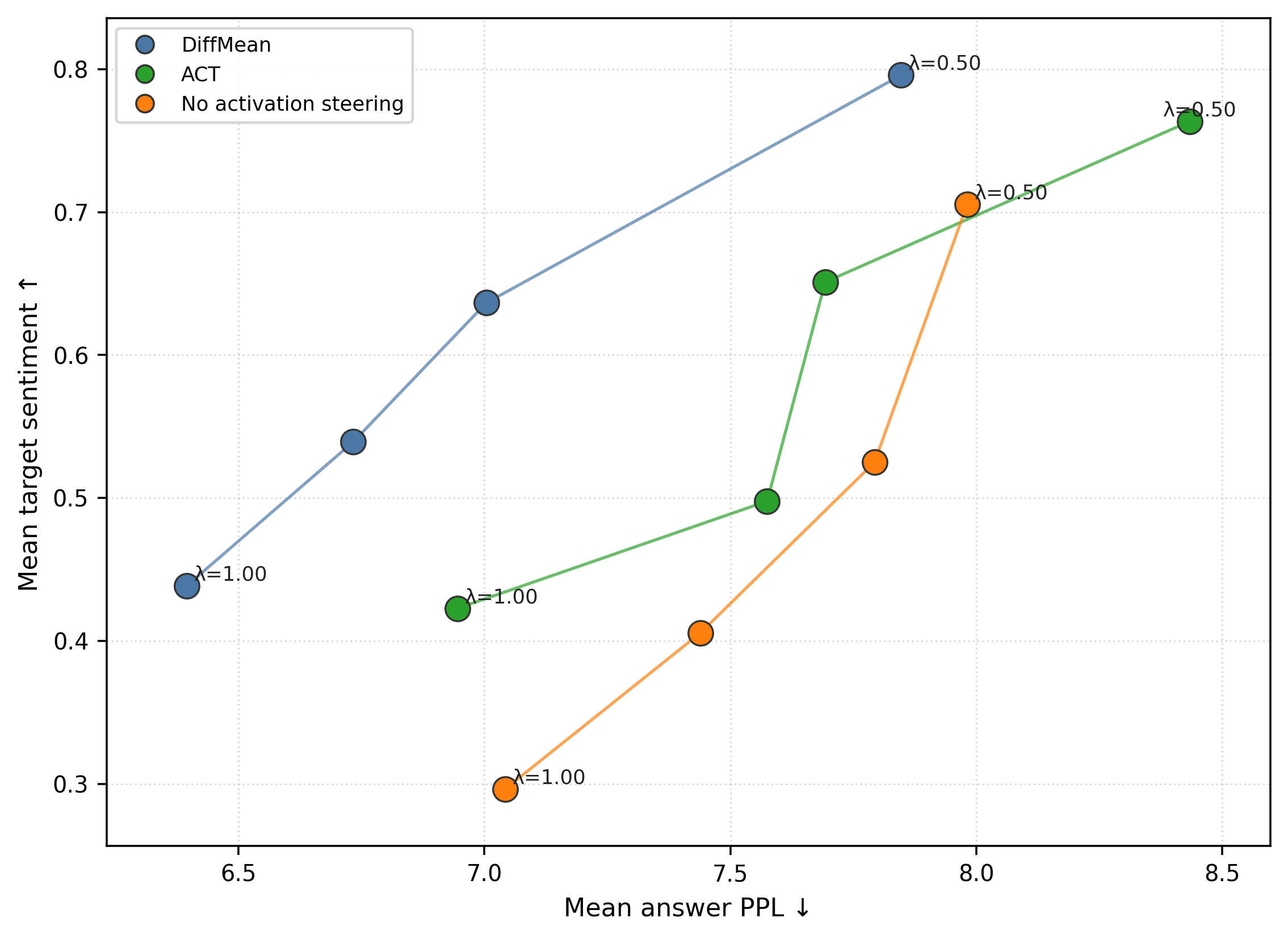}
    \caption{Counterfactual commitment-order trade-offs for Dream-7B sentiment.
    As in Fig.~\ref{fig:counterfactual-sentiment}, $u_t\equiv1$ is fixed, so
    movement along each curve changes only which provisional token is committed
    next.}
    \label{fig:dream-counterfactual}
\end{figure}
\FloatBarrier

\subsection{Sentiment: adaptive budgets}

\begin{table}[H]
\caption{\textbf{Adaptive-budget sequences, Dream-7B sentiment.}
The scoring map is fixed to confidence, so movement across each sequence
changes only the commitment budget. Target-progress-driven budgeting
($\lambda_{\mathrm{budget}}=0$) gives the strongest target and
confidence-driven budgeting ($\lambda_{\mathrm{budget}}=1$) the lowest PPL.}
\label{tab:dream-budget}
\centering
\scriptsize
\setlength{\tabcolsep}{3.4pt}
\resizebox{\linewidth}{!}{%
\begin{tabular}{@{}lccc@{}}
\toprule
Activation &
\makecell{Target\\
$\lambda_{\mathrm{budget}}=0\to0.5\to1$} &
\makecell{PPL\\
$\lambda_{\mathrm{budget}}=0\to0.5\to1$} &
\makecell{Ordering consistent\\target / PPL} \\
\midrule

None
& $0.501\!\to\!0.453\!\to\!0.432$
& $13.76\!\to\!11.45\!\to\!9.34$
& \cmark / \cmark \\

DiffMean
& $0.685\!\to\!0.638\!\to\!0.621$
& $11.33\!\to\!9.66\!\to\!8.44$
& \cmark / \cmark \\

ACT
& $0.645\!\to\!0.588\!\to\!0.523$
& $12.88\!\to\!10.46\!\to\!8.70$
& \cmark / \cmark \\

\bottomrule
\end{tabular}}
\end{table}

\begin{figure}[H]
    \centering
    {\small\bfseries No activation}\par\vspace{0.15em}
    \includegraphics[width=0.80\linewidth]{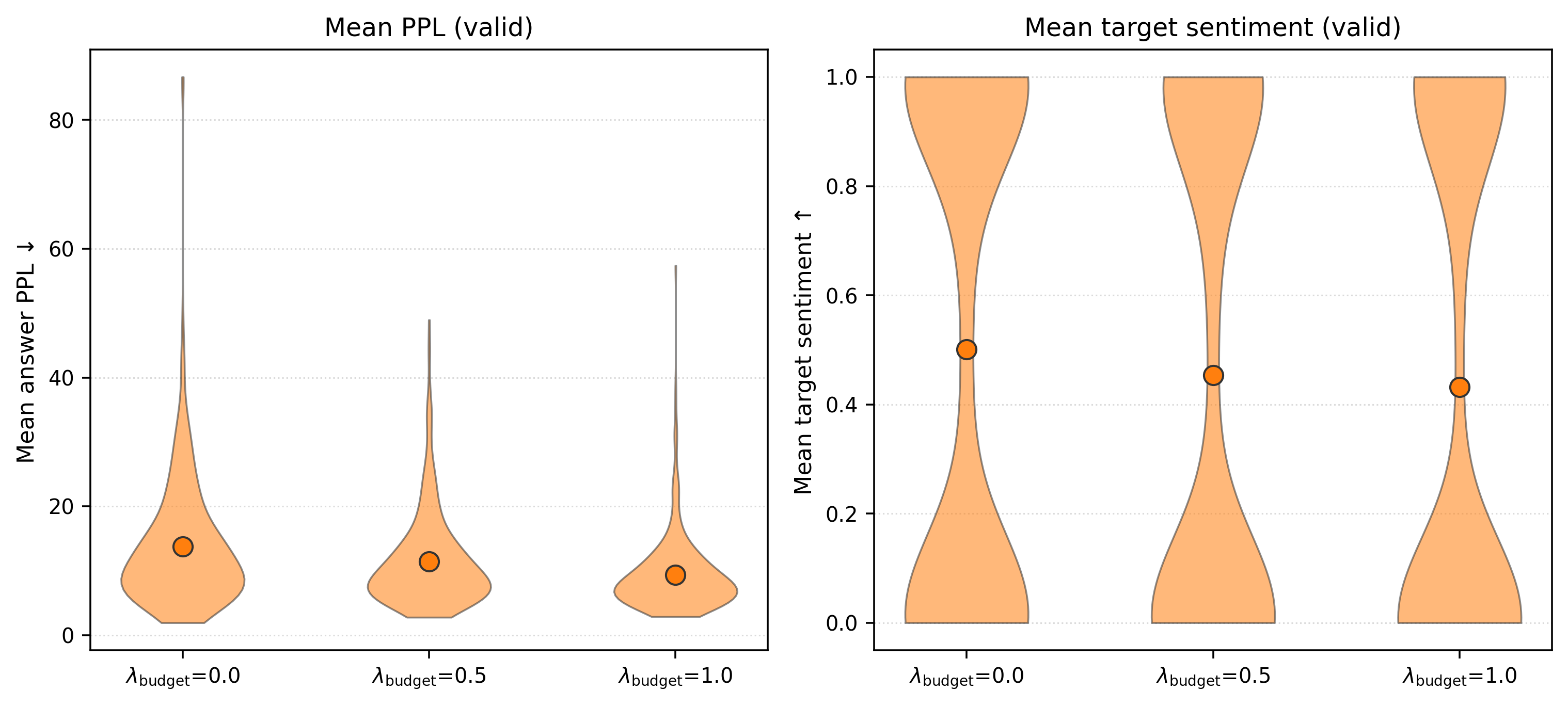}
    \vspace{0.45em}

    {\small\bfseries DiffMean}\par\vspace{0.15em}
    \includegraphics[width=0.80\linewidth]{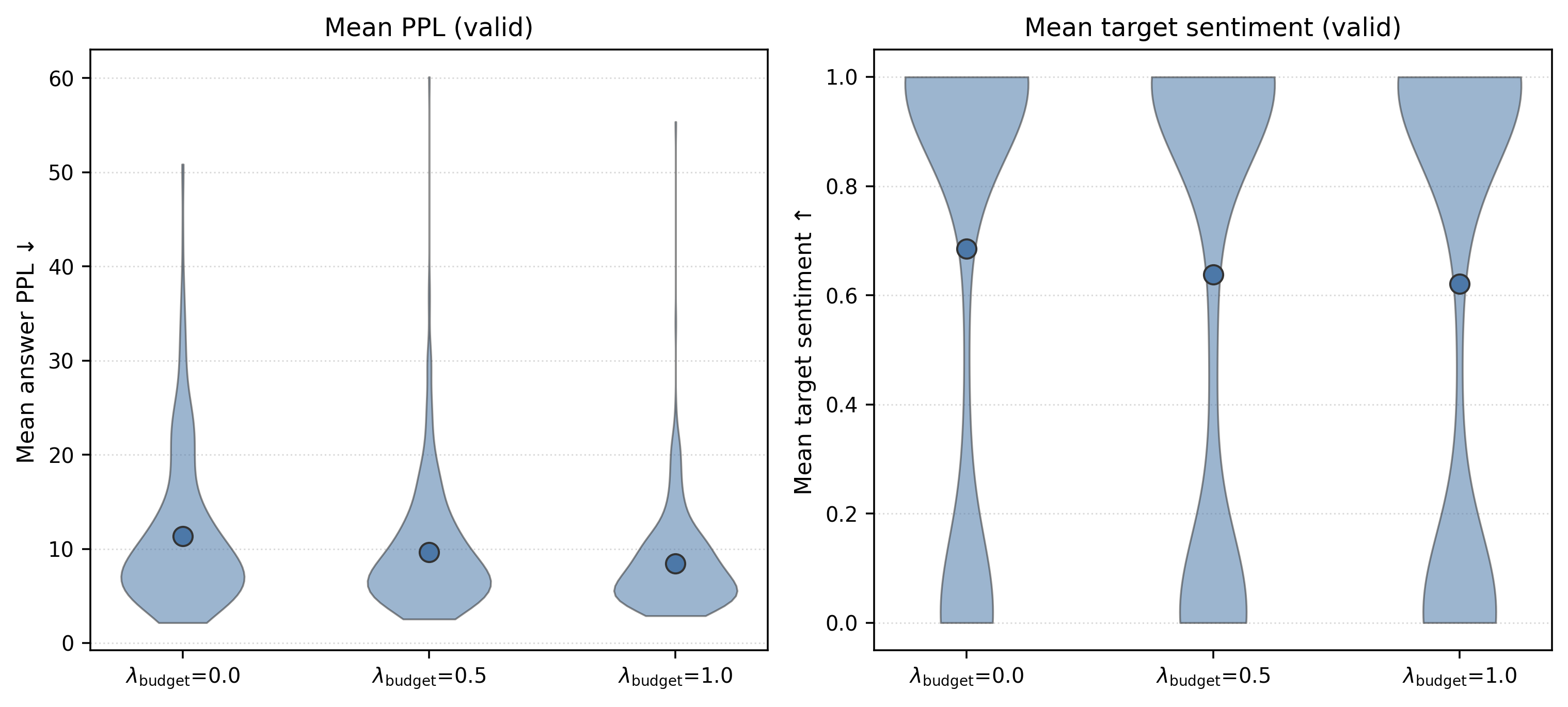}
    \vspace{0.45em}

    {\small\bfseries ACT}\par\vspace{0.15em}
    \includegraphics[width=0.80\linewidth]{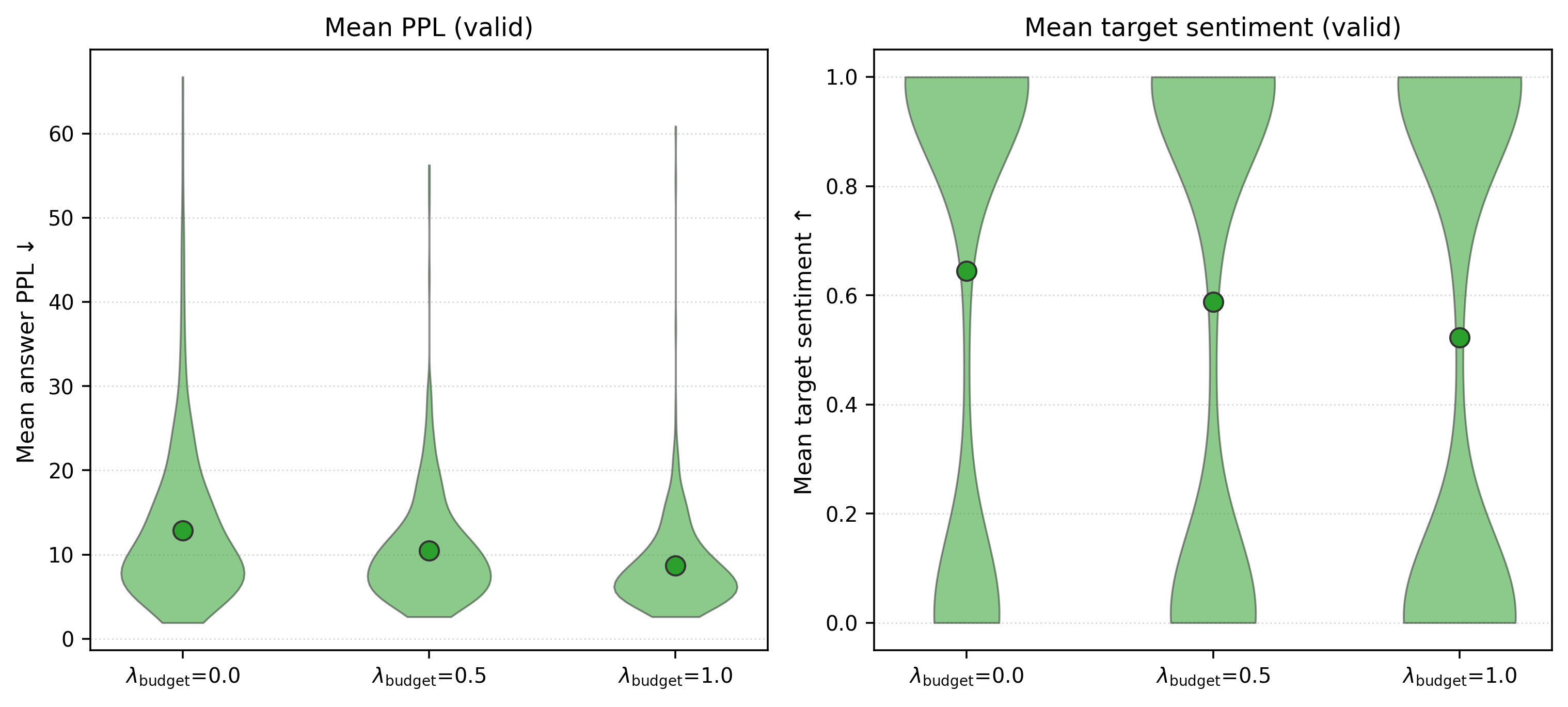}
    \vspace{0.45em}
    \caption{Adaptive-budget sweeps for Dream-7B sentiment under no activation
    steering, DiffMean, and ACT.}
    \label{fig:dream-adaptive}
\end{figure}
\FloatBarrier

\subsection{Sentiment: compute-matched comparison}

\begin{table}[H]
\caption{\textbf{Adaptive versus compute-matched stochastic fixed schedules,
Dream-7B sentiment.} Each adaptive configuration is compared with the same
scorer-free, non-adaptive stochastic schedule for that activation condition,
constructed as in Appendix~\ref{app:matched-schedule-details} and matched in
average denoising steps. Deltas are adaptive minus matched stochastic.
``Competitive'' has the same meaning as in
Table~\ref{tab:matched-stochastic-full}. Two entries are strict improvements on
both metrics at once---DiffMean at $\lambda_{\mathrm{budget}}=0$ ($+0.03$
target, $-0.8$ PPL) and no activation at $\lambda_{\mathrm{budget}}=0.5$
($+0.01$ target, $-2.2$ PPL).}
\label{tab:dream-matched}
\centering
\scriptsize
\setlength{\tabcolsep}{3.3pt}
\resizebox{\linewidth}{!}{%
\begin{tabular}{@{}lcccc@{}}
\toprule
Activation &
\makecell{Matched-fixed\\Target / PPL} &
\makecell{$\Delta$Target\\
$\lambda_{\mathrm{budget}}=0\to0.5\to1$} &
\makecell{$\Delta$PPL\\
$\lambda_{\mathrm{budget}}=0\to0.5\to1$} &
\makecell{Competitive\\$0\;/\;0.5\;/\;1$} \\
\midrule

None
& $0.441$ / $13.68$
& $+0.06\!\to\!+0.01\!\to\!-0.01$
& $+0.1\!\to\!-2.2\!\to\!-4.3$
& \cmark / \cmark / \cmark \\

DiffMean
& $0.655$ / $12.15$
& $+0.03\!\to\!-0.02\!\to\!-0.03$
& $-0.8\!\to\!-2.5\!\to\!-3.7$
& \cmark / \cmark / \cmark \\

ACT
& $0.611$ / $12.71$
& $+0.03\!\to\!-0.02\!\to\!-0.09$
& $+0.2\!\to\!-2.3\!\to\!-4.0$
& \cmark / \cmark / \cmark \\

\bottomrule
\end{tabular}}
\end{table}

\begin{figure}[H]
    \centering
    {\small\bfseries No activation}\par\vspace{0.15em}
    \includegraphics[width=0.80\linewidth]{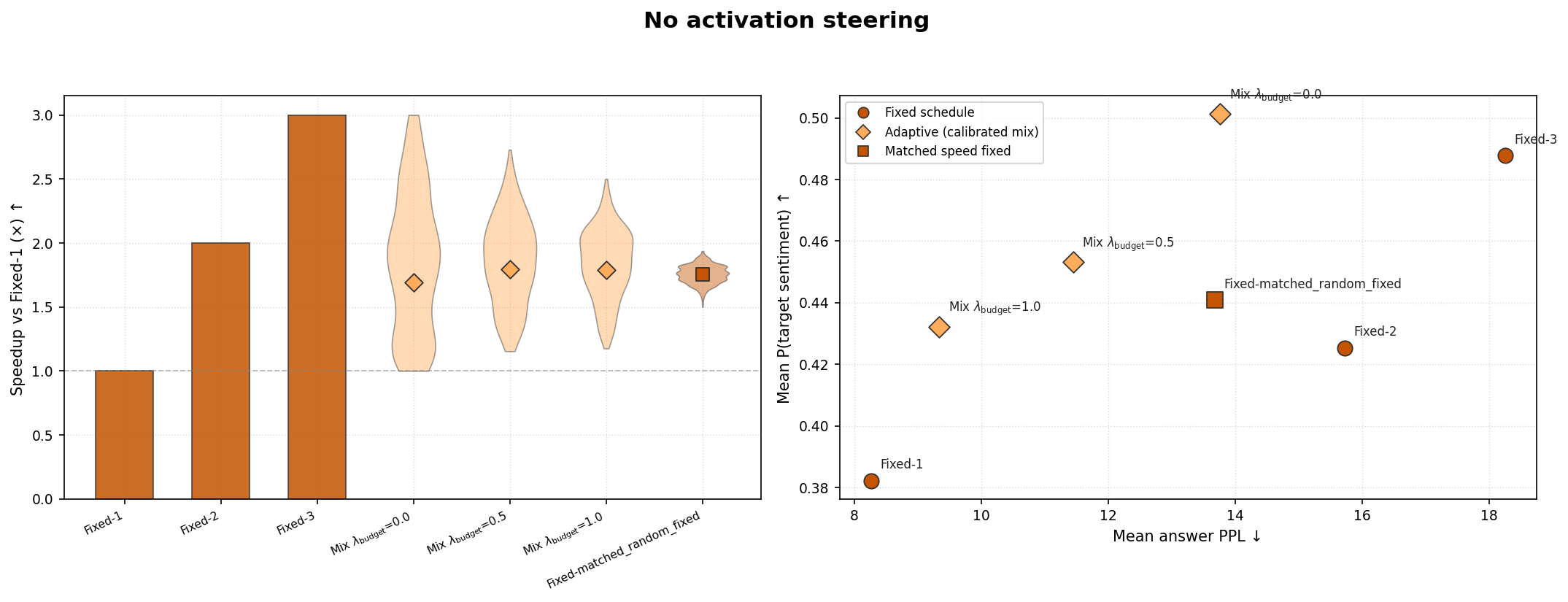}
    \vspace{0.45em}

    {\small\bfseries DiffMean}\par\vspace{0.15em}
    \includegraphics[width=0.80\linewidth]{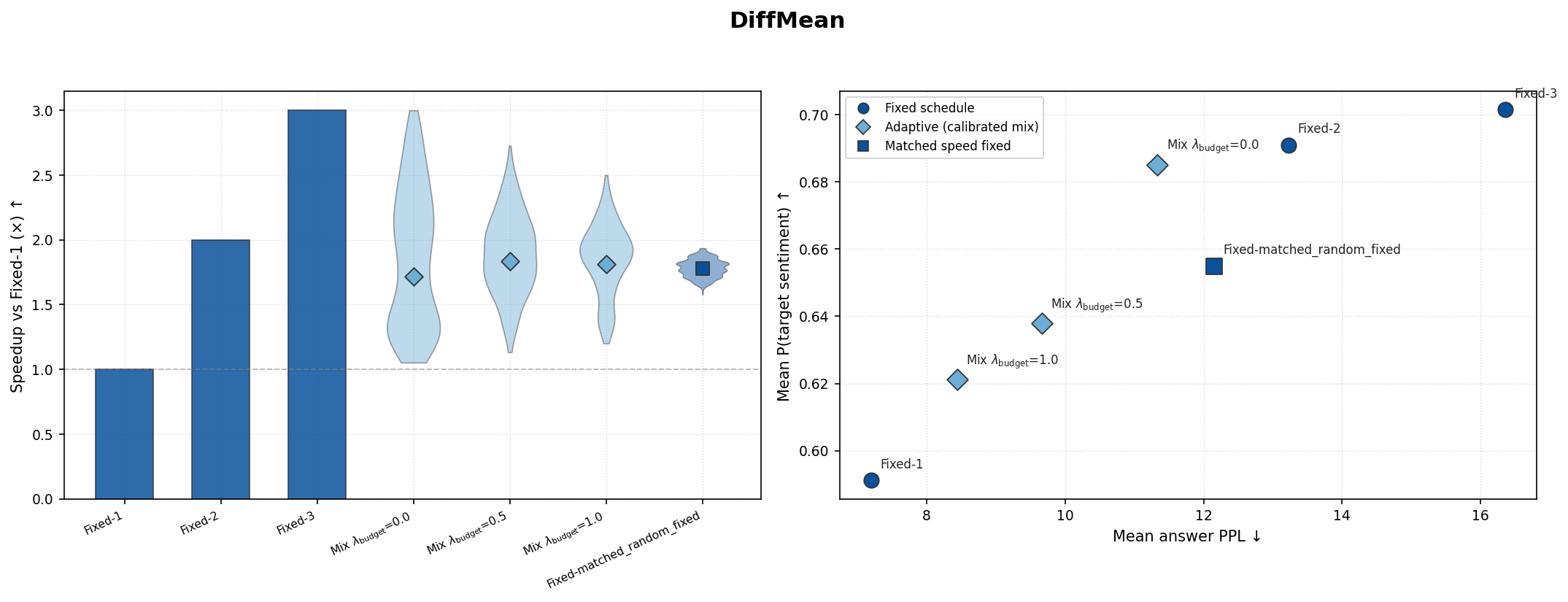}
    \vspace{0.45em}

    {\small\bfseries ACT}\par\vspace{0.15em}
    \includegraphics[width=0.80\linewidth]{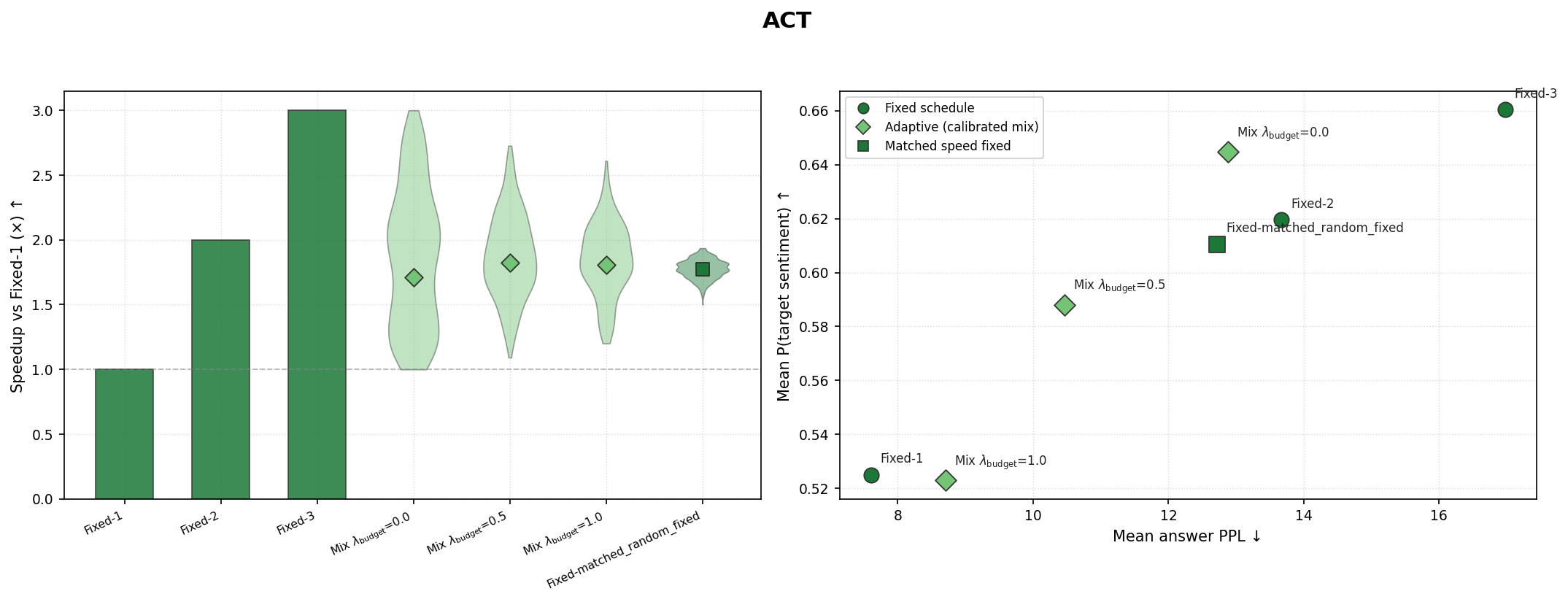}
    \vspace{0.45em}
    \caption{Adaptive versus compute-matched stochastic fixed schedules for
    Dream-7B sentiment under no activation steering, DiffMean, and ACT.
    Fixed-$1/2/3$ are shown as contextual anchors; the matched stochastic point
    is the controlled comparison.}
    \label{fig:dream-matched}
\end{figure}
\FloatBarrier

\clearpage
\subsection{NLI entailment}
\label{app:dream-nli}

Sentiment is a control target: it shares one preferred direction across examples
and admits a reusable steering direction. NLI entailment does not. It is
example-specific, recomputed for every dialogue and denoising state, and there is
no single activation direction that improves it in general. It is therefore the
setting in which commitment control has no activation-steering counterpart, and
the one that tests whether the mechanism transfers across model families for a
target type that activation steering cannot address. We evaluate it on Dream-7B
with no activation intervention, matching the LLaDA NLI configuration.

\begin{table}[H]
\caption{\textbf{Commitment-order and adaptive-budget sequences, Dream-7B NLI
entailment.} Both mechanisms behave as on LLaDA: the target score and PPL are
non-increasing as the commitment weight moves toward its confidence-driven end.
Commitment order holds $u_t\equiv1$, so its target movement occurs at a fixed
$60$ denoising steps.}
\label{tab:dream-nli-order}
\label{tab:dream-nli-budget}
\centering
\scriptsize
\setlength{\tabcolsep}{3.2pt}
\resizebox{\linewidth}{!}{%
\begin{tabular}{@{}lccc@{}}
\toprule
Control &
\makecell{Target\\(increasing $\lambda$)} &
\makecell{PPL\\(increasing $\lambda$)} &
\makecell{Ordering consistent\\target / PPL} \\
\midrule

Commitment order,
$\lambda_{\mathrm{order}}=0.5\to0.75\to0.9\to1$
& $0.597\!\to\!0.469\!\to\!0.393\!\to\!0.337$
& $27.35\!\to\!25.01\!\to\!23.50\!\to\!22.40$
& \cmark / \cmark \\

Adaptive budget,
$\lambda_{\mathrm{budget}}=0\to0.5\to1$
& $0.430\!\to\!0.336\!\to\!0.263$
& $33.28\!\to\!28.60\!\to\!24.83$
& \cmark / \cmark \\

\bottomrule
\end{tabular}}
\end{table}

Counterfactual ordering raises mean entailment from $0.337$ to $0.597$ while
perplexity rises from $22.40$ to $27.35$. Since the budget is fixed, this
movement is produced entirely by changing which provisional summary tokens
become permanent first, at identical denoising cost. The corresponding LLaDA
sweep moves entailment from $0.31$ to $0.54$ at $22.5$ to $26.6$ perplexity, so
the two families span comparable ranges on both axes. Re-scoring these same
summaries with an independent entailment model preserves the ordering on both
families (Appendix~\ref{app:heldout-scorer}).

\begin{table}[H]
\caption{\textbf{Adaptive versus compute-matched stochastic fixed schedule,
Dream-7B NLI entailment.} Deltas are adaptive minus matched stochastic, at
matched average denoising cost. All three adaptive configurations are
competitive.}
\label{tab:dream-nli-matched}
\centering
\scriptsize
\setlength{\tabcolsep}{3.3pt}
\resizebox{\linewidth}{!}{%
\begin{tabular}{@{}lcccc@{}}
\toprule
Activation &
\makecell{Matched-fixed\\Target / PPL} &
\makecell{$\Delta$Target\\
$\lambda_{\mathrm{budget}}=0\to0.5\to1$} &
\makecell{$\Delta$PPL\\
$\lambda_{\mathrm{budget}}=0\to0.5\to1$} &
\makecell{Competitive\\$0\;/\;0.5\;/\;1$} \\
\midrule

None
& $0.343$ / $31.50$
& $+0.09\!\to\!-0.01\!\to\!-0.08$
& $+1.8\!\to\!-2.9\!\to\!-6.7$
& \cmark / \cmark / \cmark \\

\bottomrule
\end{tabular}}
\end{table}

\begin{figure}[H]
    \centering
    \includegraphics[width=0.76\linewidth]{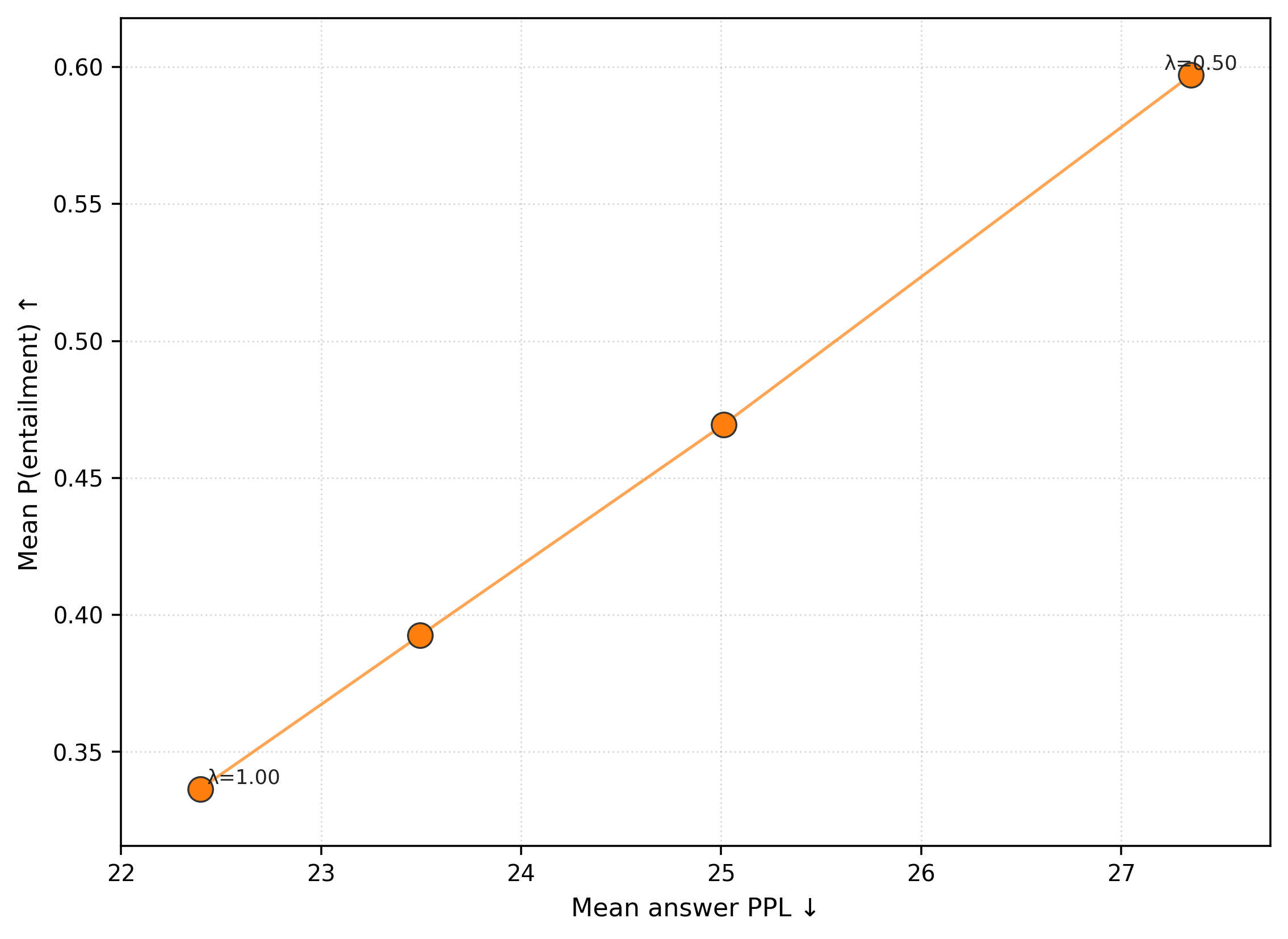}
    \caption{Counterfactual commitment-order trade-off for Dream-7B NLI
    entailment. Mean entailment probability versus PPL on SAMSum, with
    $u_t\equiv1$ and no activation steering.}
    \label{fig:dream-nli-counterfactual}
\end{figure}
\FloatBarrier

\begin{figure}[H]
    \centering
    \includegraphics[width=0.82\linewidth]{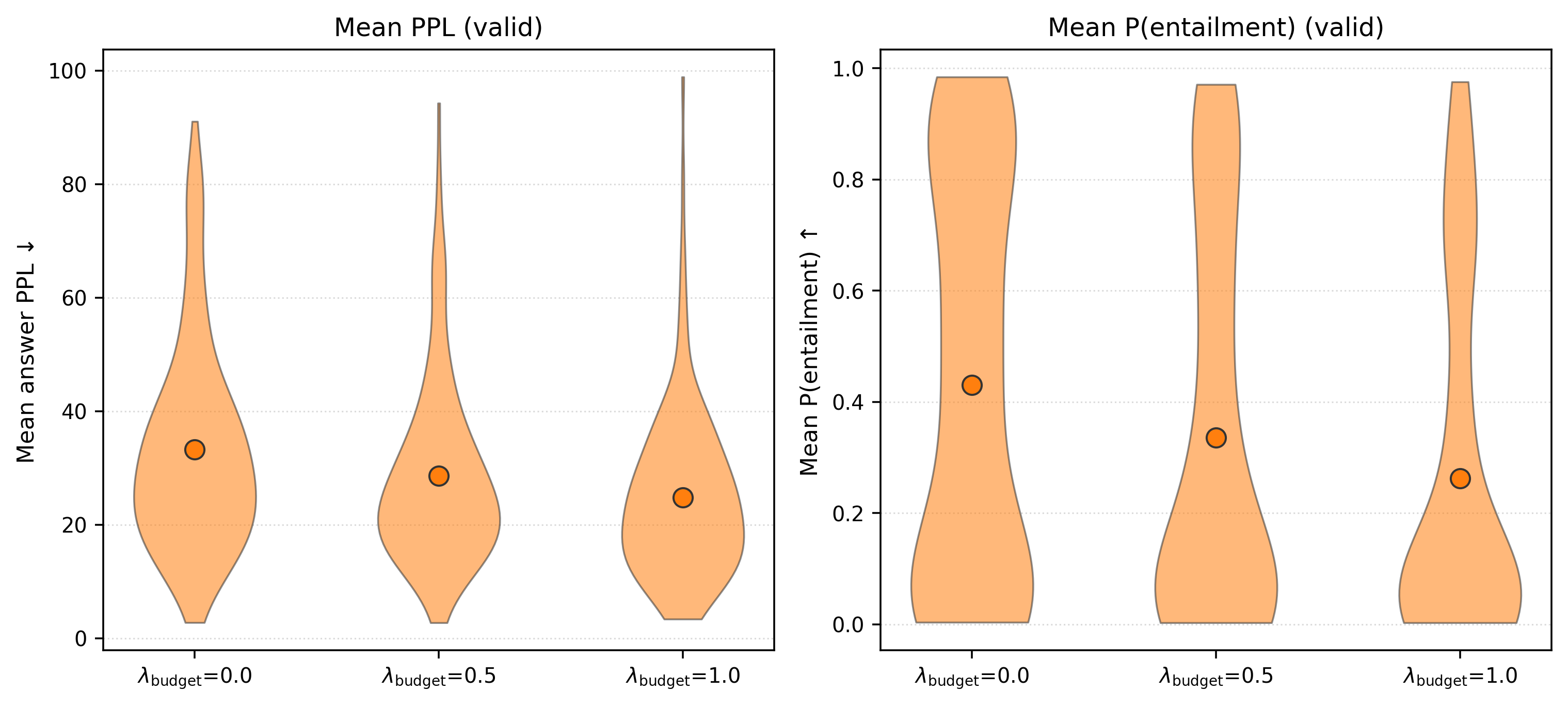}
    \caption{Adaptive-budget sweep for Dream-7B NLI entailment.}
    \label{fig:dream-nli-adaptive}
\end{figure}
\FloatBarrier

\begin{figure}[H]
    \centering
    \includegraphics[width=0.82\linewidth]{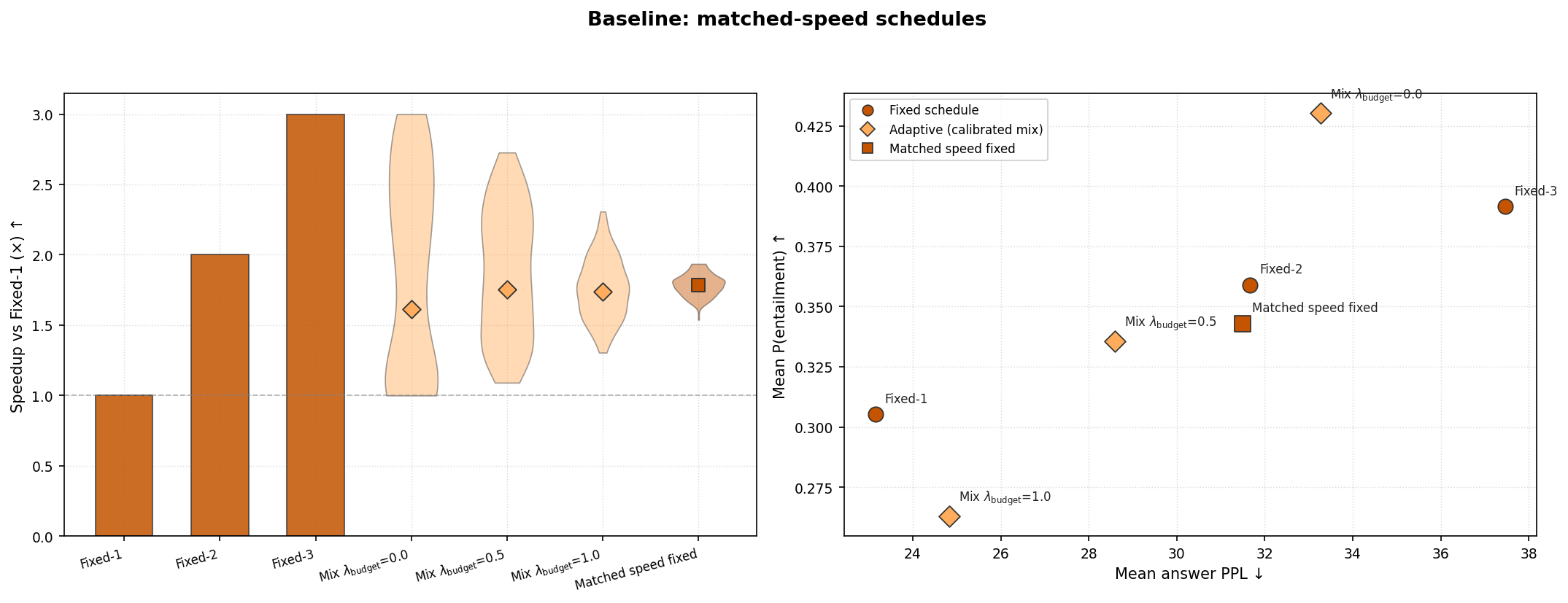}
    \caption{Adaptive versus compute-matched stochastic fixed schedule for
    Dream-7B NLI entailment. The matched baseline is scorer-free, non-adaptive,
    confidence-ranked, and matched in average denoising steps.}
    \label{fig:dream-nli-matched}
\end{figure}
\FloatBarrier

\subsection{Summary}

Both commitment axes reproduce on Dream-7B, for a control target under three
activation conditions and for a task-success target with no activation
intervention. All four commitment-order sweeps and all four adaptive-budget
sweeps are non-increasing in target and PPL, and all twelve compute-matched
comparisons are competitive, two of them strict improvements on both metrics.
This holds despite differences in training lineage, tokenizer, depth, steering
window, and intervention strength. Both the sentiment and the NLI orderings are
additionally preserved under independent scorers
(Appendix~\ref{app:heldout-scorer}). The remaining settings---subjectivity and
the two vision-language objectives---are evaluated on the LLaDA family only.

\FloatBarrier
\end{document}

%% file: math_commands.tex
\usepackage{amsmath,amsfonts,bm}

\def\eqref#1{equation~\ref{#1}}

\def\1{\bm{1}}

\DeclareMathAlphabet{\mathsfit}{\encodingdefault}{\sfdefault}{m}{sl}
\SetMathAlphabet{\mathsfit}{bold}{\encodingdefault}{\sfdefault}{bx}{n}

